\documentclass[11pt]{article}
\usepackage{amsmath,amsthm,amsfonts,amssymb}
\usepackage{mathtools}

\usepackage{graphicx}
\usepackage{subcaption}
\usepackage{float}

\usepackage{booktabs}
\usepackage{multirow}

\usepackage[authoryear]{natbib}
\usepackage[colorlinks=true,
            citecolor=blue,
            urlcolor=blue]{hyperref}

\usepackage{cleveref}
\usepackage{listings}
\usepackage{placeins}
\usepackage[dvipsnames]{xcolor}
\usepackage{listings}
\usepackage{lstbayes}
\usepackage[T1]{fontenc}
\usepackage[utf8]{inputenc}

\providecommand{\keywords}[1]
{
  \small	
  \textbf{\textit{Keywords---}} #1
}

\theoremstyle{plain}
\newtheorem{theorem}{Theorem}[section]
\newtheorem{lemma}[theorem]{Lemma}

\theoremstyle{definition}
\newtheorem{definition}[theorem]{Definition}

\theoremstyle{remark}

\title{Dirichlet Scale Mixture Priors for Bayesian Neural Networks}
\author{
August Arnstad\thanks{Department of Statistics \& Data Science, University of Oslo. Email: augusa@math.uio.no} \and
Leiv Rønneberg\thanks{Department of Statistics \& Data Science, University of Oslo. Email: ltronneb@math.uio.no} \and
Geir Storvik\thanks{Department of Statistics \& Data Science, University of Oslo. Email: geirs@math.uio.no}
}

\date{}

\begin{document}

\maketitle

\begin{abstract}
Neural networks are the cornerstone of modern machine learning, yet can be difficult to interpret, give overconfident predictions and are vulnerable to adversarial attacks. Bayesian neural networks (BNNs) provide some alleviation of these limitations, but have problems of their own. The key step of specifying prior distributions in BNNs is no trivial task, yet is often skipped out of convenience. In this work, we propose a new class of prior distributions for BNNs, the Dirichlet scale mixture (DSM) prior, that addresses current limitations in Bayesian neural networks through structured, sparsity-inducing shrinkage. Theoretically, we derive general dependence structures and shrinkage results for DSM priors and show how they manifest under the geometry induced by neural networks. In experiments on simulated and real world data we find that the DSM priors encourages sparse networks through implicit feature selection, show robustness under adversarial attacks and deliver competitive predictive performance with substantially fewer effective parameters. In particular, their advantages appear most pronounced in correlated, moderately small data regimes, and are more amenable to weight pruning. Moreover, by adopting heavy-tailed shrinkage mechanisms, our approach aligns with recent findings that such priors can mitigate the cold posterior effect, offering a principled alternative to the commonly used Gaussian priors.
\end{abstract}


\section{Introduction}
For machine learning tasks, neural networks (NNs) are widely applied in a variety of settings, due to their ability to model complex relationships in high-dimensional data. They do so by modeling responses $\mathbf{y} \in \mathbb{R}^d$ with the function
\begin{equation}
\begin{aligned}
    f_1(\mathbf{x}) &= W_1 \mathbf{x} + b_1 \\ 
    f_{\ell}(\mathbf{x}) &= W_{\ell}\varphi(f_{\ell -1}(\mathbf{x})) + b_{\ell}, \quad l=2,\ldots,L \\ 
    \hat{\mathbf{y}}(\mathbf{x}) &= f_L(\mathbf{x}),
\end{aligned}
\end{equation}
where $\mathbf{x} \in \mathbb{R}^p$ is the input, $W_\ell \in \mathbb{R}^{h_\ell \times h_{\ell-1}}$ and $b_\ell \in \mathbb{R}^{h_\ell}$ are weight matrices and bias vectors, $\varphi$ is an element-wise non-linear activation function, $\hat{\mathbf{y}}(\mathbf{x}) \in \mathbb{R}^d$ is the network output, and $p(\cdot \mid f_L(\mathbf{x}))$ denotes the likelihood model. However, due to over-parametrization and growing size, they are hard to interpret and often overconfident in their predictions \citep{arbel_primer_2023}. Bayesian neural networks (BNNs) promise to mitigate predictive overconfidence by incorporating uncertainty into the predictions, at the cost of increased computational complexity \citep{fortuin2022bayesian}. 
A BNN is a neural network in which one places a prior distribution $p(\Theta)$ over the network parameters $\Theta = \{W_{\ell}, b_{\ell}\}_{\ell =1, ..., L}$, and aims to infer the posterior distribution 
\begin{equation}
    p(\Theta \mid\mathbf{y}) = \frac{p(\mathbf{y} \mid \Theta) p(\Theta)}{p(\mathbf{y})} \ .
\end{equation}

The distributional formulation extends inference beyond single point estimates, making it possible to study uncertainty and other distributional properties of the model. This generality, however, comes at the cost of significant computational and methodological challenges. In order to define a BNN, one must choose the prior distribution such that it reflects ones prior beliefs about the parameters in the model. Specifying such beliefs is difficult, especially in BNNs where a large number of parameters with complex interactions make it unclear how prior information should be encoded. As a consequence, because specifying a prior is difficult, the standard in many BNN applications has been to choose the simplest prior of all,  isotropic Gaussian distributions, $p(\Theta)=\mathcal{N}(0, \alpha^2\mathbf{I})$ with $\alpha$ typically chosen to scale inversely with the square root of the layer width. The Gaussian prior is regarded as uninformative and has convenient sampling properties, making it a popular choice~\citep{fortuin2022priors}. However, it has recently been pointed out that the choice of prior can greatly affect the posterior distribution in BNNs, and that tempering the posterior can significantly improve performance, a phenomenon referred to as the \textit{cold posterior effect} \citep{wenzel2020goodbayesposteriordeep, tran_all_2022, fortuin2022priors}, suggesting that either the likelihood or the prior is misspecified. Because of the cold posterior effect, there has been a growing interest in more complex priors, such as sparsity-inducing priors, functional priors, structured priors, and hierarchical priors \citep{louizos_bayesian_compression_2017, ghosh2019model, tran_all_2022}. Many of the prior distributions that have been proposed are so-called scale mixture priors, where the prior structure is encoded in the variance of the prior distribution \citep{bhattacharya_dirichletlaplace_2015}. 

This paper introduces a structured extension of classical global-local scale mixture priors for BNNs, in which an additional joint scale is used to regularize parameters at a group-level alongside the usual global and local components. Building on ideas from \citet{bhattacharya_dirichletlaplace_2015, nagel2024alternative} on competitive shrinkage via Dirichlet distributions, we define our novel class as Dirichlet Scale Mixture (DSM) priors. The main idea behind the DSM prior class is to use the Dirichlet distribution to allocate a fixed variance budget in a way that is structurally natural for neural networks. Similar approaches with the Dirichlet distribution have been taken, for example for generalized linear mixed models \citep{r2d2_Yanchenko}. In our approach, all weights mapping into the same node in a hidden layer share a group-specific variance, allotted out to individual weights according to a Dirichlet component. We expect this to encourage sparse solutions, as entire nodes may be strongly shrunk through the group scale, while individual weights within a group are further shrunk at the local level.

 In this paper we analyse how this structured prior assumption translates into dependence, sparsity and effective model complexity in BNNs. We derive general theoretical properties of the a priori dependence and shrinkage behavior induced by DSM priors and investigate how these properties manifest in neural network settings. Empirically, we compare DSM priors to standard global-local alternatives and show that they consistently yield networks that are more amenable to pruning, rely on substantially fewer effective parameters, and remain competitive in terms of predictive performance.

 The remainder of the paper is organized as follows. In \Cref{sec:related_work}, a brief review of the literature is given. \Cref{sec:DSM_priors} introduces the DSM prior class, highlights similar priors and describes its application to neural network models. The dependence structure and shrinkage properties of the DSM priors are given in \Cref{sec:shrinkage_DSM_priors}, before showcasing experiments in \Cref{sec:experiments}. The article is wrapped up with a discussion in \Cref{sec:discussion}. The Supplementary Material \ref{supp} is organized into five sections. It contains additional analysis of the dependence structure; detailed lemmas and proofs of theorems; further exposition of the linearization procedure; extended experimental results including convergence diagnostics and implementation details; and a complete code example.

\section{Related work}
\label{sec:related_work}
The field of Bayesian neural networks has received significant interest due to their unique properties. In particular, their probabilistic formulation naturally incorporates Occam’s razor by favoring simpler explanations unless the data provide strong evidence for more complex models, while also remaining robust against overconfident predictions \citep{mackay_bayesian_1992, bishop_neural_1995}. However, recent findings such as the cold posterior effect, has raised questions towards both inference techniques and prior specification.\citet{wenzel2020goodbayesposteriordeep} hypothesize that techniques in deep learning may compromise the likelihood, or that the Gaussian priors are inadequate. Furthermore,  \citet{izmailov2021bayesianneuralnetworkposteriors} argue that the cold posterior effect is primarily driven by data augmentation, frequently employed in deep learning, and \citet{marek2024confidentpriorreplacecold} attribute it to model misspecification that leads to underfitting or inflated estimates of aleatoric uncertainty. \citet{fortuin2022bayesian} argues that the cold posterior is dependent on the architecture, and that data augmentation does not remove the cold posterior effect for all models. They advocate tailoring the prior based on the architecture, showing that, fully connected layers trained with stochastic gradient descent methods are heavy-tailed and recommend reflecting this in priors also for BNNs. This raises the question whether priors should mimic the heavy-tailed behavior, reinforcing it, or counteract it, depending on the desired inductive bias.

 The cold posterior effect questions both inference techniques and prior specification in Bayesian deep learning. Inference techniques such as variational dropout in neural networks are interpreted as an approximation to (deep) Gaussian processes \citep{gal16_dropout}. Although not intrinsically Bayesian, such techniques can help our understanding of the distributional properties in BNNs. Notably, \citet{molchanov2017_sparse_dropout} demonstrate that dropout produces extremely sparse neural networks with negligible accuracy loss. However, despite its empirical success, \citet{hron2018variationalbayesiandropoutpitfalls} point out that variational dropout suffers from improper priors, leading to posterior pathologies that cannot be remedied. From a more theoretical perspective, \citet{vladimirova_understanding_2019} show that the distribution of nodes in a BNN with Gaussian i.i.d. priors, become sub-Weibull distributed in deeper layers, highlighting how prior choices alone can induce strong structural properties. This further stresses the need for a deeper understanding of prior distributions in Bayesian neural networks.

The literature on BNNs primarily considers priors on the weights, as they govern the network’s functional complexity. While Gaussian i.i.d. priors on the weights are attractive due to their ease of sampling and analytical tractability, their interpretation in neural networks is difficult, as the heavy overparameterization obscures the relationship between weight distributions and the underlying data. Any prior parameterized on the weights combined with a deterministic network architecture induces a prior in the function space. Therefore, a line of work focuses on desirable properties the network should have in function space. \citet{nalisnick_2021_PREDCPs} extend the penalizing complexity prior by \citet{simpson_pc_priors_2017}, to yield a predictive complexity prior that penalizes large deviation in predictions, by comparing the network to a less complex base model. Furthermore, \citet{tran_all_2022} match the induced functional prior to a Gaussian process via an optimization scheme based on a distance measure between the GP and the network, to make the prior exhibit interpretable properties. 

Another line of priors are motivated by sparsity, not only for its computational benefits, but also because of concepts such as the lottery ticket hypothesis \citep{frankle2019lotterytickethypothesisfinding}. The lottery ticket hypothesis indicates that there exists subnetworks which give roughly the same performance as the overparameterized networks. Obtaining these networks is not trivial, but the class of sparsity-inducing priors have shown promising results. Sparsity-inducing priors have been widely studied for standard regression models, but their effect on BNNs have not been as extensively investigated. 
The classical spike-and-slab prior \citep{mitchell_bayesian_1988} induces sparsity similar to the Bernoulli dropout \citep{boluki2020learnable} and can also be used for model selection \citep{hubin2023variational}. Furthermore, one of the most popular sparsity-inducing priors is the horseshoe prior \citep{carvalho_horseshoe_2009}. With a high concentration of mass near zero and heavy tails, it shrinks most weights to zero, while allowing a few weights to escape shrinkage through locally large scales. The horseshoe prior belongs to the popular class of global-local shrinkage priors, which use one global scale to control overall shrinkage and one local scale to allow some coefficients to escape shrinkage. Many global-local priors can be expressed as part of the larger class of scale mixture Gaussian priors \citep{polson2011shrink}.  \citet{bhattacharya_dirichletlaplace_2015} conjectures, based on strong empirical evidence, that the horseshoe leads to optimal shrinkage rates. However, as the theoretical properties of the horseshoe are not fully clear, \citet{bhattacharya_dirichletlaplace_2015} introduces the Dirichlet Laplace prior, which is shown to attain optimal shrinkage. Another Gaussian scale mixture for regression models is the Dirichlet horseshoe prior \citep{nagel2024alternative}, for which we develop new theoretical foundations and extend to the neural network setting. 

Structure can also be introduced into sparsifying priors to encourage group-wise shrinkage~\citep{ghosh2019model, louizos_bayesian_compression_2017}. By letting either all incoming or all outgoing weights of a neuron share the same scale parameter, shrinkage acts on groups of weights rather than on each weight individually, with the possibility of an additional global scale controlling the overall level of sparsity. In \citet{louizos_bayesian_compression_2017}, this structure is exploited to prune entire neurons by thresholding modes of outgoing weights, thereby reducing the fixed point precision required to represent the network. On the other hand, \citet{ghosh2019model} consider incident weights and use the structure for model selection.

\section{The Dirichlet Scale Mixture (DSM) priors}
\label{sec:DSM_priors}
We now extend the global-local shrinkage framework by introducing joint regularization using the Dirichlet distribution. Let $\mathbf{w}_j = (w_{j1}, \dots, w_{jp})^{\top}$ denote a generic group of coefficients of length $p$. The Dirichlet scale mixture (DSM) prior is defined hierarchically by
\begin{equation}
\begin{aligned}
w_{jk} \mid \tau, \lambda_{j}, \xi_{jk} 
&\sim \mathcal{N}\!\left(0,\, \tau^2 \lambda_{j}^2 \xi_{jk}\right), \\
(\xi_{j1},\dots,\xi_{jp}) 
&\sim \mathrm{Dir}(\alpha_1,\dots,\alpha_p)\in \Delta^{p-1}, \\
\lambda_{j} &\sim \mathcal{P}_{\lambda}, \\
\tau &\sim \mathcal{P}_{\tau}.
\end{aligned}
\end{equation}
where $\Delta^{p-1}$ denotes the standard $p-1$ Euclidean simplex, $\alpha_k, k=1, \dots, p$ are the concentration parameters, $\tau>0$ is a global scale, $\lambda_j > 0$ a group scale and $\boldsymbol{\xi}_j$ a simplex-valued random vector with $\sum_{k=1}^p\xi_{jk}=1$. We restrict our attention to the symmetric case $\alpha_k=\alpha$ for all $k$. The distributions $\mathcal{P}_{\lambda}$ and $\mathcal{P}_{\tau}$ are unspecified positive prior distributions, governing the amount of shrinkage induced at the group and global levels, respectively. 

The intuition is to treat parameters in groups with a fixed variance budget $\tau^2\lambda_j^2$, and let the Dirichlet component distribute this variance within each group. This induces negative dependence among the $\xi_{jk}$ through the simplex constraint, coupling prior variances and promoting competition and sparsity. The DSM hierarchy thus imposes three levels of shrinkage: a global scale $\tau$, group-specific scales $\lambda_j$, and a joint allocation $(\xi_{j1},\dots,\xi_{jp})$ that couples coefficients within each group. 

The grouping used by the DSM prior is model dependent. In linear regression there is no comparable architectural grouping, and we therefore assign coefficient-specific local scales, while using a single Dirichlet vector to allocate variance across coefficients. In contrast, for models such as neural networks, meaningful groups arise naturally from the architecture itself. In the Bayesian neural network setting, we exploit this structure by assigning priors at multiple levels. We share the global scale $\tau$ across the layer, let all incoming weights to a node $j$ share a group-specific scale $\lambda_j$, and use Dirichlet components $\xi_{jk}$ to govern how variance is allocated among the incoming weights. This is similar to the ideas in \citet{r2d2_Yanchenko}, who place a Dirichlet distribution on the variance components of coefficients in generalized linear mixed models. They do so by placing a Beta prior on the coefficient of determination $R^2$, in order to induce a Beta prime prior on the total variance that is subsequently allocated via a Dirichlet distribution.

The concentration parameter $\alpha$ controls the level of sparsity within groups. Small values of $\alpha$ encourage highly uneven allocations in which only a few coefficients receive substantial variance, while $\alpha=1$ corresponds to a uniform distribution on the simplex. As $\alpha$ grows large, the Dirichlet distribution concentrates around $(1/p,\dots,1/p)$ and the allocation becomes increasingly uniform. In this regime, the dependence induced by the simplex constraint vanishes and the DSM prior reduces to a standard global-local scale mixture with group-level variance $\tau^2\lambda_j^2/p$. This limiting case connects the DSM framework directly to classical shrinkage priors, most notably the horseshoe. Depending on whether the group-level scale $\lambda_j$ is retained or replaced by parameter-specific scales, this limit recovers either a grouped or a fully local version of the horseshoe prior. It is defined by placing half-Cauchy scales both locally and globally,
\begin{equation*}
\begin{aligned}
    w_{jk} \mid \tau, \lambda_{j}
    &\sim \mathcal{N}\!\left(0,\, \tau^2 \lambda_{j}^2\right), \\
    \lambda_{j}
    &\sim \mathcal{C}^+(0,1), \\
    \tau
    &\sim \mathcal{C}^+(0,\tau_0^2).
\end{aligned}
\end{equation*}
The horseshoe has been shown to effectively capture signals in high-dimensional settings, as most parameters are shrunk aggressively toward zero while a few escape due to the locally heavy tails \citep{van_der_pas_horseshoe}. While this shrinkage profile is a key strength of the horseshoe, it also implies that coefficients escaping shrinkage are only weakly regularized.

\subsubsection*{The Dirichlet Student’s t prior}
In this paper, we will be mostly concerned with the DSM priors that use a half-Cauchy distribution for $\tau$ and let $\lambda_i$ follow a half-Student’s t distribution.
By letting $\mathcal{P}_{\lambda}=t^+_{\nu}$, the group regularization is governed by the tails of the Student’s t, which is dependent on the degrees of freedom (df), $\nu$. This means that the Dirichlet Student’s t prior allows for flexible shrinkage controlled by tuning $\nu$. When $\nu$ is small, the distribution becomes heavy-tailed, enforcing strong shrinkage for most draws while allowing a few to take large values. As $\nu$ increases, the tails become lighter and the shrinkage becomes more uniformly moderate. We define the Dirichlet Student’s t prior as
\begin{equation*}
\begin{aligned}
    w_{jk} \mid \tau, \lambda_{j}, \xi_{jk}
    &\sim \mathcal{N}\!\left(0,\, \tau^2 \lambda_{j}^2 \xi_{jk}\right), \\
    (\xi_{j1},\dots,\xi_{jp})
    &\sim \mathrm{Dir}(\alpha,\dots,\alpha) \in \Delta^{p-1}, \\
    \lambda_{j}
    &\sim t^+_{\nu}(0,1), \\
    \tau
    &\sim \mathcal{C}^+(0,\tau_0^2).
\end{aligned}
\end{equation*}
where $\tau_0$ is a hyperparameter to be chosen. By choosing $\nu=1$, the Dirichlet Student’s t prior becomes a Dirichlet horseshoe prior \citep{nagel2024alternative}, as $t^+_{1} = \mathcal{C}^+(0, 1)$. 

\subsection{Regularization of the DSM priors}
\label{subsec:regularization}
Heavy-tailed scale priors such as the half-Cauchy and low-df Student’s t can produce extremely large local scales, which may cause numerical instabilities and slow mixing in posterior sampling. Following \citet{piironen2017sparsity}, we therefore apply a soft regularization to the local scales. This modification preserves shrinkage behavior, while substantially improving computational robustness. We regularize by replacing each group scale $\lambda_j$ by a regularized version
\begin{equation}
\label{eq:tilde_lambda}
    \tilde{\lambda}_j^2 = \frac{c^2 \lambda_j^2}{c^2 + \tau^2 \lambda_j^2}, 
    \qquad c^2 \sim \mathrm{InvGamma}(a,b) \ ,
\end{equation}
where the hyperparameter $c$ controls the degree of truncation. When $c^2\gg \tau^2\lambda_j^2 
$, the prior reduces to the original heavy-tailed form, whereas for $c^2\ll \tau^2\lambda_j^2$ the local variance is bounded by $c^2$. This transformation can be applied generically to any DSM variant by substituting $\lambda_j$ with $\tilde{\lambda}_j$. The choices of $a$ and $b$ determine the effective slab behavior. Following \citet{piironen2017sparsity}, we set $a=\nu_{c^2}/2=$, $b=\nu_{c^2} s^2/2$, which induces a scaled half-Student’s t distribution with scale $s$ on the slab component. Here, $\nu_{c^2}$ denotes the degrees of freedom controlling tail heaviness, while $s$ determines the typical magnitude of coefficients that escape shrinkage. In our experiments, we set $\nu_{c^2}=4, s^2=2$, corresponding to a moderately heavy tailed slab.

\section{Properties of the DSM priors}
\label{sec:shrinkage_DSM_priors}
In this section, we characterize the dependence structure and shrinkage behavior induced by the DSM priors.
\subsection{Dependence structure}
We study the dependence induced by the DSM prior through the variance components that govern the weights within each group. In our model, the prior variance of weight $w_{jk}$ is given by
\begin{equation}
\mathrm{Var}(w_{jk} \mid \tilde{\lambda}_j^2, \xi_j)
=
\tau^2 \tilde{\lambda}_j^{\,2} \xi_{jk},
\end{equation}
where $\xi_j = (\xi_{j1},\dots,\xi_{jp}) \sim \mathrm{Dirichlet}(\alpha,\dots,\alpha)$ and $\tilde{\lambda}_j^2$ denotes the regularized local scale. To isolate the structural dependence induced by the shared scale, we focus on the variance components excluding the global factor $\tau^2$ and define
\begin{equation}
\label{eq:V_definition}
V_{jk} = \tilde{\lambda}_j^{\,2} \xi_{jk},
\qquad
V_{jl} = \tilde{\lambda}_j^{\,2} \xi_{jl},
\qquad k \neq l.
\end{equation}
A direct calculation (see Section 1 of the supplementary material \ref{ap:dependence}) yields
\begin{equation}
    \label{eq:cov_regularized_rewritten}
    \mathrm{Cov}(V_{jk}, V_{jl}) = \frac{1}{p^2(p\alpha+1)} \left(\alpha p\,\mathrm{Var}(\tilde{\lambda}_j^{\,2}) - \mathbb{E}[\tilde{\lambda}_j^{\,2}]^2 \right).
\end{equation}
The sign of the covariance is therefore governed by the square of the coefficient of variation
\begin{equation}
    CV^2(\tilde{\lambda}_j^2) = \frac{\mathrm{Var}(\tilde{\lambda}_j^{\,2})} {\mathbb{E}[\tilde{\lambda}_j^{\,2}]^2},
\end{equation}
in the sense that
\begin{equation}
    \label{eq:cov_sign_regularized_rewritten}
    \mathrm{Cov}(V_{jk}, V_{jl})
    \begin{cases}
    < 0, & \text{if } CV^2(\tilde{\lambda}_j^2) < \tfrac{1}{p\alpha}, \\
    = 0, & \text{if } CV^2(\tilde{\lambda}_j^2) = \tfrac{1}{p\alpha}, \\
    > 0, & \text{if } CV^2(\tilde{\lambda}_j^2) > \tfrac{1}{p\alpha} \ ,
    \end{cases}
\end{equation}
where the group size $p$ and the concentration parameter $\alpha$ directly modulate the threshold between negative and positive dependence.

\begin{figure}[htbp]
    \centering
    \includegraphics[width=0.9\textwidth]{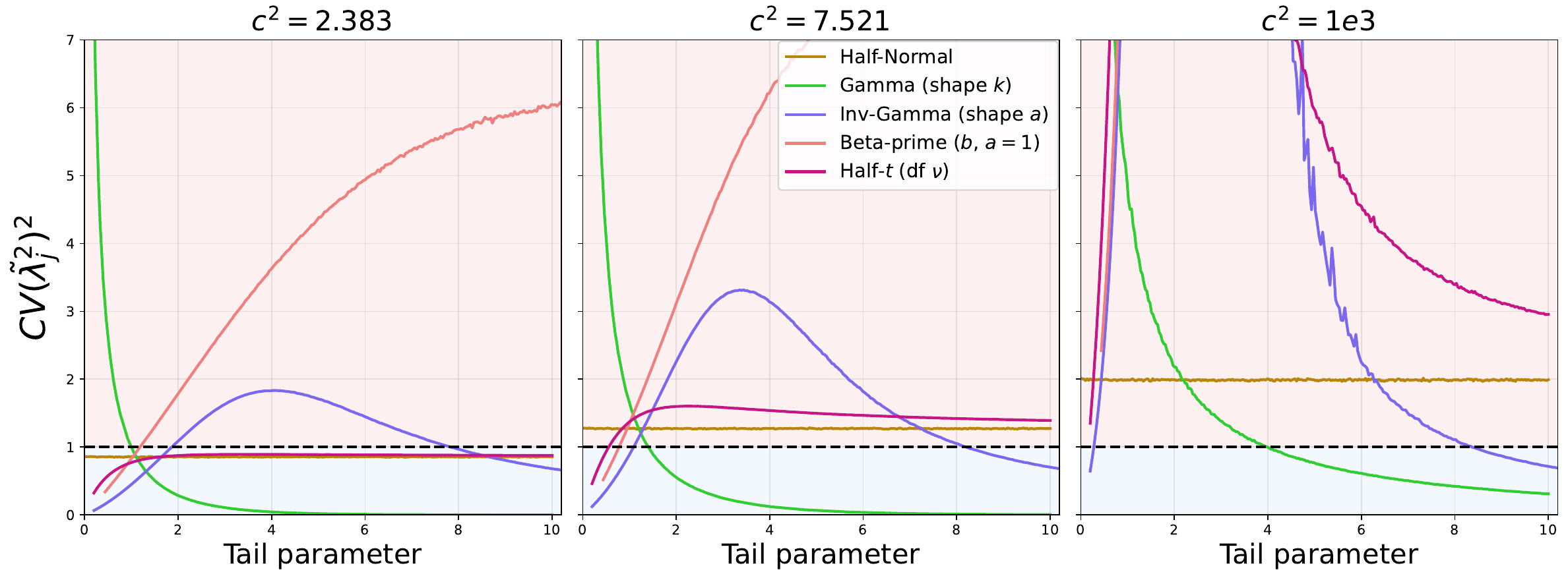}
    \caption{
    Monte Carlo estimates of the dispersion ratio $CV^2(\tilde{\lambda}_j^{\,2})$ for different scale priors as a function of their tail-controlling parameter $(\sigma, k, a, b, \nu)$. The curves are evaluated at three fixed values of the regularization parameter: (i) the median of the prior on $c$, (ii) the $0.9$ quantile of the prior on $c^2$, and (iii) a very large value of $c$, corresponding to an essentially unregularized regime. The horizontal line indicates the threshold $1/(p\alpha)$ separating negative and positive covariance regimes. Smaller values of $c$ attenuate dispersion and increasingly favor negative covariance among the variance components $\tilde{\lambda}_j^{\,2}\xi_{jk}$.
    }
    \label{fig:cov_structure_regularized}
\end{figure}

The qualitative behaviour of $CV^2(\tilde{\lambda}_j^2)$ depends on both the prior placed on $\lambda_j$ and the regularization map defining $\tilde{\lambda}_j^{\,2}$. Heavy-tailed priors on $\tilde{\lambda}_j$ tend to inflate dispersion and therefore promote positive dependence between components, whereas lighter-tailed priors favor negative dependence. An interesting special case is the Dirichlet–Laplace prior \citep{bhattacharya_dirichletlaplace_2015}, for which $\tilde{\lambda}^2_j \sim \mathrm{Gamma}(p\alpha, 1/2)$ and $CV^2(\tilde{\lambda}_j^2) = 1/p\alpha$, recovering the uncorrelated, and in fact independent, case. Furthermore, the regularization map imposes the deterministic bound $\tilde{\lambda}_j^{\,2} \le c^2/\tau^2$. As a consequence, for fixed $\tau$ and $c$, all moments of $\tilde{\lambda}_j^2$ exist even when $\lambda_j$ follows a heavy-tailed prior. In our model, the bound parameter $c^2$ is itself assigned an inverse gamma prior, meaning that the covariance sign will depend on $c^2$. Consequently, heavy-tailed priors on $\lambda_j$ still tend to increase $CV^2(\tilde{\lambda}_j^2)$ relative to light-tailed alternatives, but this effect is progressively attenuated as $c^2$ decreases. This attenuation effect is illustrated in \Cref{fig:cov_structure_regularized}, which displays the dispersion ratio $CV^2(\tilde{\lambda}_j^2)$ evaluated at three representative values of the regularization parameter $c^2$, namely the prior median of $c^2$, the $0.9$ quantile of the prior distribution of $c^2$, and a very large value of $c^2$. Large values of $c^2$ recover the behavior of the unregularized model, while smaller values enforce a stronger Gaussian envelope on the marginal weight distribution and favor negative dependence through reduced dispersion.

 A final remark is in order. The dependence described above arises through the variance components of the weights and is scaled by the global parameter $\tau$, which we take to follow a half-Cauchy prior. While $\tau$ controls the overall magnitude of the weights, its direct effect cancels out in relative quantities such as correlations, so that the dependence structure itself is independent of the global scale. The analysis above therefore characterizes genuine structural properties of the prior. At the same time, this dependence acts indirectly, entering through the variance components of the weights. This means that the induced dependence does not appear at the level of marginal means, but is expressed through relative dispersion and higher-order moments. In what follows, we shift focus to marginal shrinkage profiles, which provide a complementary and more directly interpretable summary of the regularization behavior induced by the DSM priors.

\subsection{Horseshoe for linear regression}
We now turn to shrinkage and model complexity under the DSM priors. We first study the scalar shrinkage factor $\kappa$ in a linear regression model with a horseshoe prior on the coefficients, following \citet{piironen2017sparsity}. This restriction to scalar shrinkage factors is deliberate, since for grouped parameters, shrinkage is naturally described by matrix-valued operators rather than scalars. We therefore develop the scalar theory first, and return to the grouped case when lifting the analysis to linearized Bayesian neural networks.

Consider the linear regression model
\begin{equation}
\label{eq:piironen_model}
\begin{aligned}
    y_i &= \mathbf{x}_i^{\top}\mathbf{w} + \varepsilon_i \qquad \varepsilon_i \sim \mathcal{N}(0, \sigma^2) \qquad i=1, \dots n \\
\end{aligned}
\end{equation}
where $\mathbf{x}_i, \mathbf{w} \in \mathbb{R}^{p}$ and we assume $\mathbf{X}^{\top}\mathbf{X} \approx n\mathrm{diag}(s_1, \dots s_p)$. Equip the coefficients with an unregularized horseshoe prior
\begin{equation*}
\label{eq:piironen_prior}
\begin{aligned}
    w_j \mid \tau, \lambda_j &\sim \mathcal{N}(0, \tau^2 \lambda_j^2) \qquad j=1, \dots p\\
    \lambda_j &\sim C^+(0, 1) 
\end{aligned}
\end{equation*}
where $\tau$ is some global hyperparameter. The prior on $\tau$ is not specified but is often also half-Cauchy, with the paper \citet{piironen2017sparsity} investigating the initial scale of the prior on $\tau$. From this, the posterior mean of the coefficients $\mathbf{w}$ given hyperparameters $\lambda_j$, data $(\mathbf{X}, \mathbf{y})$, for fixed $\tau, \sigma$ can be expressed as
\begin{align}
    \bar{w}_j = (1-\kappa_j)\hat{w}_j, \qquad
    \label{eq:kappa}
    \kappa_j = \frac{1}{1 + n\sigma^{-2}\tau^2s_j^2\lambda_j^2} = \frac{1}{1 + z_j^2\lambda_j^2} \ ,
\end{align}
where $\hat{w}_j$ is the ordinary least square (OLS) estimator and $\kappa_j$ is the shrinkage factor for $w_j$, with $z_j=\sqrt{n}\sigma^{-1}\tau s_j$. Intuitively, $\kappa_j=1$ means complete shrinkage of $w_j$ and $\kappa_j=0$ no shrinkage. These results hold in general for Gaussian scale mixtures for fixed $z_j$ \citep{piironen2017sparsity}. By now letting $\lambda_j$ follow an i.i.d. half Cauchy prior for all $j$, for fixed $\tau$ and $\sigma$, one can show that $\kappa_j$ follows the a priori distribution
\begin{align}
    \label{eq:kappa_prior}
    p(\kappa_j \mid \sigma, \tau) = \frac{1}{\pi} \frac{z_j}{(z_j^2-1)\kappa_j +1} \frac{1}{\sqrt{\kappa_j}\sqrt{1-\kappa_j}} \ ,
\end{align}
with 
\begin{align*}
    \mathbb{E}_{\lambda_j}[\kappa_j\mid\sigma, \tau] = \frac{1}{1+z_j}, \qquad \mathrm{Var}_{\lambda_j}(\kappa_j\mid\sigma, \tau) = \frac{z_j}{2(1+z_j)^2} \ .
\end{align*}
Another valuable property one can obtain from the shrinkage factor, is the effective number of nonzero coefficients. For a given $\tau$, \citet{piironen2017sparsity} define this as
\begin{equation}
\label{eq:piironen_m_eff}
    m_{\mathrm{eff}} = \sum_{j=1}^p(1-\kappa_j) \ ,
\end{equation}
and it effectively counts the number of times $\kappa_j$ is close to zero. This measure can be used as an indicator for effective model size. From the expectation and variance of the shrinkage factor, \citet{piironen2017sparsity} further develop the expectation and variance of $m_{\mathrm{eff}}$ and use these to choose the global prior scale $\tau_0$. In the regularized horseshoe, \citet{piironen2017sparsity} shows that one obtains the shrinkage coefficient 
\begin{equation}
    \tilde{\kappa}_j = (1-b_j)\kappa_j + b_j \qquad b_j = \frac{1}{1+n\sigma^2s_j^2c^2}
\end{equation}
where $\kappa_j$ is the original shrinkage coefficient. Thus, the theoretical results are modified by a shift from the interval $(0,1)$ to $(b_j,1)$, which is a result of truncating $\lambda_j$.

\subsection{DSM for linear regression}
We now develop an analogous shrinkage theory for the DSM priors. Consider again the regression in \Cref{eq:piironen_model}, but now assign $w_j$ the unregularized Dirichlet Horseshoe prior
\begin{align}
\label{eq:DSM_beta}
    w_j\mid\tau, \lambda_j, \xi_j \sim \mathcal{N}(0, \tau^2 \lambda_j^2 \xi_j) 
\end{align}
where $\boldsymbol{\xi} = (\xi_1,\dots,\xi_p) \sim \mathrm{Dir}(\alpha,\dots,\alpha) \in \Delta^{p-1}$ and again assume $\mathbf{X}^{\top}\mathbf{X}\approx n\mathrm{diag}(s_1, \dots, s_p)$.  
In the absence of a natural higher-level grouping in linear regression, we adopt the finest possible grouping by treating each coefficient as its own group and assigning an individual local scale $\lambda_j$. This choice preserves direct comparability with the horseshoe analysis of \citet{piironen2017sparsity} while isolating the effect of the Dirichlet
variance allocation. 

By noting that each component $\xi_{j}$ marginally follows a $\mathrm{Beta}(\alpha, (p-1)\alpha)$ distribution under symmetry, we can state that, given hyperparameters $\lambda_j, \xi_j$, data $(\mathbf{X}, \mathbf{y})$, for fixed $\tau, \sigma$, the marginal shrinkage factor takes the form
\begin{align}
\label{eq:kappa_dirichlet}
    \kappa_{ j} = \frac{1}{1+n\sigma^{-2}\tau^2s_j^2 \lambda^2_{j}\xi_{j}} = \frac{1}{1+z_j^2\lambda^2_{j}\xi_{j}} \ ,
\end{align}
with $z_j$ defined as before.
We now put forth a theorem that will characterize the marginal shrinkage imposed by the DSM priors. Hoere, we denote by ${}_2F_1$  the generalized hypergeometric function with $p=2$, $q=1$ \citep{andrews_askey_roy_1999}.

\begin{theorem}
Let $w_j$ follow the DSM prior with global scale $\tau$, group scale $\lambda_j \sim t^+_{\nu}(0, 1)$ and local scale $\xi_j \sim \mathrm{Beta}(\alpha, (p-1)\alpha)$ marginally. Assume a$z_j =\sqrt{n}\sigma^{-1}\tau s_j > 0$ to be fixed and given. The  marginal prior distribution of $\kappa_j$ as per \Cref{eq:kappa_dirichlet}, with the accompanying assumptions, can be written as
\begin{align*}
    p(\kappa_j \mid \tau, \sigma) &=\tilde{C}(\nu, z_j) \cdot \frac{(\alpha)_{\nu/2}}{(p\alpha)_{\nu/2}}\cdot \frac{\kappa_j^{\frac{\nu}{2}-1}}{(1-\kappa_j)^{\frac{\nu}{2}+1}}\cdot{}_2F_1\!\left(\begin{matrix} \frac{\nu+1}{2}, \alpha+\frac{\nu}{2} \\ p\alpha+\frac{\nu}{2}\end{matrix}; -\frac{\kappa_j\nu z_j^2}{1-\kappa_j} \right) 
\end{align*}
where $\tilde{C}(\nu, z_j) = \frac{\Gamma(\frac{\nu+1}{2})}{\sqrt{\nu\pi}\Gamma(\frac{\nu}{2})}  \nu^{\frac{\nu + 1}{2}} z_j^{\nu}$.

When $\nu=1$, we obtain 
\begin{align*}
    p(\kappa_j \mid \tau, \sigma) &= \frac{1}{\pi} \frac{z_j}{(1-\kappa_j)\sqrt{\kappa_j}\sqrt{1-\kappa_j}} \frac{(\alpha)_{1/2}}{(p\alpha)_{1/2}}{}_2F_1\!\left(\begin{matrix} 1, \alpha + \frac{1}{2} \\ p\alpha + \frac{1}{2} \end{matrix}; -\frac{\kappa_j z_j^2}{1-\kappa_j} \right) \ .
\end{align*}
The expectation is in that case given by
\begin{align*}
    \mathbb{E}_{\xi_j}[\kappa_{j} \mid \tau, \sigma] & = {}_2F_1\!\left(\begin{matrix} 1, \alpha \\ p\alpha \end{matrix}; z_j^2 \right) - z_j\frac{(\alpha)_{1/2}}{(p\alpha)_{1/2}}{}_2F_1\!\left(\begin{matrix} 1, \alpha+\frac{1}{2} \\ p\alpha + \frac{1}{2} \end{matrix}; z_j^2 \right) .
\end{align*}
\label{thm:theorem_2}
\end{theorem}
\noindent
A proof can be found in the supplementary material \ref{ap:proofs_and_such}.

The above theorem extends the well-known horseshoe shrinkage result by identifying the a priori marginal distribution of $\kappa_j$ under the DSM prior with Student’s t distribution for the local scales. 
\begin{figure}[h]
    \centering
    \includegraphics[width=0.8\textwidth]{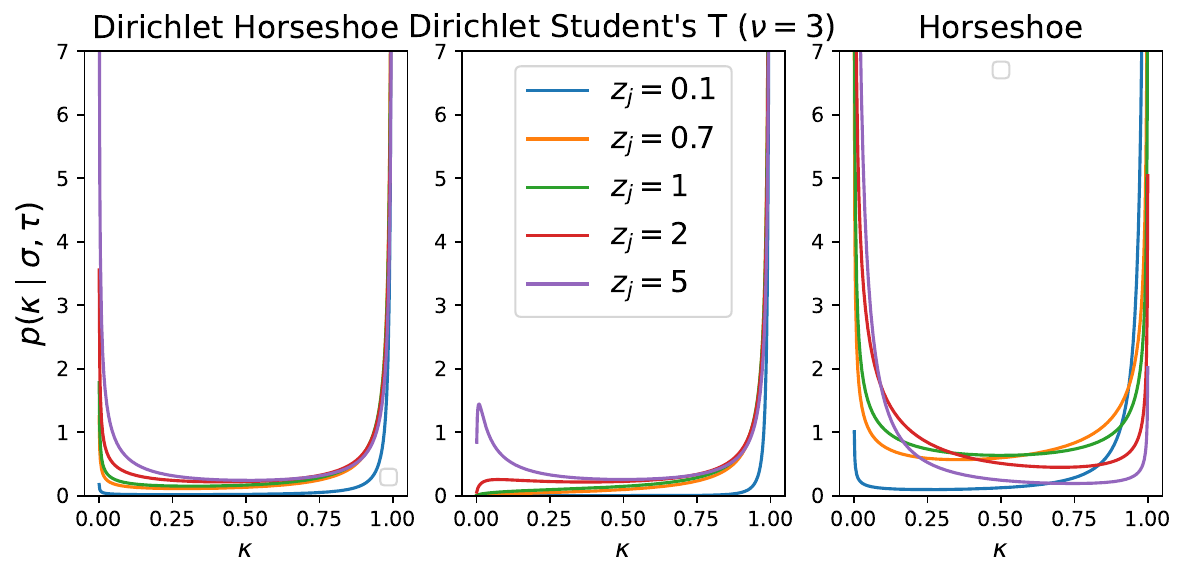}
    \caption{
    $p(\kappa \mid \sigma, \tau)$ in the Dirichlet horseshoe ($\nu=1$), Dirichlet Student’s t (\Cref{thm:theorem_2}), and the classical horseshoe (\Cref{eq:kappa_prior}). $\kappa=1$ indicates full shrinkage, and $\kappa=0$ indicates no shrinkage at all. It is clear that the Dirichlet methods shrinks more aggressively, as the shrinkage factor $\kappa$ has more mass close to $1$, than for the horseshoe.}
    \label{fig:p_kappa}
\end{figure}
   In \Cref{fig:p_kappa}, we see that the marginal prior on $\kappa$ induced from the Dirichlet horseshoe is essentially a scaled version of the original horseshoe, which is not surprising as the Dirichlet distribution is symmetric. The Dirichlet Student’s t prior on the other hand lacks the divergent mass at zero, and yields quite different prior shrinkage profiles compared to those from horseshoe distributed local scales.

 The results above characterize the marginal shrinkage induced by DSM priors in the classical linear regression setting under orthogonality $\mathbf{X}^\top \mathbf{X} \approx n \mathrm{diag}(s_1,\dots,s_p)$. In this regime, the effective shrinkage matrix is diagonal and each coefficient $w_j$ admits a scalar shrinkage factor $\kappa_j$, whose prior distribution we can describe in closed form for both Dirichlet–horseshoe and Dirichlet–Student’s t priors. In neural networks, the situation is more intricate. The likelihood depends on the weights through a complicated interaction structure. To relate the DSM shrinkage behaviour to this setting, we first linearize a single-hidden-layer BNN around a reference point and identify the analogue of the scalar shrinkage factor in terms of a matrix-valued operator acting on the weights.

\subsubsection{Linearizing the single-layer BNN}
To extend the scalar shrinkage analysis to Bayesian neural networks, we now place the DSM priors hierarchically on a single-hidden layer neural network ($L=2$) as previously described. Then, we linearize the network around a reference point $(\mathbf{w}_{1,0},\mathbf{b}_{1,0},\mathbf{w}_{L,0},b_{L,0})$. This yields a locally linear model in which shrinkage is naturally described by a matrix-valued operator rather than a scalar factor. Let $\mathbf{J}_w$ denote the Jacobian of the network output with respect to the vectorized input weights $\mathbf{w}_1$, and let $\Sigma_y$ denote the marginal covariance of the linearized likelihood.  Full expressions for $\mathbf{J}_w$, $\Sigma_y$, and the linearization are given in Section 3 of the Supplementary material \ref{ap:linearization_full}. Conditioned on the DSM hyperparameters $(\tau, \boldsymbol{\lambda}, \boldsymbol{\xi})$, we obtain the approximate linear Gaussian model
\begin{equation*}
y^*\mid \mathbf{w}_1 \sim \mathcal{N}(\mathbf{J}_w\mathbf{w}_1,\Sigma_y), \qquad \mathbf{w}_1 \sim \mathcal{N}(0,\tau^2\Psi),
\end{equation*}
where the prior covariance matrix $\Psi \in \mathbb{R}^{pH \times pH}$ is diagonal with entries
\begin{equation*}
\Psi_{(k,j),(k,j)} = \lambda_j^2 \, \xi_{kj}, \qquad k = 1,\dots,p,\;\; j = 1,\dots,H.
\end{equation*}
Standard Gaussian conditioning gives
\begin{equation*}
\mathbf{w}_1\mid y^* \sim  \mathcal{N}\!\left( (P+S)^{-1}S\hat{\mathbf{w}}, \; (P+S)^{-1} \right),
\end{equation*}
where
\begin{equation*}
P=\tau^{-2}\Psi^{-1}, \qquad  S=\mathbf{J}_w^\top\Sigma_y^{-1}\mathbf{J}_w, \qquad \hat{\mathbf{w}}=(\mathbf{J}_w^\top\Sigma_y^{-1}\mathbf{J}_w)^{-1}\mathbf{J}_w^\top\Sigma_y^{-1}y^*.
\end{equation*}
The matrix
\begin{equation}
\label{eq:shrinkage_matrix}
    K := (P+S)^{-1}S =I - (P+S)^{-1}P
\end{equation}
is a shrinkage matrix, generalizing the scalar shrinkage $\kappa$ from linear regression. It governs the shrinkage of the ordinary least squares estimator $\hat{\mathbf{w}}$ and forms the basis for our subsequent analysis of shrinkage in Bayesian neural networks.

\subsubsection{General shrinkage}
In the case of $S$ being diagonal, we recover the scalar shrinkage, $\kappa_{ij}$ for $w_{ij}$, as seen for the linear regression. In neural networks, however, the Jacobian structure generally makes $S$ non-diagonal, so shrinkage acts along data-adapted directions rather than coordinate-wise. To analyze this, we exploit the spectral structure of the shrinkage operator. 

As derived in Section 3 of the supplementary material \ref{ap:linearization}, the matrix admits the whitened form
\begin{align*}
    (P+S)^{-1}S = P^{-1/2}(I+G)^{-1}G P^{1/2}, 
    \qquad 
    G = P^{-1/2} S P^{-1/2} \ .
\end{align*}
Diagonalizing $G = U \Omega U^\top$ with $\Omega = \mathrm{diag}(\omega_1,\dots,\omega_{pH})$
yields shrinkage along the generalized eigenvectors satisfying
$S \mathbf{u}_j = \omega_j P \mathbf{u}_j$. In these directions, we obtain a shrinkage factor
analogous to equation (\ref{eq:kappa})
\begin{align}
    \kappa(\mathbf{u}_j) 
    = \frac{1}{1+\tau^2 \psi^2_{\mathrm{eff}}(\mathbf{u}_j)\, \mathbf{u}_j^{\top} S \mathbf{u}_j}, 
    \qquad 
    \psi_{\mathrm{eff}}^2(\mathbf{u}_j) 
    = \left( \mathbf{u}_j^{\top} \Psi^{-1} \mathbf{u}_j\right)^{-1} .
\end{align}
Thus, each generalized eigendirection $\mathbf{u}_j$ behaves like a scalar problem with an effective local scale $\psi_{\mathrm{eff}}(\mathbf{u}_j)$ and an effective data term $\mathbf{u}_j^\top S \mathbf{u}_j$. Furthermore, using the cyclic invariance of the trace, one finds
\begin{equation}
\begin{aligned}
\label{eq:m_eff}
    \mathrm{tr}\big((P+S)^{-1}S\big) 
    &= \mathrm{tr}\big((I+G)^{-1}G\big)  
    = \sum_{j=1}^{pH}\frac{\omega_j}{1+\omega_j} \ ,
\end{aligned}
\end{equation}
which provides a direct analogue to the effective model size $m_{\mathrm{eff}}$ in equation  (\ref{eq:piironen_m_eff}). Consequently, as this trace is aggregated over generalized directions rather than coordinates, it remains valid for non-diagonal $S$.

In contrast to the coordinate-wise shrinkage in a linear regression, the shrinkage in a BNN acts in the generalized eigen-directions $\mathbf{u}_j$ of the pair $(S,P)$. Because each hidden unit depends only on its own incoming weights, the Jacobian $\mathbf{J}_w$ has a unitwise structure when the weights are vectorized, and the matrix $S = \mathbf{J}_w^\top \mathbf{J}_w$ is therefore close to block-diagonal, with blocks corresponding to the sets of weights feeding into individual hidden nodes. Consequently, many generalized eigenvectors $\mathbf{u}_j$ are effectively localized within a single hidden unit. Within each block, the effective scale $\psi_{\mathrm{eff}}(\mathbf{u}_j)$ captures how the DSM prior aggregates node-level shrinkage through $\lambda_j$, with the Dirichlet weights $\xi_{ih}$ controlling relative contributions of individual inputs.  
The factor $\mathbf{u}_j^{\top} S \mathbf{u}_j$ reflects the data geometry and noise level, and under mild regularity assumptions (see Supplementary material, Section 3, for details), one can establish the bounds
\begin{align*}
1-\frac{1}{1+\psi_{\mathrm{eff}}^2(\mathbf{u}_j)\tau^{2}\tfrac{\Theta(n)}{\sigma^2+\Theta(n)}} \;\le\; 1-\frac{1}{1+\psi_{\mathrm{eff}}^2(\mathbf{u}_j)\tau^{2}\mathbf{u}_j^\top S \mathbf{u}_j} \;\le\; 1-\frac{1}{1+\psi_{\mathrm{eff}}^2(\mathbf{u}_j)\tau^{2}\sigma^{-2}\Theta(n)} \ ,
\end{align*}
where $\Theta(n)$ denotes a quantity that is bounded above and below by positive constants times $n$, and $\sigma^2$ is the observation noise variance in the likelihood. These bounds make explicit how the sample size $n$, noise level $\sigma^2$ and DSM scales jointly control the amount of shrinkage in each mode $u_j$.

\section{Experiments}
\label{sec:experiments}
The experimental analysis focuses on the baseline Gaussian prior, the horseshoe prior and two instances of DSM priors, namely the Dirichlet horseshoe prior (DSM-HS) and the Dirichlet Student’s t prior (DSM-ST), where the latter is specified with $\nu=3$ degrees of freedom. First, a constructed linear regression example is used to study shrinkage and effective dimensionality in a controlled sparse setting with correlated predictors. Second, the priors are used in BNNs on a simulated regression task, with known interactions and sparsity, providing insight into how the shrinkage behaviour observed in linear models carries over to neural networks. Lastly, BNNs are fitted to real-world datasets, which serve as benchmarks with less explicit structure and allow us to assess the practical utility of the priors. 

 We study how our prior construction encourages sparsity by investigating two distinct pruning schemes. In the first approach, which we refer to as \emph{prune per sample}, pruning is applied independently to each posterior draw of the network parameters. That is, each sampled network is pruned based on its own weight magnitudes, and predictions are obtained by averaging over the resulting pruned networks. This scheme preserves posterior variability in the sparsity pattern, but leads to sample-specific network structures. In the second approach, which we refer to as \emph{posterior prune}, pruning is performed at the level of the posterior distribution. Here, a single pruning mask is constructed from the posterior mean of the absolute weights and applied across all posterior samples. Predictions are then formed by averaging over these consistently pruned networks. This scheme yields a single, interpretable sparsity structure representative of the posterior. Since the networks considered in this study are relatively small, the attainable level of sparsity is inherently limited, and we expect larger architectures would permit substantially higher pruning rates. Moreover, posterior pruning is inherently more aggressive than prune-per-sample, as it enforces a global sparsity pattern across all posterior draws.
 
   Across all experiments, we use a single-hidden-layer feedforward Bayesian neural network with $H=16$ neurons. 
Posterior inference is performed using Hamiltonian Monte Carlo with the No-U-Turn Sampler (NUTS), drawing $M=1000$ samples per chain from $4$ independent chains, after a warm-up period of $M_{\mathrm{warmup}}=1000$ iterations per chain.  The same sampling configuration is used for both linear and neural-network models. To ensure comparable shrinkage behavior across models, we follow the recommendation of \citet{piironen2017sparsity} for setting the global scale parameter,
\begin{equation*}
    \tau_0 = \frac{p_0}{p - p_0}\,\frac{\sigma}{\sqrt{N}},
\end{equation*}
where $N$ denotes the sample size, $p$ the input dimensionality, $p_0$ an a priori guess of the number of relevant covariates, and $\sigma$ the noise scale. We fix $p_0 = 4$ throughout, encouraging moderate sparsity while remaining agnostic about the exact degree. The noise variance is assigned the prior $\sigma^2 \sim \mathrm{Inv\text{-}Gamma}(3,2)$, corresponding to $\mathbb{E}[\sigma^2]=\mathrm{Var}(\sigma^2)=1$. We view this as mildly informative, anchoring the variance at the unit scale after standardization and stabilizing the induced global shrinkage level $\tau_0$. In all experiments, the local scales in the DSM and Horseshoe priors are regularized as described in Section \ref{subsec:regularization}. All prior specifications are held fixed across linear and neural-network models.

\subsection{A linear regression example}
\begin{figure}[h]
    \centering
    \includegraphics[width=0.9\linewidth]{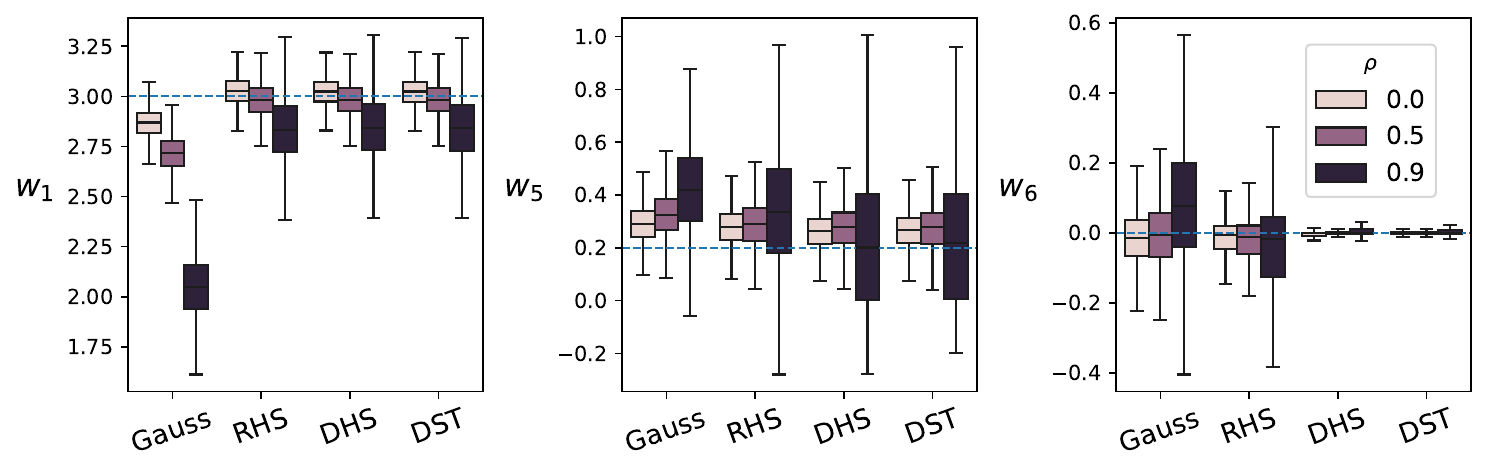}
    \caption{
    Boxplot of posterior samples from $(w_1, w_5, w_6)$ for the linear regression model, for different correlations. The dotted blue line represents the underlying, true coefficient.}
    \label{fig:linreg_betas}
\end{figure}

Based on the regression in (\ref{eq:piironen_model}), we construct a linear regression example. We set $N=250$, $p=10$ and generate data $\mathbf{X}\in \mathbb{R}^{N \times p}$, with $\mathbf{X}$ standard normally distributed with pairwise correlation $\rho$ between all covariates. The response is then generated as $\mathbf{y} = \mathbf{X}\mathbf{w} + \varepsilon$ where $\varepsilon \sim \mathcal{N}(0, 1)$ and $\mathbf{w} \in \mathbb{R}^{p}$ given by $w_1=3.0, w_2=-2.0, w_3=1.5, w_4=0.8, w_5=0.2$ and $w_6=\cdots=w_{10}=0$. Upon fitting the model, the coefficients are given the DSM prior as in (\ref{eq:DSM_beta}), and we fit the models using $80\%$ of the full dataset, for instances of $\rho \in \{0.0, 0.5, 0.9\}$. The remaining $20\%$ of the dataset is held out for validating the models. In Figure \ref{fig:linreg_betas}, a histogram of sampled coefficients are shown, with the true coefficient as the dotted line. In this sparse regime, the Gaussian prior clearly stands out with its poor estimates compared to the other priors. The regularized horseshoe (RHS) and the DSM priors (DHS and DST) show very similar performance across all correlations for coefficients that are truly nonzero. It can be noted that for the smallest coefficient, $w_5$, the DSM priors give a good estimate on average, but show a larger spread than the Gaussian. However, for $w_6$ which is truly zero, the DSM priors seem to shrink this much stricter than the RHS.

\begin{figure}[t]
    \centering
    \includegraphics[width=0.9\linewidth]{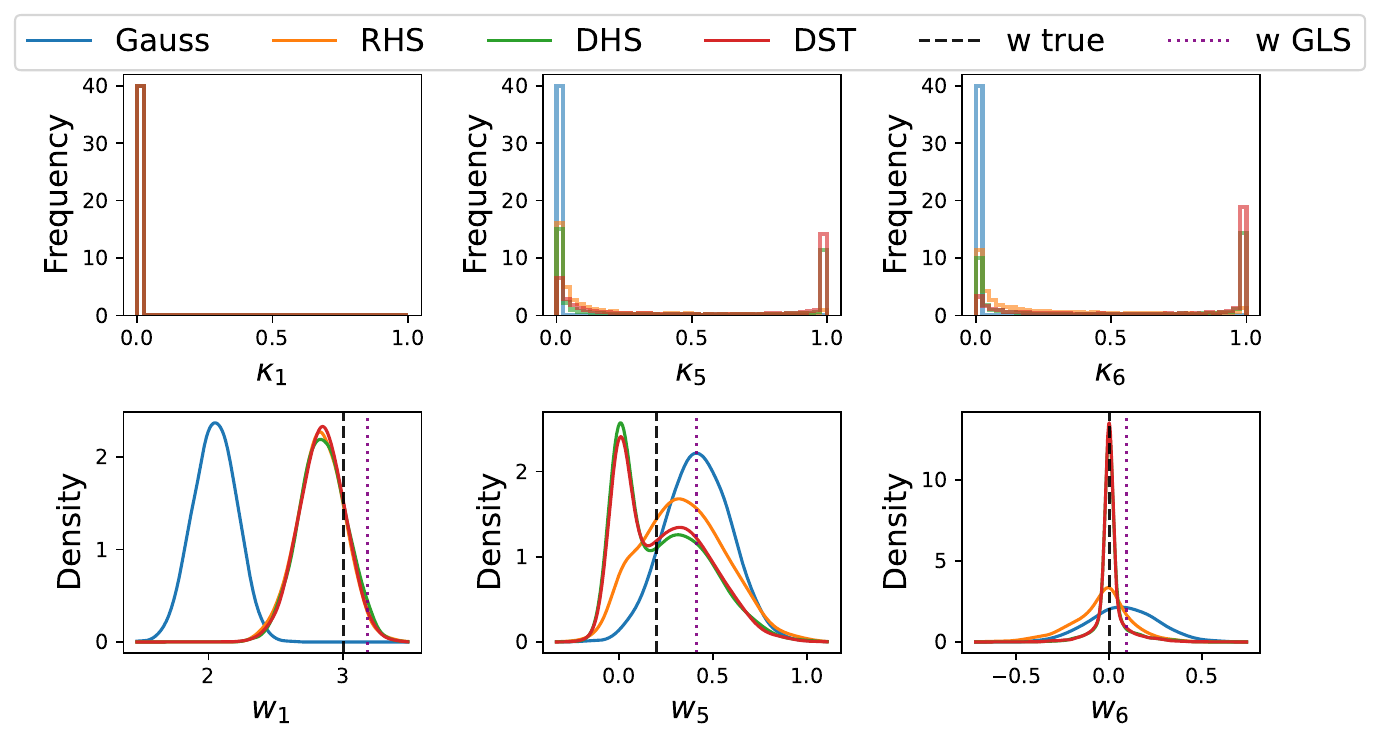}
    \caption{
    Estimated density of posterior samples for $w_1, w_5, w_6$ (left) and histogram of $\kappa_1, \kappa_5, \kappa_6$ (right) for the linear regression model, for $\rho=0.9$. The dashed black line represents the true coefficient, and the dotted purple line the GLS estimate.}
    \label{fig:linreg_beta_vs_kappa}
\end{figure}

This is further reflected in Figure \ref{fig:linreg_beta_vs_kappa}, where we display posterior samples of three coefficients $w_1, w_5$ and $w_6$, alongside the associated $\kappa$ values from (\ref{eq:kappa_dirichlet}). For $w_1$, all models seem to yield little to no shrinkage, which is of course expected, as this is the largest coefficient. It seems that the DSM priors shrink the smallest nonzero coefficient, $w_5$, more than the RHS does. For the zero coefficient $w_6$, the DSM priors exhibit superior shrinkage, giving a much more narrow estimate centered about zero.
The linear regression experiment highlights a key distinction between the DSM priors and the regularized horseshoe. While both priors perform similarly for large coefficients, the DSM priors impose stronger shrinkage on weak signals. In particular, the smallest nonzero coefficient is shrunk more under the DSM priors, whereas truly zero coefficients are more tightly concentrated around zero. This effect is not explained by a simple rescaling of the prior, as adjusting the global scale by $\sqrt{p}$ yields similar behavoir. Instead, the shrinkage pattern reflects the normalization induced by the Dirichlet distribution when $\alpha$ is small, which concentrates mass on a few active coefficients. Now, we turn to Bayesian neural networks.

\subsection{Friedman dataset, regression}
To evaluate the priors we consider the regression dataset proposed in \citet{friedman1991multivariate}, a popular benchmark for regression trees characterized by both interactions and sparsity \citep{Prado2021_Bayesian_additive_regression_trees}. 
The objective is to model the response $\mathbf{y} = f(\mathbf{x}) + \varepsilon$ with
\begin{gather}
\label{eq:friedman_data}
    f(\mathbf{x}) = 10\sin(\pi X_1 X_2) + 20\left(X_3 - \frac{1}{2}\right)^2 + 10X_4 + 5X_5 \ ,
\end{gather}
where the $p=10$ covariates $\mathbf{X}$ are generated uniformly on the hypercube $[0,1]^{10}$, with only the first five covariates contributing to the response, and $\varepsilon$ is a standard normal variable. In addition to the independent setting, we also consider a correlated regime. 
To construct correlated uniform covariates, we specify a target Spearman correlation $S_{ij}$ between covariates $i$ and $j$ and map it to a Gaussian copula using the relation $R_{ij} = 2\sin(\pi S_{ij}/6)$, where $R_{ij}$ denotes the corresponding Pearson correlation. Samples are then drawn from the Gaussian copula and transformed coordinatewise using the standard normal CDF, yielding uniformly distributed covariates with the desired dependence structure. This construction imposes a correlation, while preserving uniform marginals. For both the independent and correlated regimes, we generate fifteen datasets to fit the models, five for each sample size $N \in \{100, 200, 500\}$. The models are fit using a single-layer BNN with $H=16$ hidden units, $\tanh$ activation and evaluated on $N_{\mathrm{test}}=1000$ samples from the data generating process, with a previously unseen seed. As noted by Friedman (1991), the signal-to-noise ratio is high ($\mathrm{SNR}=4.8$).
 
\begin{figure}[tb]
\centering
\begin{minipage}{0.45\linewidth}
    \centering
    \includegraphics[width=\linewidth]{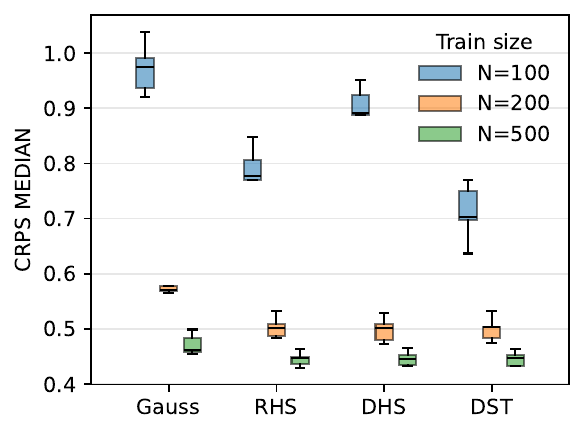}
\end{minipage}
\hfill
\begin{minipage}{0.45\linewidth}
    \centering
    \includegraphics[width=\linewidth]{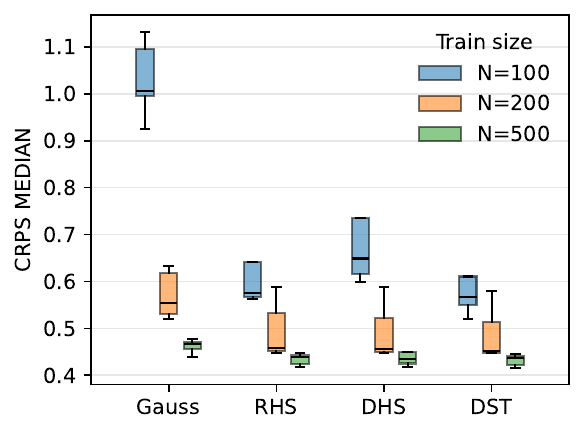}
\end{minipage}
\caption{
Boxplots of seed-level median CRPS for each model and training sample size on the independent Friedman (left) and correlated Friedman (right) datasets. For each training size $N \in \{100,200,500\}$, five independent datasets are used. Each box summarizes the five median CRPS values, where each value is computed from posterior predictive ensembles evaluated on a large generated test set.}
\label{fig:Friedman_crps}
\end{figure}

To evaluate the Friedman models, all performance metrics are computed separately for each random seed and subsequently aggregated over the five seeds corresponding to the same training sample size $N$. In Figure \ref{fig:Friedman_crps}, the continuous ranked probability score (CRPS) of the models is shown across training sample sizes and dependence regimes. The results are consistent with those reported in Table \ref{tab:friedman_rmse}, with the shrinkage priors generally yielding superior predictive performance compared to the Gaussian baseline. For $N=100$, the DST model attains the lowest predictive error in both the independent and correlated settings, while the DHS model performs less favorably in this small-sample regime. As the sample size increases, the differences between the sparsity-inducing priors narrow, with all three achieving similar performance for $N=200$ and $N=500$. In contrast, the Gaussian prior consistently results in higher higher error across all settings, indicating inferior predictive performance relative to the shrinkage-based alternatives. 
\begin{table}[tb]
\centering
\scalebox{0.8}{
\begin{tabular}{lccc|ccc}
\toprule
& \multicolumn{3}{c|}{Uncorrelated} 
& \multicolumn{3}{c}{Correlated} \\
\cmidrule(lr){2-4} \cmidrule(lr){5-7}
Model 
& N=100 & N=200 & N=500
& N=100 & N=200 & N=500 \\
\midrule
Gauss
& 2.601 (0.035) & 1.443 (0.042) & 1.150 (0.018)
& 2.547 (0.083) & 1.459 (0.060) & 1.143 (0.017) \\
RHS  
& 2.079 (0.033) & 1.260 (0.035) & 1.113 (0.018)
& 1.583 (0.107) & 1.233 (0.063) & 1.057 (0.011) \\
DHS   
& 2.359 (0.036) & \textbf{1.243} (0.037) & \textbf{1.106} (0.018)
& 1.846 (0.199) & 1.232 (0.056) & 1.057 (0.011) \\
DST   
& \textbf{1.887} (0.042) & 1.252 (0.036) & 1.107 (0.017)
& \textbf{1.515} (0.099) & \textbf{1.215} (0.060) & \textbf{1.049} (0.009) \\
\bottomrule
\end{tabular}
}
\caption{
Boxplots of posterior mean RMSE for each model and training sample size on the correlated Friedman data. For each $N \in \{100,200,500\}$, five independent datasets are used. For each dataset, predictions are formed by averaging over posterior draws and RMSE is computed on a large generated test set.
}
\label{tab:friedman_rmse}
\end{table}

\begin{figure}[tb]
    \centering
    \subfloat[Independent features\label{fig:m_eff_friedman_independent}]{
        \includegraphics[width=0.46\linewidth]{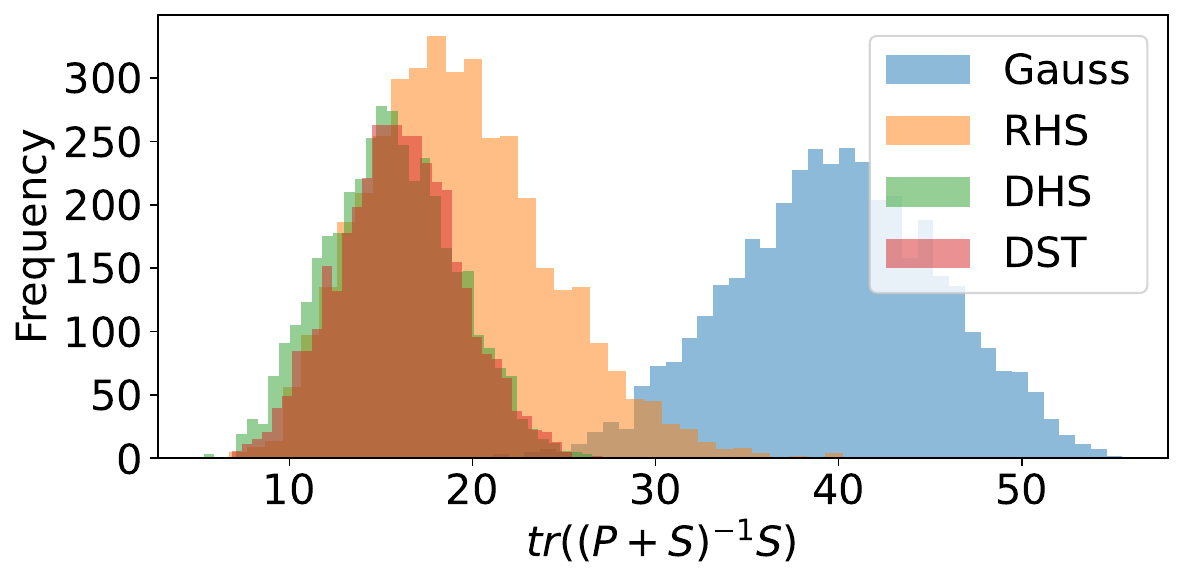}
    }
    \hfill
    \subfloat[Correlated features\label{fig:m_eff_friedman_correlated}]{
        \includegraphics[width=0.46\linewidth]{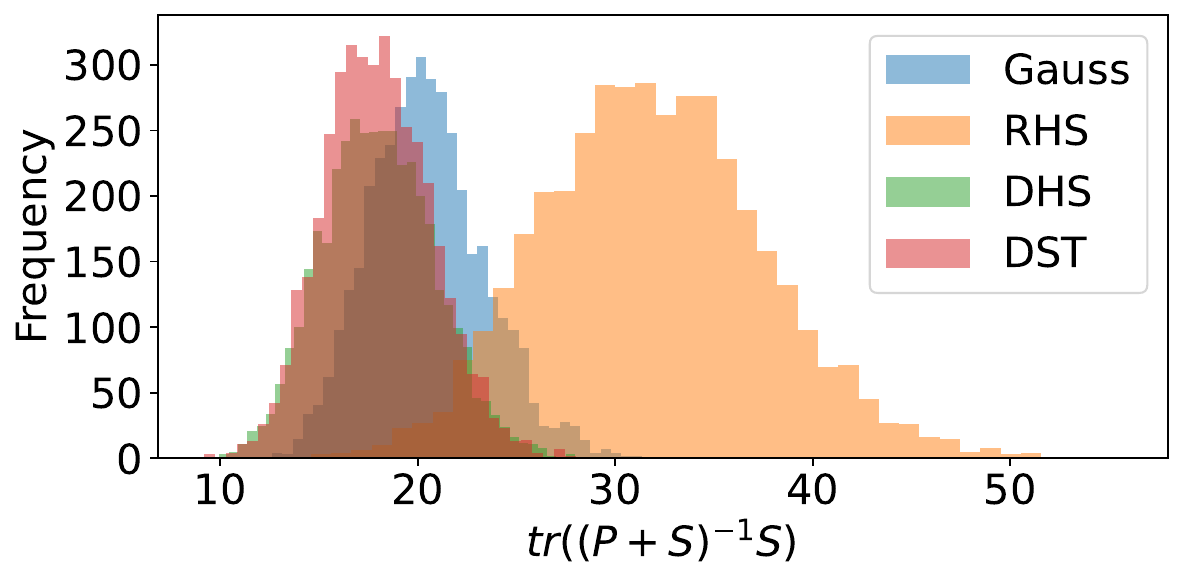}
    }
    \caption{
    Trace plots of the effective number of non-zero parameters  $m_{\mathrm{eff}} = \mathrm{tr}\!\big((P+S)^{-1} S\big)$ for different models on the Friedman dataset with independent and correlated input features. Each curve corresponds to a single fitted model and shows $m_{\mathrm{eff}}$ across 4000 posterior draws.
    }
    \label{fig:m_eff_friedman_side_by_side}
\end{figure}

To assess the complexity induced by the different priors, we focus on the case $N=100$, where prior effects are most pronounced. In Figure \ref{fig:m_eff_friedman_side_by_side}, we report the trace of the shrinkage matrix (\ref{eq:shrinkage_matrix}) across posterior samples. For independent covariates, the Gaussian prior exhibits substantially larger effective dimensionality than the shrinkage-based models. In the correlated setting, the regularized horseshoe yields the highest effective complexity among the sparsity-inducing priors. In both regimes, the DSM priors produce the lowest effective number of parameters, suggesting stronger overall shrinkage. This reduction in complexity is achieved without a corresponding loss in predictive accuracy. The same qualitative behavior is observed in the eigenvalue spectra of the shrinkage matrix.

\begin{figure}[tb]
\centering
\subfloat[Independent Friedman\label{fig:Friedman_sparsity}]{
  \includegraphics[width=0.48\linewidth]{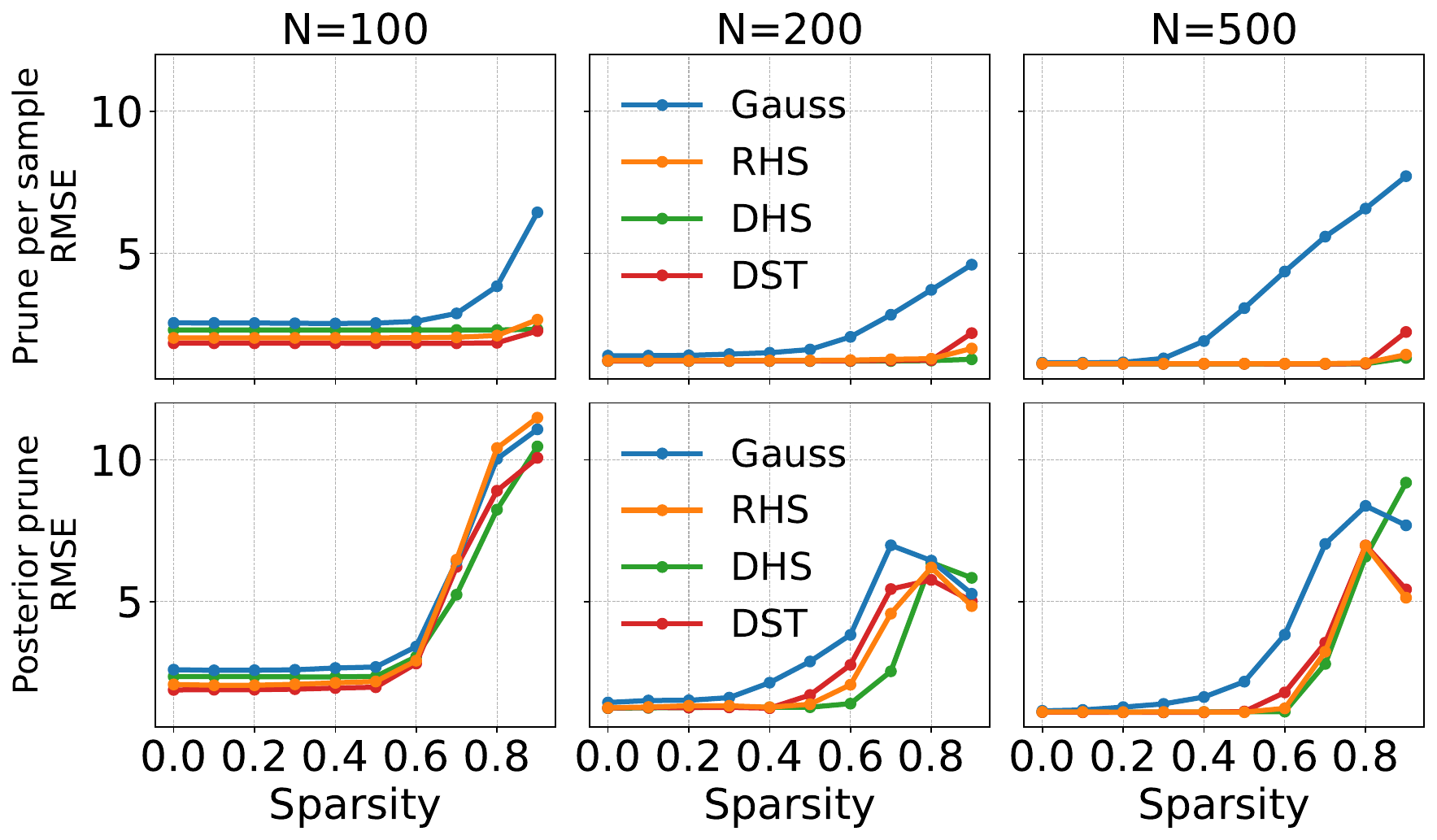}
}
\hfill
\subfloat[Correlated Friedman\label{fig:Friedman_correlated_sparsity}]{
  \includegraphics[width=0.48\linewidth]{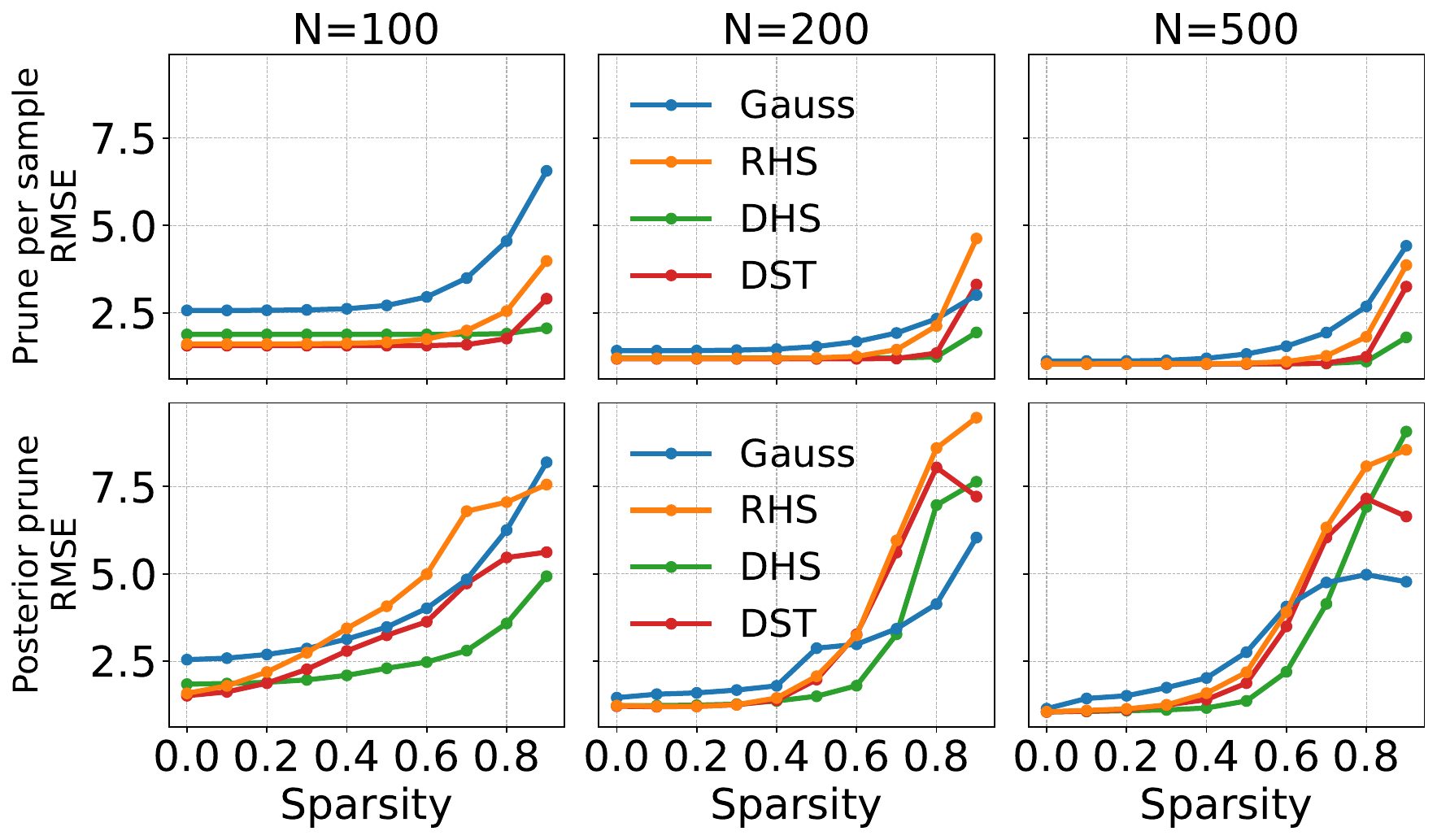}
}
\caption{Posterior mean RMSE as a function of sparsity level in Bayesian neural networks on Friedman datasets, aggregated across random seeds. The left panel shows the independent case and the right panel shows the correlated case. The upper panels correspond to the prune-per-sample scheme, the lower panels show posterior pruning.}

\end{figure}

We next examine how the different priors respond to explicit sparsification through pruning. For the independent Friedman dataset (Figure \ref{fig:Friedman_sparsity}), across all sparsity levels, the Gaussian model deteriorates markedly faster than the shrinkage-based priors. The RHS, DHS, and DST models display similar robustness in this setting, with no clear separation between them. For the correlated Friedman dataset (Figure \ref{fig:Friedman_correlated_sparsity}), the differences between sparsity-inducing priors become more pronounced. While the Gaussian prior again shows rapid performance degradation, the RHS model also exhibits reduced robustness to pruning. Among the DSM priors, the DHS model maintains lower RMSE across a wider range of sparsity levels, whereas the DST model shows intermediate behavior, performing better than the RHS but worse than the DHS. Overall, these results suggest that the DSM priors provide a favorable balance between predictive accuracy, model complexity, and robustness to aggressive sparsification, particularly in the presence of correlated covariates.

\subsection{Abalone dataset, regression}
A classic UCI regression dataset is the Abalone dataset \citep{abalone}, containing data from physical measurements on abalone shells. The categorical sex variable is encoded as an ordinal numerical covariate. Furthermore, the target of the regression is the number of rings the shell has, which determines the age of the abalone. The dataset consists of $N=4177$ observations with $p=8$ features, many of which exhibit strong positive correlations. We use an $80/20$ train–test split. For the Abalone dataset, we include the predictive negative log-likelihood as a performance measure.
 
\begin{figure}[tb]
\centering
\subfloat[\label{fig:abalone_crps}]{
  \includegraphics[width=0.36\linewidth]{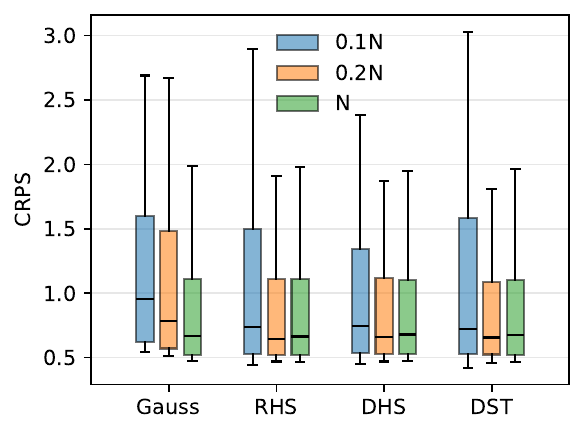}
}
\hfill
\subfloat[\label{fig:abalone_sparsity}]{
  \includegraphics[width=0.60\linewidth]{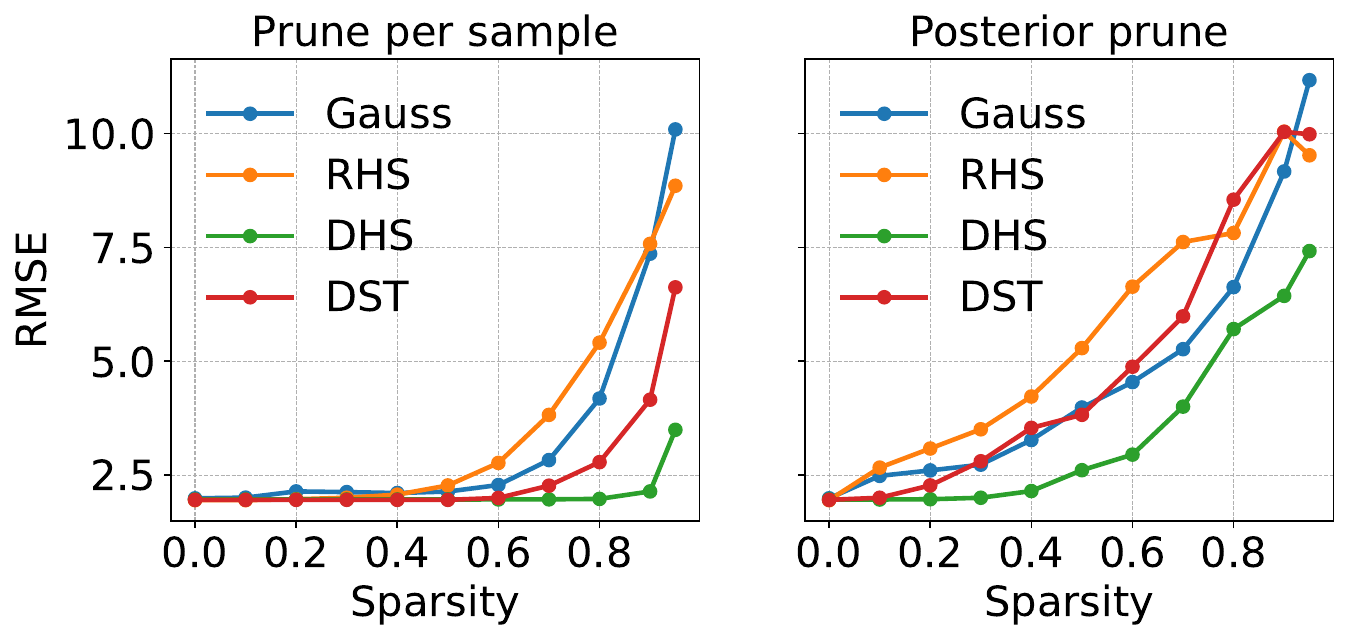}
}
\caption{a) Pointwise CRPS distributions for each model on the Abalone dataset at three training sizes. Each box summarizes CRPS across test points computed from posterior predictive ensembles. b) Posterior mean RMSE as a function of sparsity level for the Abalone dataset. The left column shows sample-wise pruning, where masks are recomputed for each posterior draw, while the right column shows posterior pruning using a single global mask per model.}
\end{figure}

In Figure \ref{fig:abalone_crps} and Table \ref{tab:abalone_performance}, predictive performance across priors is broadly comparable, with only moderate differences observed across training sizes. For the smallest training fraction ($0.1N$), the DHS attains the lowest predictive error, followed by the RHS, while the Gaussian prior again performs worse across all metrics. Notably, the DST model exhibits a larger CRPS spread in this small sample regime, indicating greater predictive uncertainty. At $0.2N$, all sparsity-inducing priors achieve nearly identical performance, with only marginal differences between RHS, DHS, and DST. When trained on the full dataset, performance converges further, with all models yielding similar RMSE and PNLL values, and overlapping CRPS distributions. Overall, these results suggest that on this relatively large dataset, predictive accuracy and uncertainty metrics provide limited separation between priors except for the clear gap to the Gaussian baseline.

\FloatBarrier
\begin{table}[tb]
\centering
\scalebox{0.8}{
\begin{tabular}{lcc|cc|cc}
\toprule
& \multicolumn{2}{c|}{$0.1N$} 
& \multicolumn{2}{c|}{$0.2N$} 
& \multicolumn{2}{c}{$N$} \\
\cmidrule(lr){2-3} \cmidrule(lr){4-5} \cmidrule(lr){6-7}
Model 
& RMSE & PNLL 
& RMSE & PNLL
& RMSE & PNLL \\
\midrule
Gauss & 2.918 & 2.534 & 2.662 & 2.433 & 1.990 & 2.103 \\
RHS   & 2.578 & 2.434 & 2.243 & 2.232 & \textbf{1.949} & \textbf{2.087} \\
DHS   & \textbf{2.475} & \textbf{2.365} & 2.240 & \textbf{2.229} & 1.965 & 2.096 \\
DST   & 2.702 & 2.497 & \textbf{2.236} & \textbf{2.229} & 1.956 & \textbf{2.087} \\
\bottomrule
\end{tabular}
}
\caption{
Posterior mean RMSE and test-set negative log-likelihood on the Abalone dataset for three training sizes (10\%, 20\%, and full data).
}
\label{tab:abalone_performance}
\end{table}

Clearer differences emerge when examining robustness to sparsification. As shown in Figure \ref{fig:abalone_sparsity}, the DSM models remain stable under substantial pruning. Under the prune-per-sample scheme, the DHS and DST models maintains near constant RMSE until approximately $90\%$ and $80\%$ sparsity, respectively, after which performance degrades. In contrast, the Gaussian and RHS models exhibit a noticeable increase in RMSE at substantially lower sparsity levels, with degradation beginning around $40$–$50\%$ sparsity. This separation is even more pronounced under posterior pruning, where the RHS model deteriorates rapidly, while the DHS prior preserves predictive accuracy over a much wider sparsity range. The pruning behavior of the RHS model in this very correlated setting is consistent with its behavior in the correlated Friedman experiments.
 
\begin{figure}[tb]
\centering
\begin{minipage}[t]{0.55\linewidth}
  \centering
  \includegraphics[width=\linewidth]{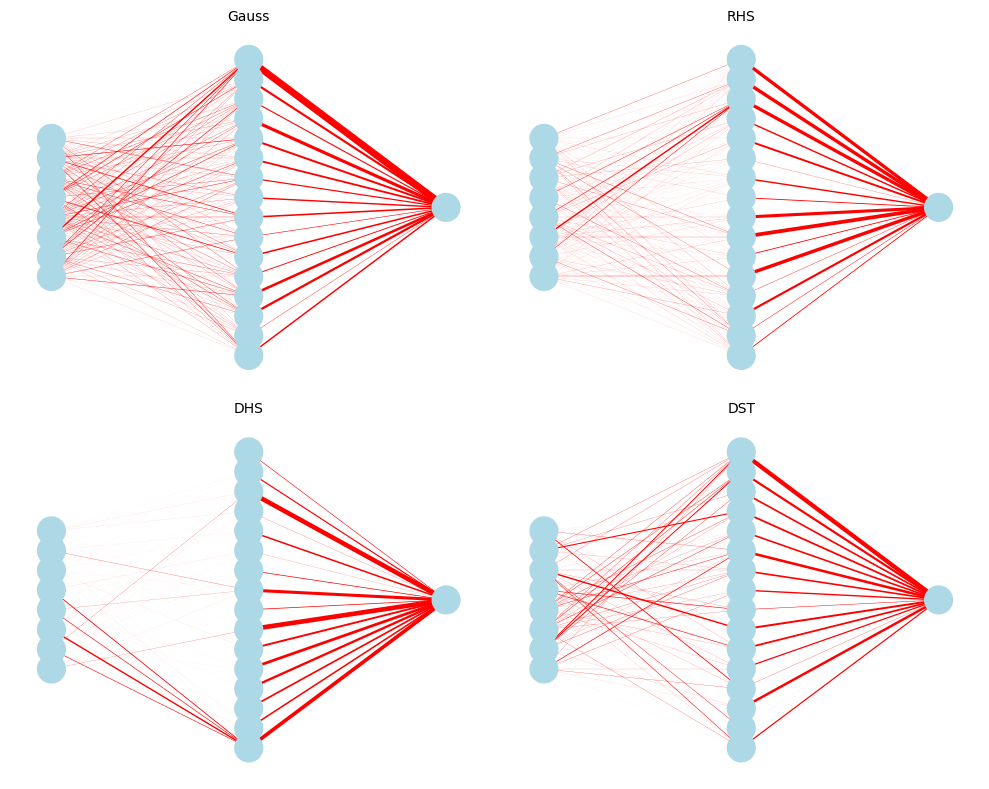}\\
  \vspace{2pt}
\end{minipage}
\hfill
\begin{minipage}[t]{0.4\linewidth}
  \centering
  \vspace{-13em}

  \footnotesize
  \setlength{\tabcolsep}{4pt}
  \renewcommand{\arraystretch}{0.9}
  \begin{tabular}{lcccc}
  \toprule
  Feature & Gauss & RHS & DHS & DST \\
  \midrule
  Whole weight   & 3.48 & 2.26 & 2.23 & 2.14 \\
  Shucked weight & 2.69 & 1.37 & 1.68 & 1.36 \\
  Shell weight   & 0.98 & 0.96 & 0.95 & 1.00 \\
  Viscera weight & 0.59 & 0.58 & 0.53 & 0.60 \\
  Diameter       & 0.29 & 0.30 & 0.29 & 0.03 \\
  Height         & 0.38 & 0.19 & 0.21 & 0.24 \\
  Length         & 0.49 & 0.46 & 0.17 & 0.12 \\
  Sex            & 0.18 & 0.17 & 0.12 & 0.14 \\
  \bottomrule
  \end{tabular}\\
  \vspace{2pt}
\end{minipage}
\caption{The left figure shows the posterior mean network, in which the thickness of edges are proportional to the mean absolute value of weights. The right table displays feature importances from a SHAP analysis for the Abalone dataset.}
\label{fig:abalone_network_and_shap}
\end{figure}

We also conducted a SHAP analysis using the KernelExplainer framework \citep{Lundberg_shap_neurips, shap_kernelexplainer}. SHAP values are based on Shapley values from cooperative game theory and measure the marginal contribution of each feature to the model prediction. The resulting summaries are shown together with a visualization of the posterior mean network in Figure \ref{fig:abalone_network_and_shap}. The DHS prior induces substantially sparser input-to-hidden connectivity, with a larger proportion of weights shrunk effectively to zero compared to the Gaussian, RHS and DST models. The concentration of mass on fewer connections might be the reason the DHS model is particularly robust to pruning.
Finally, the sparsity patterns induced by the Dirichlet Horseshoe model enables a degree of feature-level interpretability. The largest posterior weight magnitudes for the DHS model is consistently associated to shucked weight (node ${6}$), with height, whole weight, and viscera weight (nodes ${4,5,7}$) also receiving substantial emphasis. This aligns with the SHAP values, which similarly indicate these variables as most influential.

\subsection{Breast cancer dataset, classification}
Another commonly used UCI dataset is the breast cancer dataset \citep{breast_cancer_wisconsin}. The data has $p=30$ features on $N=569$ observations computed from a digitized image of breast mass, that describe the characteristics of the cell nuclei. The response indicates whether the tumor is malignant or benign, i.e. we perform a binary classification. As before, the models are trained on $80\%$ of the full dataset, and $20\%$, 114 observations, is held out for testing. We calculate the accuray (Acc), negative log-likelihood (NLL) and expected calibration error (ECE) for the models in Table \ref{tab:bc_performance}. It is a relatively easy classification task, and we see that all models achieve high accuracy and performance overall is similar.
\begin{table}[tb]
\centering
\scalebox{0.8}{
\begin{tabular}{l|c|c|c}
\toprule
\cmidrule(lr){2-4}
Model & Acc & NLL & ECE  \\
\midrule
Gauss & 0.9386 & 0.1288 & \textbf{0.0267} \\
RHS   & \textbf{0.9649} & 0.1005 & 0.0320 \\
DHS   & \textbf{0.9649} & \textbf{0.0943} & 0.0305 \\
DST   & \textbf{0.9649} & 0.1004 & 0.0291 \\
\bottomrule
\end{tabular}
}
\caption{Posterior mean performance of networks on the Breastcancer dataset.}
\label{tab:bc_performance}
\end{table}

To analyze how robust the posterior BNNs we obtain are, we rely on the methods presented in \citet{cardelli_statistical_2019}. In this, a notion of robustness and of safety are presented, and we briefly describe them here. 
\begin{definition}
\textbf{Robustness: }Consider a neural network $f^{\mathbf{w}}$ with training set $\mathcal{D}$. Let $x^*$ be a test point and $T \subseteq \mathbb{R}^p$ a bounded set. For a given $\delta  >  0$, define $p_1$ as the probability
\begin{equation}
    p_1 = \mathbb{P}(\{\exists x \in T : \lVert\sigma(f^{\mathbf{w}}(x^*)) - \sigma(f^{\mathbf{w}}(x)) \rVert_q \geq \delta \} |\mathcal{D}) \ ,
\end{equation}
where $\sigma$ is the softmax output of the classifier and $\lVert \cdot \rVert_q$ is a given norm or seminorm (we apply the $\ell_2$-norm). For $0\leq \eta \leq 1$, we say that $f^{\mathbf{w}}$ is robust with probability at least $1-\eta$ in $x^*$ with respect to $T$ and perturbation $\delta$ \textit{iff} $p_1 \leq \eta$. 
\end{definition}
This means that $p_1$ represents the probability that there exists $x \in T$ such that the output of the softmax layer deviates from $x$ more than a given threshold $\delta$. $x^*$ is not necessarily an element of $T$, but if it is, $p_1$ assess the robustness to local perturbations. $p_1$ is relative to the output of the softmax layer, and thus the stochasticity of it comes only from the distribution over weights.  
\begin{definition} 
\textbf{Safety: }Let $f^{\mathbf{w}}$ be a neural network, $\mathbf{m}(x) = \arg \max_j \sigma_j(f^{\mathbf{w}}(x))$ denote the predicted class label and $\mathcal{D}$ denote the training data. Then define
\begin{equation}
    p_2 = \mathbb{P}(\{\exists x \in T: \mathbf{m}(x^*) \neq \mathbf{m}(x) \} |\mathcal{D}) \ .
\end{equation}
For $0 \leq \eta \leq 1$, the model $f^{\mathbf{w}}$ is said to be \textit{safe} with probability at least $1-\eta$ in $x^*$ with respect to $T$ \textit{iff} $p_2 \leq \eta$. 
\end{definition}
The stochasticity of $p_2$ includes both the distribution over weights and the noise of the modeled process. This means that for regions of the input space where the model is more uncertain of what class to assign the output, $p_2$ will take on a higher value. \\ \\ 
To obtain the estimates of $p_1$ and $p_2$ we set up an FGSM (Fast Gradient Sign Method) adversarial attack scheme; see \citet{yuan2019adversarial} for more details. For each model we sample a random subset of the test set, and generate adversarial examples by adding small perturbations to the input in the direction of the gradient of cross entropy loss w.r.t. the input. The perturbation is bounded by the $\ell_{\infty}$-ball of radius $\varepsilon$ and the threshold $\delta$ is chosen in fractions of $\varepsilon$. The FGSM attack is applied to each of the $M=100$ posterior network samples, and the resulting adversarial outcomes are aggregated across samples to estimate both $p_1$ and $p_2$. While $p_1$ varies smoothly as a confidence-based measure, $p_2$ reflects whether adversarial perturbations induce label changes and is therefore typically $0$ or $1$ at the run level, with intermediate values appearing only when averaged. 
Consequently, we report the results for $p_2$ binned into safe if $p_2=0$, partially safe if $p_2 \in (0, 1)$, and unsafe if $p_2 =1$, rather than treating it as a continuous quantity 
 
\begin{figure}[tb]
\centering
\begin{minipage}[t]{0.55\linewidth}
  \centering
  \includegraphics[width=0.8\linewidth]{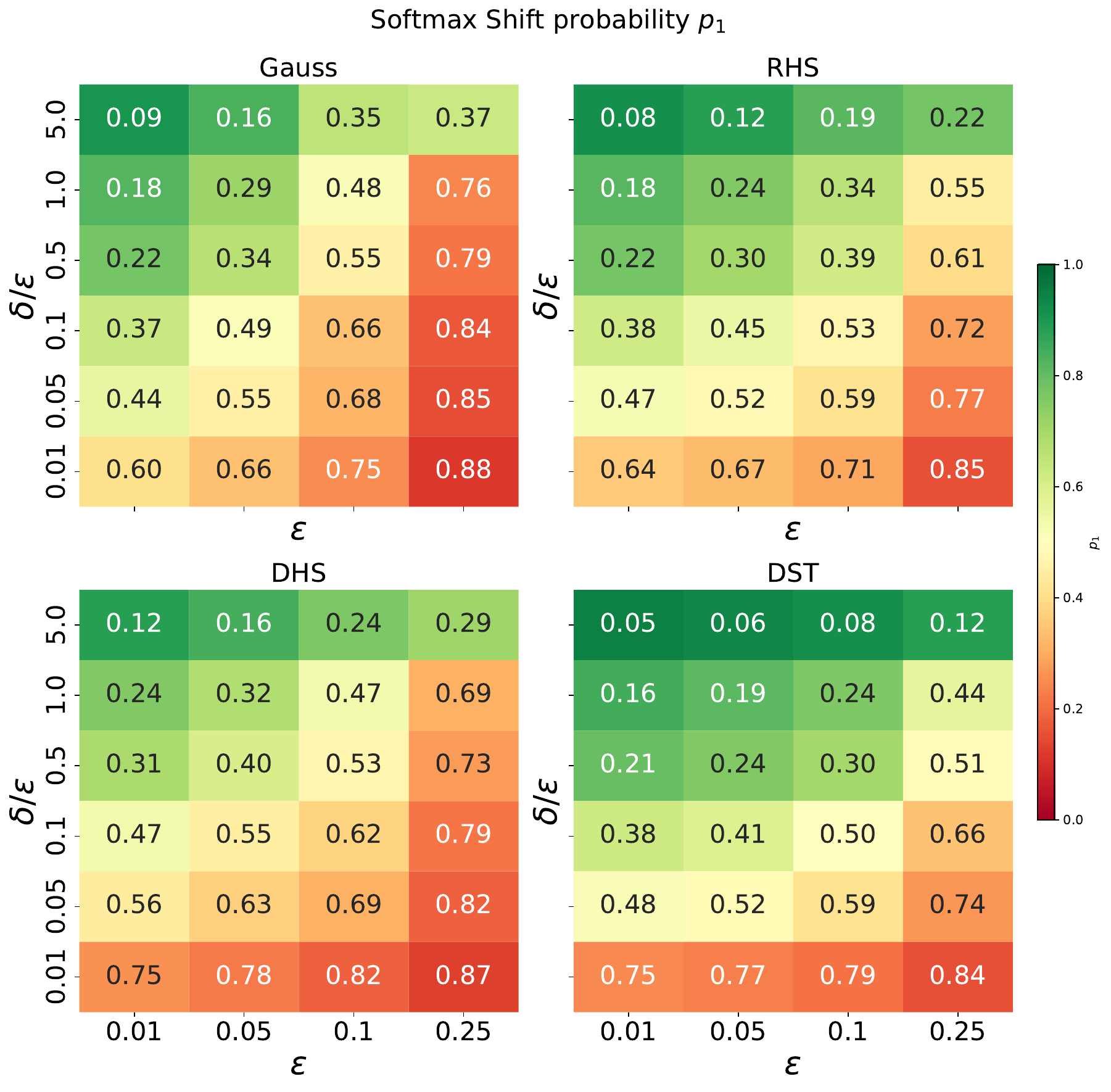}\\
  \vspace{2pt}
\end{minipage}
\hfill
\begin{minipage}[t]{0.4\linewidth}
  \centering
  \vspace{-11em}

  \footnotesize
  \setlength{\tabcolsep}{4pt}
  \renewcommand{\arraystretch}{0.9}
    \begin{tabular}{lccc}
    \toprule
    Model 
    & S & P & U \\
    \midrule
    Gaussian              
    & 0.583 & 0.270 & 0.147 \\
    RHS 
    & 0.583 & 0.395 & \textbf{0.022} \\
    DHS   
    & 0.627 & 0.340 & 0.033 \\
    DST
    & \textbf{0.761} & 0.211 & 0.029 \\
    \bottomrule
    \end{tabular}\\
  \vspace{2pt}
\end{minipage}
\caption{Left: Robustness of the softmax probabilities in the classification on the breastcancer dataset. The x-axis displays the disturbance $\varepsilon$, and the y-axis displays the fraction $\delta/\varepsilon$. Right: Fractions of safe (S), partially safe (P), and unsafe (U) outcomes from the $M=4000$ samples, defined in terms of the probability of adversarial label change $p_2$, on the breast cancer dataset.}
\label{fig:breastcancer}
\end{figure}

From Figure \ref{fig:breastcancer} we observe that the DST model requires substantially larger perturbations to induce changes in the softmax outputs, indicating stronger local stability. This is consistent with Table \ref{fig:breastcancer}, where DST achieves the highest fraction of safe outcomes and one of the lowest fractions of unsafe outcomes. The RHS model also performs well in terms of robustness, exhibiting relatively good local stability and the smallest fraction of unsafe outcomes overall. The DHS model achieves a high proportion of safe outcomes, but seems to be more sensitive to local perturbations in terms of $p_1$. In contrast, the Gaussian model shows weaker robustness, with the lowest fraction of safe outcomes and the highest proportion of unsafe cases.

\section{Discussion}
\label{sec:discussion}
This work has presented a new class of sparsity-inducing priors, the Dirichlet Scale Mixture (DSM) priors, for Bayesian neural networks. By utilizing the natural hierarchy of neural networks, shrinkage is imposed globally, at node level and at weight level. Each node is assigned with an a priori variance budget, which is distributed to the incoming weights using the Dirichlet distribution. Thus, the priors encourage node level sparsity, as well as competition among the weights mapping into the same node. This leads to a more interpretable network model, that can be pruned extensively while maintaining predictive performance.
 
   We first analyzed the dependence structure induced among the variance components, showing that the sign of the covariance between variance terms is governed by dimensionality, geometry, and the tail behavior of the group-specific prior. Secondly, the marginal shrinkage imposed was theoretically developed for linear regression models, before lifting the concepts to a single hidden layer neural network. Then, a comprehensive empirical analysis of how the models perform in practice was carried out. A linear regression example served as preliminary results, before BNNs were equipped with DSM priors. The experiments with the neural networks on both simulated and real datasets show that the DSM priors have attractive properties. They express  structured representations, allowing the network to be sparsified and sparsified to a greater extent than competing priors. Compared to Gaussian and regularized horseshoe priors, the DSM priors achieve comparable or better predictive performance with far fewer effective parameters, they show superior performance at high sparsity levels, and they remain relatively stable under adversarial perturbations such as FGSM attacks. Importantly, their performance is particularly strong in small to moderate data regimes, a setting where Bayesian machine learning is especially beneficial due to its ability to provide calibrated uncertainty estimates and incorporate domain knowledge. By concentrating prior probability mass on a small subset of the weights, the DSM priors enable networks that are somewhat interpretable and offer a form of implicit feature selection in the input to hidden layer connections.
 
   The DSM priors are marginally equivalent to placing independent Beta distributions on the individual weights. This begs the question of how such a Beta prior would perform compared to the Dirichlet-based models. Our investigations found that the predictive performances are almost identical, and that the main differences can be seen in the number of effective parameters and when pruning the models. The Dirichlet models use less parameters, and can be pruned to the same, or to a larger, degree. Details regarding this investigation can be found in Section 4 of the supplementary material \ref{ap:supplementary}.
 
   Convergence of MCMC samplers in Bayesian neural networks is well known to be challenging due to multimodality and parameter non-identifiability. In Section 4 of the supplementary material, we present and discuss convergence diagnostics for the network models considered in this work.

   Many aspects of the DSM priors have not yet been investigated, but this preliminary work opens up a lot of possible pathways for further development. One aspect of particular interest is to study the behavior of heavy-tailed priors as the network size increases, especially in highly overparameterized regimes that are increasingly common in modern neural network practice. Exploring such settings may require inference techniques beyond standard HMC or substantially increased computational resources, but recent methodological advances make this direction both feasible and relevant. Another possibility, which could be pursued independently or in combination with the above, is to explore richer dependence structures within the prior. Imposing stronger structural constraints may lead to more efficient representations, potentially reducing computational cost while also yielding a clearer picture of how different parts of the network contribute to the final prediction.

\section*{Funding}
This work was supported by the Research Council of Norway through its Centre of Excellence Integreat -- The Norwegian Centre for Knowledge-driven Machine Learning, project number 332645.

\newpage

\bibliographystyle{plainnat}
\bibliography{bibliography}

\newpage

\vskip 0.2in
\newpage
\clearpage
\section{Supplementary Material}
\label{supp}
\addcontentsline{toc}{section}{Supplementary Material}

The supplementary material is included below, with five sections:
\begin{itemize}
    \item Supplementary A: Dependence structure derivations.
    \item Supplementary B: Proofs of theoretical results.
    \item Supplementary C: Details on the linearization of the BNN.
    \item Supplementary D: Additional results and experiments, experimental details and a convergence assessment.
    \item Supplementary E: Stan code example of a DSM prior, and link to full code on GitHub.
\end{itemize}

\subsection*{Supplementary A: Dependence structure derivations}
\label{ap:dependence}
We here give derivations of the dependence structure of the DSM priors.
\subsubsection*{Covariance of variance terms.}
Let $i \neq j$ and consider
\begin{equation*}
\begin{aligned}
    X_{jk} &= \lambda_j \xi_{jk},
    \qquad
    X_{jl} = \lambda_j \xi_{jl}, \qquad k \neq l \\
    \xi_{j} &= (\xi_{j1}, ..., \xi_{jp}) \sim \mathrm{Dirichlet}(\alpha,\dots,\alpha), \quad \lambda_j \perp \xi_j,
\end{aligned}
\end{equation*}
with $\lambda_j>0$. By independence, we have
\begin{align*}
    \mathrm{Cov}(X_{jk},X_{jl})
    &= \mathbb{E}[\lambda_j^2]\mathbb{E}[\xi_{jk}\xi_{jl}] - \mathbb{E}[\lambda_j]^2\mathbb{E}[\xi_{jk}]\mathbb{E}[\xi_{jl}]  \\
    &= \mathbb{E}[\lambda_j^2]\left(\mathrm{Cov}(\xi_{jk}, \xi_{jl}) + \mathbb{E}[\xi_{jk}]\mathbb{E}[\xi_{jl}] \right) - \frac{\mathbb{E}[\lambda_j]^2}{p^2}  \\
    &= \mathbb{E}[\lambda_j^2]\left(-\frac{1}{p^2(p\alpha+1)} + \frac{1}{p^2}\right) - \frac{\mathbb{E}[\lambda_j]^2}{p^2}  \\
    &= \frac{p\alpha}{p^2(p\alpha+1)}(\mathrm{Var}(\lambda_j) + \mathbb{E}[\lambda_j]^2) - \frac{\mathbb{E}[\lambda_j]^2}{p^2}  \\
    & = \frac{1}{p^2(p\alpha + 1)} \left(p\alpha\mathrm{Var}(\lambda_j) - \mathbb{E}[\lambda_j]^2 \right)
\end{align*}
Since no properties of the distribution of $\lambda_j$ is used in the proof, the expressions above are also valid if $\lambda_j$ is replaced by the regularized version $\tilde\lambda_j$.

\subsection*{Supplementary B: Lemmas and proofs of theorems}
\label{ap:proofs_and_such}
In this section, we prove the theorems stated in the main text, which includes defining and proving three lemmas and use of some auxillary Pochhammer identities.
\label{ap:proofs}
\begin{lemma}[Expectation of transformed Beta variable I]
Consider the Dirichlet component with marginal $\xi_{j} \sim \mathrm{Beta}(\alpha, (p-1)\alpha)$ and the transformation $\xi_j \mapsto \frac{\xi_j^k}{(1+s\xi)^a}$, where $s  >  -1$ is a constant independent of $\xi_j$ and $a\in \mathbb{C}$. The expectation of this transform is then
\begin{align*}
    \mathbb{E}_{\xi_j}\left[ \frac{\xi_j^k}{(1+s\xi_j)^a } \right] & = \frac{(\alpha)_k}{(p\alpha)_k}\sum_{n=0}^{\infty} \frac{(a)_n(\alpha+k)_n}{(p\alpha+k)_n} \frac{(-s)^n}{n!} = \frac{(\alpha)_k}{(p\alpha)_k}{}_2F_1\!\left(\begin{matrix} a, \alpha+k \\ p\alpha + k \end{matrix}; -s \right)
\end{align*}
where $(x)_n=\frac{\Gamma(x+n)}{\Gamma(x)}$ denotes the Pochhammer symbol \citep{AbramowitzStegun1964}, and ${}_pF_q(a_1, \cdots, a_p;b_1, \cdots, b_q; z)$ is the generalized hypergeometric function as defined in \citep{NISTHandbook2010}. 
\label{lemma:prop_1}
\end{lemma}

\subsubsection*{Proof of Lemma 6.1}
\label{ap:proof_prop_1}
Euler’s integral representation of the hypergeometric function ${}_2F_1$ is
\begin{equation*}
    B(b, c-b){}_2F_1\!\left(\begin{matrix} a, b \\ c \end{matrix}; z \right) = \int_0^1\frac{t^{b-1}(1-t)^{c-b-1}}{(1-zt)^a}dt
\end{equation*}
where $B(\cdot, \cdot)$ denotes the Beta function. The minimal conditions for the integral representation of the hypergeometric function is that 
\begin{equation*}
    \Re(c) > \Re(b) > 0,  \qquad |\arg(1-z)| < \pi
\end{equation*}
where the latter is satisfied for real $z$ when $z \notin [1, \infty)$ \citep{andrews_askey_roy_1999}.
Now, let $\xi \sim \mathrm{Beta}(\alpha, \beta)$ and consider
\begin{equation*}
\begin{aligned}
    \mathbb{E}_{\xi}\left[ \frac{\xi^k}{(1+s\xi)^a } \right] & = \frac{1}{B(\alpha, \beta)} \int_0^1\frac{\xi^{k+\alpha-1}(1-\xi)^{\beta-1}}{(1+s\xi)^a} d\xi \\
    & = \frac{1}{B(\alpha, \beta)} B(\alpha + k, \beta){}_2F_1\!\left(\begin{matrix} a, \alpha+k \\ \beta + \alpha + k \end{matrix}; -s \right) \\
    & = \frac{(\alpha)_k}{(\alpha + \beta)_k}{}_2F_1\!\left(\begin{matrix} a, \alpha+k \\ \beta + \alpha + k \end{matrix}; -s \right)
\end{aligned}
\end{equation*}
where $(x)_n$ denotes the Pochhammer symbol. We now check our conditions. With $b=\alpha+k$ and $c=\beta+\alpha+k$ it is clear that $\Re(c) > \Re(b) > 0$ as long as $k > -\alpha$. Furthermore, if $z\notin [1, \infty)$, then $-z = s \notin(-\infty, -1]$.\hfill $\blacksquare$

\begin{lemma}[Expectation of transformed Beta variable II]
Let $\xi_j \sim \mathrm{Beta}(\alpha, \beta)$, and let $k, s, a$ be as in Lemma \ref{lemma:prop_1}.
The expectation of the transform $\xi_j \mapsto \frac{\xi_j^k}{(1+s\sqrt{\xi})^a}$ is
\begin{align*}
    \mathbb{E}_{\xi}\left[ \frac{\xi^k}{(1+s\sqrt{\xi})^{a} } \right] =\frac{(\alpha)_k}{(\alpha+\beta)_k}&\Bigg[{}_3F_2\!\left(\begin{matrix} \frac{a}{2}, \frac{a+1}{2}, \alpha+k \\ \frac{1}{2}, \alpha+\beta+k\end{matrix}; s^2 \right) \\
    &-sa \frac{(\alpha + k)_{1/2}}{(\alpha+\beta + k)_{1/2}}{}_3F_2\!\left(\begin{matrix} \frac{a+1}{2}, \frac{a+2}{2}, \alpha +k+ \frac{1}{2} \\ \frac{3}{2}, \alpha+\beta +k+ \frac{1}{2}\end{matrix}; s^2 \right)\Bigg] \ ,
\end{align*} 
and for the special case $\beta=(p-1)\alpha, k=0, a=1$ we obtain for the transform $\xi_j \mapsto \frac{1}{1+s\sqrt{\xi_j}}$ that
\begin{align*}
    \mathbb{E}_{\xi_j}\left[\frac{1}{1+s\sqrt{\xi_j}} \right] & = {}_2F_1\!\left(\begin{matrix} 1, \alpha \\ p\alpha \end{matrix}; s^2 \right) - s\frac{(\alpha)_{1/2}}{(p\alpha)_{1/2}}{}_2F_1\!\left(\begin{matrix} 1, \alpha + \frac{1}{2} \\ p\alpha+ \frac{1}{2}\end{matrix}; s^2 \right)
\end{align*} 
\label{prop:prop_3}
\end{lemma}

\subsubsection*{Auxiliary Pochhammer identities}
\label{ap:auxillary_pochhammer}
To ease the derivations in the proof of Lemma \ref{prop:prop_3}, we state the following Pochhammer identites
\begin{equation}
\label{eq:poch_id_1}
\begin{aligned}
    \frac{(\frac{1}{2})_m}{(\frac{3}{2})_m} &= \frac{\frac{1}{2}(\frac{1}{2}+1)\cdots (\frac{1}{2}+m-1)}{\frac{3}{2}(\frac{3}{2}+1)\cdots (\frac{3}{2}+m-1)} = \frac{\frac{1}{2}}{(\frac{3}{2} + m -1)} = \frac{1}{2m+1}
\end{aligned} 
\end{equation}
\begin{equation}
\label{eq:poch_id_2}
\begin{aligned}
    (2m)! &= \Gamma(2m+1)= \frac{2^{2m}\Gamma(m+\tfrac12)\Gamma(m+1)}{\sqrt{\pi}}= \frac{2^{2m}\left(\frac{1}{2}\right)_{m}\sqrt{\pi}m!}{\sqrt{\pi}}= 2^{2m}m!\!\left(\tfrac12\right)_m
    \end{aligned} 
\end{equation}
\begin{equation}
\label{eq:poch_id_3}
\begin{aligned}
    (a)_{2m} &= 2^{2m}\left( \frac{a}{2}\right)_{m}\left(\frac{a+1}{2} \right)_{m} \\ (a)_{2m+1} &= a2^{2m}\left(\frac{a+1}{2} \right)_{m}\left( \frac{a+2}{2}\right)_{m}
\end{aligned} 
\end{equation}
\begin{equation}
\label{eq:poch_id_4}
\begin{aligned}
    \left(a+b\right)_{c+d} &= \tfrac{\Gamma(a+b+c+d)}{\Gamma(a+b)} =\tfrac{\Gamma(a+b+c)}{\Gamma(a+b)}\tfrac{\Gamma(a+b+c+d)}{\Gamma(a+b+c)}  = \left(a+b\right)_{c}\left(a+b+c\right)_{d}
\end{aligned} 
\end{equation}
\begin{equation}
\label{eq:poch_id_5}
\begin{aligned}
    \left(\frac{a}{2}\right)_{m+1} &= \frac{a}{2}\left(\frac{a}{2}+1\right)\left(\frac{a}{2}+2\right)\cdots \left(\frac{a}{2}+m\right) = \frac{a}{2}\left(\frac{a+2}{2}\right)_{m} 
\end{aligned} 
\end{equation}

\subsubsection*{Proof of Lemma 6.2}
\label{ap:proof_prop_3} Let $\xi \sim \mathrm{Beta}(\alpha, \beta), \quad \alpha,\beta > 0$. Let $a \in \mathbb{N}, a\geq1, k > -\alpha$, then for any real $s > -1$, we propose that
\begin{equation*}
\begin{aligned}
    \mathbb{E}_{\xi}\left[ \frac{\xi^k}{(1+s\sqrt{\xi})^{a} } \right] =\frac{(\alpha)_k}{(\alpha+\beta)_k}&\Bigg[{}_3F_2\!\left(\begin{matrix} \frac{a}{2}, \frac{a+1}{2}, \alpha+k \\ \frac{1}{2}, \alpha+\beta+k\end{matrix}; s^2 \right) \\
    &-sa \frac{(\alpha + k)_{1/2}}{(\alpha+\beta + k)_{1/2}}{}_3F_2\!\left(\begin{matrix} \frac{a+1}{2}, \frac{a+2}{2}, \alpha +k+ \frac{1}{2} \\ \frac{3}{2}, \alpha+\beta +k+ \frac{1}{2}\end{matrix}; s^2 \right)\Bigg]
\end{aligned} 
\end{equation*}
where ${}_3F_2$ is defined, following \citet{andrews_askey_roy_1999}, as
\begin{equation*}
\begin{aligned} 
    {}_3F_2\!\left(\begin{matrix} a_1, a_2, a_3 \\ b_1, b_2\end{matrix}; t \right) = \sum_{n=0}^{\infty}\frac{(a_1)_n(a_2)_n(a_3)_n}{(b_1)_n(b_2)_n}\frac{t^n}{n!} \ .
\end{aligned} 
\end{equation*}
To prove this proposition, define
\begin{equation*}
\begin{aligned}
    F(s) & \coloneq \mathbb{E}_{\xi}\left[\frac{\xi^k}{(1+s\sqrt{\xi})^a } \right] =  \frac{1}{B(\alpha, \beta)} \int_0^1\frac{1}{(1+s\sqrt{\xi})^a}\xi^{k+\alpha-1}(1-\xi)^{\beta-1} d\xi \\
    H(s) & \coloneq \frac{(\alpha)_k}{(\alpha+\beta)_k}\Bigg[{}_3F_2(\dots; s^2 ) - sa\frac{(\alpha+k)_{1/2}}{(\alpha+\beta+k)_{1/2}}{}_3F_2(\dots; s^2 )\Bigg]
\end{aligned} 
\end{equation*}
Consider the case of $|s| < 1$ and the binomial series
\begin{equation*}
\begin{aligned}
    \left(1+s\sqrt{\xi}\right)^{-a} &= \sum_{n=0}^{\infty} \binom{-a}{n}s^n\xi^{n/2}  
    =\sum_{n=0}^{\infty}\frac{(a)_n}{n!}(-s)^{n}\xi^{n/2} \\
\end{aligned} 
\end{equation*}
which is absolutely convergent since $\xi \in [0, 1]$. 
This gives 
\begin{equation*}
\begin{aligned}
    F(s) &= \frac{1}{B(\alpha, \beta)} \int_0^1 \sum_{n=0}^{\infty}\frac{(a)_n}{n!}(-s)^{n}\xi^{n/2} \xi^{k+\alpha-1}(1-\xi)^{\beta-1}  d\xi
\end{aligned} 
\end{equation*}
Now define $|f_n(\xi)| := |\frac{(a)_n}{n!}(-s)^{n}\xi^{n/2} \xi^{k+\alpha-1}(1-\xi)^{\beta-1}|$ and develop
\begin{equation*}
\begin{aligned}
    |f_n| & \leq \bigg|\frac{(a)_n}{n!}\bigg||s|^{n}\xi^{n/2} \xi^{k+\alpha-1}(1-\xi)^{\beta-1} \\
    & \leq \bigg|\frac{(a)_n}{n!}\bigg||s|^{n} \xi^{k+\alpha-1}(1-\xi)^{\beta-1} 
\end{aligned} 
\end{equation*}
such that
\begin{equation*}
\begin{aligned}
    \int_0^1 \sum_{n=0}^{\infty} |f_n| &\leq \int_0^1 \sum_{n=0}^{\infty}\bigg|\frac{(a)_n}{n!}\bigg| |s|^{n} \xi^{k+\alpha-1}(1-\xi)^{\beta-1} d\xi \\
    & = \int_0^1 \xi^{k+\alpha-1}(1-\xi)^{\beta-1} \sum_{n=0}^{\infty} \bigg|\frac{(a)_n}{n!}\bigg||s|^{n} d\xi
\end{aligned} 
\end{equation*}
Now define
\begin{equation*}
\begin{aligned}
    S\coloneq \sum_{n=0}^{\infty}\big|\frac{(a)_n}{n!}\big||s|^{n} \ .
\end{aligned} 
\end{equation*}
Since $a\in\mathbb{N}, a\geq 1$ we have
\begin{equation*}
\begin{aligned}
    \frac{(a)_n}{n!} = \frac{\Gamma(a+n)}{n!\Gamma(a)} = \frac{(a+n-1)!}{n!(a-1)!} = \binom{a+n-1}{n}
\end{aligned} 
\end{equation*}
hence
\begin{equation*}
\begin{aligned}
    S=\sum_{n=0}^{\infty}\binom{a+n-1}{n}|s|^{n} \ ,
\end{aligned} 
\end{equation*}
which is a negative binomial series, which for $|s|<1$ satisfies
\begin{equation*}
\begin{aligned}
    S=\sum_{n=0}^{\infty}\binom{a+n-1}{n}|s|^{n} = (1-|s|)^{-a} < \infty \ .
\end{aligned} 
\end{equation*}
Consequently, 
\begin{equation*}
\begin{aligned}
    \int_0^1\sum_{n=0}^{\infty} |f_n| &\leq  S\int_0^1 \xi^{k+\alpha-1}(1-\xi)^{\beta-1} d\xi < \infty \ ,
\end{aligned} 
\end{equation*}
and $\sum_{n=0}^{\infty} |f_n| \in L^1(0,1)$ is absolutely integrable on $(0, 1)$. This invokes the Fubini-Tonelli theorem, so a swap of the integral and summation is justified, yielding
\begin{align*}
    F(s) &= \frac{1}{B(\alpha, \beta)} \int_0^1 \sum_{n=0}^{\infty} \frac{(a)_n}{n!}(-s)^{n}\xi^{n/2} \xi^{k+\alpha-1}(1-\xi)^{\beta-1} d\xi \\
    &= \frac{1}{B(\alpha, \beta)} \sum_{n=0}^{\infty}\frac{(a)_n}{n!}(-s)  ^{n}\int_0^1  \xi^{k+\alpha+n/2-1}(1-\xi)^{\beta-1}d\xi \\
    &= \frac{1}{B(\alpha, \beta)} \sum_{n=0}^{\infty}\frac{(a)_n}{n!}(-s)^{n}B(k+\alpha+n/2, \beta) \\
    &= \sum_{n=0}^{\infty}\frac{(a)_n}{n!}(-s)^{n}\frac{B(k+\alpha+n/2, \beta)}{B(\alpha, \beta)} \\
    &= \sum_{n=0}^{\infty}\frac{(a)_n(\alpha)_{k+n/2}}{(\alpha + \beta)_{k+n/2}}\frac{(-s)^{n}}{n!} \\
    &= \frac{(\alpha)_k}{(\alpha+\beta)_k}\sum_{n=0}^{\infty}\frac{(a)_n(\alpha+k)_{n/2}}{(\alpha + \beta+k)_{n/2}}\frac{(-s)^{n}}{n!} 
\end{align*} 
To evaluate this expression, we want to split the series into its even and odd parts. Since this is a regrouping of terms, we must first verify that the series is
absolutely convergent (for $|s|<1$) before splitting. Consider the representation
\begin{equation*}
\begin{aligned} 
    F(s)=\frac{1}{B(\alpha,\beta)}\sum_{n=0}^{\infty}\frac{(a)_n}{n!}(-s)^n B\!\left(k+\alpha+\frac n2,\beta\right).
\end{aligned} 
\end{equation*}
Using the integral form of the Beta function,
\begin{equation*}
\begin{aligned} 
    B(x,\beta)=\int_0^1 t^{x-1}(1-t)^{\beta-1}\,dt,
\end{aligned} 
\end{equation*}
define, for fixed $\beta>0$,
\begin{equation*}
\begin{aligned} 
    g_x(t):=t^{x-1}(1-t)^{\beta-1}, \qquad t\in(0,1).
\end{aligned} 
\end{equation*}
If $x_2>x_1>0$, then for all $t\in(0,1)$ we have $t^{x_2-1}\le t^{x_1-1}$, hence
\begin{equation*}
\begin{aligned} 
    g_{x_2}(t)\le g_{x_1}(t).
\end{aligned} 
\end{equation*}
Since $g_x(t)\ge0$ and measurable, monotonicity of the Lebesgue integral gives
\begin{equation*}
\begin{aligned} 
    B(x_2,\beta)=\int_0^1 g_{x_2}(t)\,dt \le \int_0^1 g_{x_1}(t)\,dt = B(x_1,\beta).
\end{aligned} 
\end{equation*}
Thus $x\mapsto B(x,\beta)$ is decreasing on $(0,\infty)$. In particular, since $k+\alpha>0$,
\begin{equation*}
\begin{aligned} 
    B\!\left(k+\alpha+\frac n2,\beta\right)\le B(k+\alpha,\beta) \quad\text{for all } n\ge0.
\end{aligned} 
\end{equation*}
Therefore,
\begin{equation*}
\begin{aligned} 
    \sum_{n=0}^\infty  \left|\frac{(a)_n}{n!}(-s)^n\frac{B(k+\alpha+n/2,\beta)}{B(\alpha,\beta)}\right| \le C\sum_{n=0}^\infty \frac{(a)_n}{n!}|s|^n, \qquad C:=\frac{B(k+\alpha,\beta)}{B(\alpha,\beta)}.
\end{aligned} 
\end{equation*}
For $|s|<1$, the right-hand side equals $C(1-|s|)^{-a}<\infty$, and hence the series is absolutely convergent. Consequently, we may regroup terms and split the series into its even and odd parts. Note the auxiliary identities of Supplementary \ref{ap:auxillary_pochhammer} and first consider $n=2m$
  
\begin{align*}
    \sum_{n=0}^{\infty}\frac{(a)_n(\alpha + k)_{n/2}}{(\alpha + \beta + k)_{n/2}}&\frac{(-s)^{n}}{n!}
     = \sum_{m=0}^{\infty}\frac{(a)_{2m}(\alpha+k)_{m}}{(\alpha + \beta+k)_{m}}\frac{(-s^2)^{m}}{(2m)!} \\
    & \overset{\ref{eq:poch_id_2}, \ref{eq:poch_id_3}}{=} 
    \sum_{m=0}^{\infty}2^{2m}\left(\frac{a}{2}\right)_m\left(\frac{a+1}{2}\right)_m\frac{(\alpha+k)_{m}}{(\alpha + \beta+k)_{m}}\frac{1}{\left(\frac{1}{2}\right)_m2^{2m}}\frac{(-s^2)^{m}}{m!} \\
    & = {}_3F_2\!\left(\begin{matrix} \frac{a}{2}, \frac{a+1}{2}, \alpha+k \\ \frac{1}{2}, \alpha+\beta+k\end{matrix}; s^2 \right) \ .
\end{align*} 
 
Then consider $n=2m+1$
  
\begin{align*}
    &\sum_{n=0}^{\infty}\frac{(a)_n(\alpha+k)_{n/2}}{(\alpha + \beta+k)_{n/2}}\frac{(-s)^{n}}{n!}
     = \sum_{m=0}^{\infty}\frac{(a)_{2m+1}(\alpha+k)_{m+1/2}}{(\alpha + \beta+k)_{m+1/2}}\frac{(-s)^{2m+1}}{(2m+1)!} \\
    & \overset{\eqref{eq:poch_id_2}, \eqref{eq:poch_id_3},\eqref{eq:poch_id_4}}{=} -s\tfrac{(\alpha+k)_{1/2}}{(\alpha+\beta+k)_{1/2}}\sum_{m=0}^{\infty}2^{2m+1}\left(\tfrac a2\right)_{m+1}\left(\tfrac{a+1}{2}\right)_m\frac{(\alpha+k + \frac{1}{2})_{m}}{(\alpha + \beta+k + \frac{1}{2})_{m}}\frac{(s^2)^{m}}{(2m+1)(2m)!} \\
    & \overset{\eqref{eq:poch_id_1}, \eqref{eq:poch_id_5}}{=}  -s\tfrac{(\alpha+k)_{1/2}}{(\alpha+\beta+k)_{1/2}}\sum_{m=0}^{\infty}\tfrac{2a}{2}\left(\tfrac{a+2}{2}\right)_{m}\left(\tfrac{a+1}{2}\right)_m\frac{(\alpha +k+ \frac{1}{2})_{m}}{(\alpha + \beta+k + \frac{1}{2})_{m}}\frac{(s^2)^{m}}{(2m+1)\left( \frac{1}{2}\right)_mm!} \\
    & \overset{ \eqref{eq:poch_id_2}}{=}  -sa\tfrac{(\alpha+k)_{1/2}}{(\alpha+\beta+k)_{1/2}}\sum_{m=0}^{\infty}\left(\tfrac{a+1}{2}\right)_m\left(\tfrac{a+2}{2}\right)_{m}\frac{(\alpha +k+ \frac{1}{2})_{m}}{(\alpha + \beta +k+ \frac{1}{2})_{m}}\frac{(s^2)^{m}}{\left( \frac{3}{2}\right)_mm!} \\
    & = -sa \tfrac{(\alpha+k)_{1/2}}{(\alpha+\beta+k)_{1/2}}{}_3F_2\!\left(\begin{matrix} \frac{a+1}{2}, \frac{a+2}{2}, \alpha +k+ \frac{1}{2} \\ \frac{3}{2}, \alpha+\beta +k+ \frac{1}{2}\end{matrix}; s^2 \right) 
\end{align*} 
 
and then we finally arrive at
  
\begin{align*}
    F(s) &= \mathbb{E}_{\xi}\left[ \frac{\xi^k}{(1+s\sqrt{\xi})^{a} } \right] \\
    &= \frac{(\alpha)_k}{(\alpha+\beta)_k}
    \Bigg[
      {}_3F_2\!\left(
        \begin{matrix}
          \frac{a}{2}, \frac{a+1}{2}, \alpha+k \\
          \frac{1}{2}, \alpha+\beta+k
        \end{matrix}
        ; s^2
      \right) \\
    &\qquad\qquad
      -\, s a \frac{(\alpha+k)_{1/2}}{(\alpha+\beta+k)_{1/2}}
      {}_3F_2\!\left(
        \begin{matrix}
          \frac{a+1}{2}, \frac{a+2}{2}, \alpha +k+ \frac{1}{2} \\
          \frac{3}{2}, \alpha+\beta +k+ \frac{1}{2}
        \end{matrix}
        ; s^2
      \right)
   \Bigg] 
    = H(s).
\end{align*} 
 
Now, we have shown that on the domain $s\in S = (-1, 1)$, $F(s)=H(s)$. Moreover, $F$ is real analytic on $(-1, \infty)$, since for every $s_0 > -1$, the integrand admits a power series expansion in $s-s_0$ with a positive radius of convergence, meaning that it can be integrated term by term to give a locally convergent power series representation of $F(s)$. The hypergeometric function $H(s)$ is real analytic by definition. As both $F, H$ are analytic, the domain $D=(-1, \infty)$ is open and connected with $S\subseteq D$ and $S$ has accumulation points in $D$, the identity theorem allows us to analytically continue into domain $D$ such that $F(s)=H(s)$ on $(-1, \infty)$. Thus, the expectation holds for all $s > -1$. \\ \\
For the case of $k=0, a=1$, the expectation reduces because of equal factors in the hypergeometric function, and by recalling that the marginal was parametrized by $\xi \sim \mathrm{Beta}(\alpha, (p-1)\alpha)$ we obtain the expression in Lemma \ref{prop:prop_3}:
  
    \begin{align*}
    \mathbb{E}_{\xi_j}\left[\frac{\xi^0}{(1+s\sqrt{\xi_j})^1} \right] &= \mathbb{E}_{\xi_j}\left[\frac{1}{1+s\sqrt{\xi_j}} \right] \nonumber\\
    &= {}_3F_2\!\left(\begin{matrix} \frac{1}{2}, 1, \alpha \\ \frac{1}{2}, \alpha+\beta\end{matrix}; s^2 \right) -s \frac{(\alpha)_{1/2}}{(\alpha+\beta)_{1/2}}{}_3F_2\!\left(\begin{matrix} 1, \frac{3}{2}, \alpha + \frac{1}{2} \\ \frac{3}{2}, \alpha+\beta + \frac{1}{2}\end{matrix}; s^2 \right) \\
    &={}_2F_1\!\left(\begin{matrix} 1, \alpha \\ p\alpha \end{matrix}; s^2 \right) - s\frac{(\alpha)_{1/2}}{(p\alpha)_{1/2}}{}_2F_1\!\left(\begin{matrix} 1, \alpha + \frac{1}{2} \\ p\alpha+ \frac{1}{2}\end{matrix}; s^2 \right)\hfill\nonumber
    \end{align*} 
 
\hfill$\blacksquare$

\subsubsection*{Proof of Theorem 4.1 for $\nu=1$}
\label{ap:proof_theorem_1}
Let $\xi_j \sim \mathrm{Beta}(\alpha, (p-1)\alpha)$ and derive the distribution using Lemma \ref{lemma:prop_1} with $k=\frac{1}{2}, a=1$
  
\begin{align*}
    p(\kappa_j \mid \tau, \sigma) &= \int_0^1 p(\kappa_j \mid \tau, \sigma, \xi_j)p(\xi_j) d\xi_j \\
    & = \int_0^1 \frac{1}{\pi} \frac{1}{\sqrt{\kappa_j}\sqrt{1-\kappa_j}} \frac{z_j\sqrt{ \xi_j}}{(\xi_jz_j^2-1)\kappa_j +1}  p(\xi_j) d\xi_j \\
    & = \int_0^1 \frac{1}{\pi} \frac{z_j}{\sqrt{\kappa_j}\sqrt{1-\kappa_j}} \frac{\xi_j^{\frac{1}{2}}}{(1-\kappa_j)(1+\frac{\kappa \xi_j z_j^2}{1-\kappa_j})}  p(\xi_j) d\xi_j\\
    & = \frac{1}{\pi} \frac{z_j}{(1-\kappa_j)\sqrt{\kappa_j}\sqrt{1-\kappa_j}} \int_0^1\frac{\xi_j^{\frac{1}{2}}}{(1+s\xi_j)}  p(\xi_j) d\xi_j\\
    & = \frac{1}{\pi} \frac{z_j}{(1-\kappa_j)\sqrt{\kappa_j}\sqrt{1-\kappa_j}} \mathbb{E}_{\xi_j}\left[ \frac{\xi_j^{\frac{1}{2}}}{1+s\xi_j } \right] \\
    & = \frac{1}{\pi} \frac{z_j}{(1-\kappa_j)\sqrt{\kappa_j}\sqrt{1-\kappa_j}} \frac{(\alpha)_{1/2}}{(p\alpha)_{1/2}}{}_2F_1\!\left(\begin{matrix} 1, \alpha + \frac{1}{2} \\ p\alpha + \frac{1}{2} \end{matrix}; -s \right)
\end{align*} 
 
where $s=\frac{\kappa_j z_j^2}{1-\kappa_j}$. The expectation can be derived, using Lemma \ref{prop:prop_3}, as
  
\begin{align*}
    \mathbb{E}_{\xi_j}[\kappa_{j} \mid \tau, \sigma] & = \mathbb{E}_{\xi_j}\big[\mathbb{E}_{\lambda_j}[\kappa_{j} \mid \tau, \sigma, \xi_j]\big] \\
    & = \mathbb{E}_{\xi_j}\bigg[\frac{1}{1+z_j\sqrt{\xi_j}}\bigg] \\
    & = {}_2F_1\!\left(\begin{matrix} 1, \alpha \\ p\alpha \end{matrix}; z_j^2 \right) - z_j\frac{(\alpha)_{1/2}}{(p\alpha)_{1/2}}{}_2F_1\!\left(\begin{matrix} 1, \alpha+\frac{1}{2} \\ p\alpha + \frac{1}{2} \end{matrix}; z_j^2 \right)
\end{align*}
and the variance, again using Lemma \ref{prop:prop_3},
\begin{align*}
    \mathrm{Var}_{\xi_j}[\kappa_{j} \mid \tau, \sigma]  &= \mathbb{E}_{\xi_j}[\text{Var}_{\lambda}(\kappa_j) \mid \tau, \sigma, \xi_j] +  \mathrm{Var}_{\xi_j}(\mathbb{E}_{\lambda}[\kappa_{j} \mid \tau, \sigma, \xi_j]) \\
    & = \mathbb{E}_{\xi_j}\bigg[\frac{z_j\sqrt{\xi_j}}{2(1+z_j\sqrt{\xi_j})^2}\bigg] +  \text{Var}_{\xi_j}\left(\frac{1}{1+z_j\sqrt{\xi_j}}\right) \\
    &= \frac{z_j}{2} \mathbb{E}_{\xi_j}\bigg[\frac{\sqrt{\xi_j}}{(1+z_j\sqrt{\xi_j})^2}\bigg]  + \mathbb{E}_{\xi_j}\bigg[ \frac{1}{(1+z_j\sqrt{\xi_j})^2} \bigg] - \left(\mathbb{E}_{\xi_j}\bigg[ \frac{1}{1+z_j\sqrt{\xi_j}} \bigg] \right)^2 \\
    &= \frac{z_j}{2}\frac{(\alpha)_{1/2}}{(p\alpha)_{1/2}}
   \Bigg[
      {}_3F_2\!\left(
        \begin{matrix}
          1, \tfrac{3}{2}, \alpha+\tfrac{1}{2} \\
          \tfrac{1}{2}, p\alpha+\tfrac{1}{2}
        \end{matrix}
        ; z_j^2
      \right)
    \\[4pt]
    &\qquad\qquad
      -\, 2 z_j
        \frac{(\alpha+\tfrac{1}{2})_{1/2}}{(p\alpha+\tfrac{1}{2})_{1/2}}
        {}_2F_1\!\left(
          \begin{matrix}
            2, \alpha+1 \\
            p\alpha+1
          \end{matrix}
          ; z_j^2
        \right)
   \Bigg]
    \\
    & + {}_3F_2\!\left(\begin{matrix} 1, \frac{3}{2}, \alpha \\ \frac{1}{2}, p\alpha\end{matrix}; z_j^2 \right) -2z_j \frac{(\alpha+\frac{1}{2})_{1/2}}{(p\alpha+\frac{1}{2})_{1/2}}{}_2F_1\!\left(\begin{matrix} 2, \alpha +\frac{1}{2} \\  p\alpha +\frac{1}{2}\end{matrix}; z_j^2 \right)  \\
    &- \Bigg[{}_2F_1\!\left(\begin{matrix} 1, \alpha \\ p\alpha \end{matrix}; z_j^2 \right) - z_j\frac{(\alpha)_{1/2}}{(p\alpha)_{1/2}}{}_2F_1\!\left(\begin{matrix} 1, \alpha+\frac{1}{2} \\ p\alpha + \frac{1}{2} \end{matrix}; z_j^2 \right) \Bigg]^2 
\end{align*}
\hfill $\blacksquare$

\begin{lemma}[A priori distribution of shrinkage factor for student T local scale]
Let $\kappa = \frac{1}{1+z^2\lambda^2}$ in which $z$ is assumed fixed, and $\lambda$ follow a positively truncated Student T distribution with $\nu$ degrees of freedom. Then $\kappa$ follows the distribution
\begin{align*}
    p_{\kappa}(\kappa \mid z) &= \frac{\Gamma(\frac{\nu+1}{2})}{\sqrt{\nu\pi}\Gamma(\frac{\nu}{2})}  \frac{1}{(1-\kappa)^{\frac{\nu}{2}+1}}\nu^{\frac{\nu + 1}{2}}\kappa^{\frac{\nu}{2}-1} z^{\nu}\left(1+\frac{\kappa\nu z^2}{1-\kappa}\right)^{-\frac{\nu+1}{2}}
\end{align*}
\label{prop:prop_2}
\end{lemma}
\subsubsection*{Proof of Lemma 6.3}
\label{ap:proof_prop_2}
Let $\kappa = \frac{1}{1+z^2\lambda^2}$ in which $s$ is assumed fixed, and $\lambda$ follow a half Student T distribution with $\nu$ degrees of freedom (half Cauchy coincides with $\nu=1$). 
We thus have 
\begin{equation}
\begin{aligned}
    p_{\lambda}(\lambda) &= \frac{2\Gamma(\frac{\nu+1}{2})}{\sqrt{\nu\pi}\Gamma(\frac{\nu}{2})} \left(1 + \frac{\lambda^2}{\nu} \right)^{-\frac{\nu+1}{2}}, \quad
    \lambda = \frac{1}{z}\sqrt{\frac{1-\kappa}{\kappa}}, \quad
    \bigg|\frac{d\lambda}{d\kappa}\bigg| = \frac{1}{2z}\frac{1}{\kappa^{3/2}\sqrt{1-\kappa}}
\end{aligned}
\end{equation}
which then means we obtain
  
\begin{align*}
    p_{\kappa}(\kappa) &= p_{\lambda}\left(\frac{1}{z}\sqrt{\frac{1-\kappa}{\kappa}} \right) \bigg|\frac{d\lambda}{d\kappa}\bigg| \\
    & = \frac{2\Gamma(\frac{\nu+1}{2})}{\sqrt{\nu\pi}\Gamma(\frac{\nu}{2})}\frac{1}{\kappa^{3/2}\sqrt{1-\kappa}}  \frac{1}{2z} \left(1+\frac{\left(\frac{1}{z}\sqrt{\frac{1-\kappa}{\kappa}}\right)^2}{\nu} \right)^{-\frac{\nu+1}{2}} \\
    & = \frac{\Gamma(\frac{\nu+1}{2})}{\sqrt{\nu\pi}\Gamma(\frac{\nu}{2})}  \frac{1}{\kappa^{3/2}\sqrt{1-\kappa}}\frac{1}{z} \left(1+\frac{1-\kappa}{\nu z^2\kappa} \right)^{-\frac{\nu+1}{2}} \\
    & = \frac{\Gamma(\frac{\nu+1}{2})}{\sqrt{\nu\pi}\Gamma(\frac{\nu}{2})}  \frac{1}{\kappa^{3/2}\sqrt{1-\kappa}}\frac{1}{z}\left(\frac{\nu z^2\kappa + 1-\kappa}{\nu z^2\kappa} \right)^{-\frac{\nu+1}{2}} \\
    & = \frac{\Gamma(\frac{\nu+1}{2})}{\sqrt{\nu\pi}\Gamma(\frac{\nu}{2})}\frac{1}{\kappa^{3/2}\sqrt{1-\kappa}} \frac{1}{z} \left(\frac{\kappa(\nu z^2-1) + 1}{\nu z^2\kappa} \right)^{-\frac{\nu+1}{2}} \\
    & = \frac{\Gamma(\frac{\nu+1}{2})}{\sqrt{\nu\pi}\Gamma(\frac{\nu}{2})} \frac{1}{\kappa^{3/2}\sqrt{1-\kappa}}\frac{1}{z}\left(\nu z^2\kappa \right)^{\frac{\nu+1}{2}} \left(\kappa(\nu z^2-1) + 1 \right)^{-\frac{\nu+1}{2}} \\
    & =  \frac{\Gamma(\frac{\nu+1}{2})}{\sqrt{\nu\pi}\Gamma(\frac{\nu}{2})}  \frac{1}{(1-\kappa)^{\frac{\nu}{2}+1}}\nu^{\frac{\nu + 1}{2}} z^{\nu}\kappa^{\frac{\nu}{2}-1}\left(1+\frac{\kappa\nu z^2}{1-\kappa}\right)^{-\frac{\nu+1}{2}}
\end{align*}
As a sanity check, we insert $\nu=1$ to make sure we agree with Piironen
\begin{align*}
    p_{\kappa}(\kappa) &= \frac{\Gamma(1)}{\sqrt{\pi}\Gamma(\frac{1}{2})}  \left(1+\frac{\kappa z^2}{1-\kappa}\right)^{-1}\frac{1}{(1-\kappa)^{\frac{1}{2}+1}} z\kappa^{\frac{1}{2}-1} \\
    &= \frac{z}{\pi}\left(\frac{1-\kappa}{(\kappa(z^2-1)+1)}\right)^{-1}\frac{1}{(1-\kappa)^{\frac{3}{2}}\kappa^{\frac{1}{2}}} \\
    & = \frac{z}{\pi}\frac{1}{\left(\kappa(z^2-1) + 1 \right)}\frac{1}{\sqrt{\kappa(1-\kappa)}}
\end{align*}
which is exactly what \citet{piironen2017sparsity} has. 

\subsubsection*{Proof of Theorem 4.1 for general $\nu$}
\label{ap:proof_theorem_2}
The distribution of $\kappa_j$, using Lemma \ref{prop:prop_2}, can be written as
  
\begin{align*}
    p(\kappa_j \mid \tau, \sigma) &=\int_0^1 p(\kappa_j \mid \tau, \sigma, \xi_j)p(\xi_j)d\xi_j \\
    & = \int_0^1\frac{\Gamma(\frac{\nu+1}{2})}{\sqrt{\nu\pi}\Gamma(\frac{\nu}{2})}\nu^{\frac{\nu + 1}{2}}z_j^{\nu}\frac{\kappa^{\frac{\nu}{2}-1}}{(1-\kappa)^{\frac{\nu}{2}+1}} \xi_j^{\nu/2}\left(1+\frac{\kappa\nu \xi_jz_j^2 }{1-\kappa}\right)^{-\frac{\nu+1}{2}} p(\xi_j)d\xi_j \\
    &= \tilde{C}(\nu, z_j)\frac{\kappa^{\frac{\nu}{2}-1}}{(1-\kappa)^{\frac{\nu}{2}+1}}\int_0^1 \xi_j^{\nu/2}\left(1+\frac{\kappa\nu \xi_jz_j^2}{1-\kappa}\right)^{-\frac{\nu+1}{2}}p(\xi_j)d\xi_j \\
    &= \tilde{C}(\nu, z_j)\frac{\kappa^{\frac{\nu}{2}-1}}{(1-\kappa)^{\frac{\nu}{2}+1}}\int_0^1 \xi_j^{\nu/2}\left(1+s\xi_j\right)^{-\frac{\nu+1}{2}}p(\xi_j) d\xi_j \\
    &= \tilde{C}(\nu, z_j)\frac{\kappa^{\frac{\nu}{2}-1}}{(1-\kappa)^{\frac{\nu}{2}+1}} \mathbb{E}_{\xi_j}\left[\frac{\xi_j^{\nu/2}}{\left(1+s\xi_j\right)^{\frac{\nu+1}{2}}}\right]
\end{align*}
 
where 
  
\begin{align*}
    \tilde{C}(\nu, z_j) &= \frac{\Gamma(\frac{\nu+1}{2})}{\sqrt{\nu\pi}\Gamma(\frac{\nu}{2})}\nu^{\frac{\nu + 1}{2}}z_j^{\nu} \\
    s &=\frac{\kappa\nu z_j^2}{1-\kappa}
\end{align*}
 
which by using Lemma \ref{lemma:prop_1} with $k=\nu/2$, $a=\frac{\nu+1}{2}$ yields
\begin{equation}
\begin{aligned}
    p(\kappa_j \mid \tau, \sigma) &= \tilde{C}(\nu, z_j) \frac{(\alpha)_{\nu/2}}{(p\alpha)_{\nu/2}}\frac{\kappa^{\frac{\nu}{2}-1}}{(1-\kappa)^{\frac{\nu}{2}+1}}{}_2F_1\!\left(\begin{matrix} \frac{\nu+1}{2}, \alpha + \frac{\nu}{2} \\ p\alpha + \frac{\nu}{2}  \end{matrix}; -\frac{\kappa\nu z_j^2}{1-\kappa} \right) \ .
\end{aligned}
\end{equation}

\subsection*{Supplementary C: Linearization}
\label{ap:linearization_full}
In this supplement we give the details underlying the linearized Gaussian model and posterior for $\mathbf{w}_1$ used in the article.
\label{sec:linearization}
Let $\mathbf{X}\in\mathbb{R}^{n\times p}$ with rows $\mathbf{x}_i^{\top}$, and consider a single hidden layer with weights $W_1\in\mathbb{R}^{H\times p}$, biases $\mathbf{b}_1\in\mathbb{R}^{H}$, output weights $W_L\in\mathbb{R}^{1\times H}$, and output bias $b_L$. Define the hidden activations
\[
\boldsymbol{\phi}_i
= \varphi(W_1\mathbf{x}_i+\mathbf{b}_1)\in\mathbb{R}^{H},
\]
and collect them in the feature matrix
\[
\Phi(\mathbf{w}_1,\mathbf{b}_1)
=
\begin{bmatrix}
\boldsymbol{\phi}_1^{\top}\\[-2pt]
\vdots\\[-2pt]
\boldsymbol{\phi}_n^{\top}
\end{bmatrix}
\in\mathbb{R}^{n\times H},
\]
where $\mathbf{w}_1=\mathrm{vec}(W_1^\top)\in\mathbb{R}^{pH}$ and $\mathbf{w}_L=\mathrm{vec}(W_L^\top)\in\mathbb{R}^{H}$. The network output and observation model are
\begin{align*}
    f(\mathbf{w}_1,\mathbf{b}_1,\mathbf{w}_L,b_L)
    &= \Phi(\mathbf{w}_1,\mathbf{b}_1)\mathbf{w}_L + b_L\,\mathbf{1}_n, \\
    y &= f(\mathbf{w}_1,\mathbf{b}_1,\mathbf{w}_L,b_L) + \varepsilon,
    \qquad \varepsilon \sim \mathcal{N}(0,\,\sigma^2 I_n).
\end{align*}
We place a DSM prior on the input weights and standard Gaussian priors on the remaining parameters
\begin{gather*}
    \mathbf{w}_1 \sim \mathcal{N}\!\big(0,\;\tau^2\,\Psi\big), 
    \quad \Psi=\mathrm{diag}(\lambda_1^2\xi_1,\ldots,\lambda_{pH}^2\xi_{pH}), \\
    \mathbf{b}_1 \sim \mathcal{N}(0,I_H), \qquad
    \mathbf{w}_L \sim \mathcal{N}(0, I_H), \qquad
    b_L \sim \mathcal{N}(0, 1) \ ,
\end{gather*}
and linearize the network around a reference point  $(\mathbf{w}_{1,0},\mathbf{b}_{1,0},\mathbf{w}_{L,0}, b_{L,0})$. Writing $\Phi_0 := \Phi(\mathbf{w}_{1,0},\mathbf{b}_{1,0})$ and defining the Jacobians
\begin{align*}
    \mathbf{J}_w &= \frac{\partial \big(\Phi(\mathbf{w}_1,\mathbf{b}_1)\mathbf{w}_L\big)}
               {\partial\,\mathbf{w}_1}\Big|_{(\mathbf{w}_{1,0},\mathbf{b}_{1,0},\mathbf{w}_{L,0})} 
          \in \mathbb{R}^{n \times pH}, \\
    \mathbf{J}_b &= \frac{\partial \big(\Phi(\mathbf{w}_1,\mathbf{b}_1)\mathbf{w}_L\big)}
               {\partial\,\mathbf{b}_1}\Big|_{(\mathbf{w}_{1,0},\mathbf{b}_{1,0},\mathbf{w}_{L,0})} 
          \in \mathbb{R}^{n \times H},
\end{align*}
a first-order Taylor expansion yields
\begin{align*}
    \Phi(\mathbf{w}_1,\mathbf{b}_1)\mathbf{w}_L
    \;\approx\;
    \Phi_0 \mathbf{w}_L
    + \mathbf{J}_w (\mathbf{w}_1 - \mathbf{w}_{1,0})
    + \mathbf{J}_b (\mathbf{b}_1 - \mathbf{b}_{1,0}) .
\end{align*}
Absorbing constants into the response by letting $y^* := y  + \mathbf{J}_w\,\mathbf{w}_{1,0} + \mathbf{J}_b\,\mathbf{b}_{1,0}$, the linearized model is
\begin{align*}
    \mathbf{y}^* 
    \;\approx\;
    \mathbf{J}_w\,\mathbf{w}_{1} + \mathbf{J}_b\,\mathbf{b}_1 + \Phi_0 \mathbf{w}_L + b_L\mathbf{1}_n + \varepsilon.
\end{align*}
Conditioning on $(\tau,\lambda,\xi)$, we can integrate out $(\mathbf{b}_1,\mathbf{w}_L,b_L)$ to obtain the marginal likelihood
\[
    \mathbf{y}^* \mid \mathbf{w}_1 \sim \mathcal{N}(\mathbf{J}_w \mathbf{w}_1,\Sigma_y),
\]
with
\begin{align*}
    \Sigma_y &=  J_b J_b^{\top}+ \Phi_0\Phi_0^{\top}
               + \mathbf{1}_n\mathbf{1}_n^{\top} + \sigma^2 I_n
               \;\in\; \mathbb{R}^{n\times n}.
\end{align*}
Together with the prior $\mathbf{w}_1\sim\mathcal{N}(0,\tau^2\Psi)$, this defines a Gaussian prior–likelihood pair.

\subsubsection*{Linearized posterior distribution}
\label{ap:linearization}
It is a well known property of the normal distributions that for
\begin{align}
    \begin{bmatrix}
        \mathbf{w} \\
        \mathbf{y}^*
    \end{bmatrix} &\sim \mathcal{N}\left( \begin{bmatrix}
        0 \\
        0
    \end{bmatrix}, \begin{bmatrix}
        \tau^2\Psi & \tau^2\Psi \mathbf{J}^{\top} \\
        \tau^2\mathbf{J}\Psi & \Sigma_{y} + \tau^2\mathbf{J}\Psi \mathbf{J}^{\top}
    \end{bmatrix}\right) \ ,
\end{align}
we have
\begin{align}
    \mathbf{w}\mid \mathbf{y}^* & \sim \mathcal{N}\left(\mathbf{\bar w}, \bar{\Sigma}_w\right) \ , \\
    \mathbf{\bar w} &= \tau^2\Psi \mathbf{J}^{\top}(\Sigma_{y}+\tau^2\mathbf{J}\Psi \mathbf{J}^{\top})^{-1}\mathbf{y}^* \ , \\
    \bar{\Sigma}_w &= \tau^2\Psi - \tau^2\Psi \mathbf{J}^{\top} (\Sigma_{y}+\tau^2\mathbf{J}\Psi \mathbf{J}^{\top})^{-1}\mathbf{J}\tau^2\Psi \ .
\end{align}
This can be rewritten by defining the matrices
\begin{align*}
    P = \tau^{-2}\Psi^{-1} \quad S=\mathbf{J}^{\top}\Sigma_y^{-1}\mathbf{J} \ ,
\end{align*}
and using the Woodbury identity
\begin{align*}
    (\mathcal{A}+\mathcal{U}\mathcal{C}\mathcal{V})^{-1} = \mathcal{A}^{-1}-\mathcal{A}^{-1}\mathcal{U}(\mathcal{C}^{-1}+\mathcal{V}\mathcal{A}^{-1}\mathcal{U})^{-1}\mathcal{V}\mathcal{A}^{-1}
\end{align*}
For the covariance matrix $\bar{\Sigma}_w$, the identity is applied straightforward by defining the following relations
\begin{align*}
    \mathcal{A}=\tau^{-2}\Psi^{-1} \quad \mathcal{U}=\mathbf{J}^{\top} \quad \mathcal{V}=\mathbf{J} \quad \mathcal{C}=\Sigma_y^{-1} \ , \\
\end{align*}
to obtain
\begin{align*}
    \bar{\Sigma}_w &= \tau^2\Psi - \tau^2\Psi \mathbf{J}^{\top} (\Sigma_{y}+\tau^2\mathbf{J}\Psi \Psi \mathbf{J}^{\top})^{-1}\mathbf{J}\Psi \tau^2\Psi \\
    & = \mathcal{A}^{-1} - \mathcal{A}^{-1}\mathcal{U}(\mathcal{C}^{-1}+\mathcal{V}\mathcal{A}^{-1}\mathcal{U})^{-1}\mathcal{V}\mathcal{A}^{-1} \\
    & = (\tau^{-2}\Psi^{-1} + \mathbf{J}^{\top}\Sigma^{-1}_y\mathbf{J}\Psi )^{-1} \\
    & = (P+S)^{-1} \ .
\end{align*}
For the expectation, $\mathbf{\bar w}$, define
\begin{align*}
    \mathcal{A}=\Sigma_y &\quad \mathcal{U}=\mathbf{J} \quad \mathcal{V}=\mathbf{J}^{\top} \quad \mathcal{C}=\tau^2\Psi \ ,
\end{align*}
such that 
\begin{align*}
    \mathbf{\bar w} &= \tau^2\Psi \mathbf{J}^{\top}(\Sigma_{y}+\tau^2\mathbf{J}\Psi \Psi \mathbf{J}^{\top})^{-1}\mathbf{y}^* \\
    & = \mathcal{C}\mathcal{V}(\mathcal{A}+\mathcal{U}\mathcal{C}\mathcal{V})^{-1}\mathbf{y}^* = \mathcal{C}\mathcal{V}\left(\mathcal{A}^{-1} - \mathcal{A}^{-1}\mathcal{U}(\mathcal{C}^{-1}+\mathcal{V}\mathcal{A}^{-1}\mathcal{U})^{-1}\mathcal{V}\mathcal{A}^{-1}\right)\mathbf{y}^* \\
    & = \tau^2\Psi \mathbf{J}^{\top}\left(\Sigma_y^{-1} - \Sigma_y^{-1}\mathbf{J}(\tau^{-2}\Psi^{-1}+\mathbf{J}^{\top}\Sigma_y^{-1}\mathbf{J})^{-1}\mathbf{J}^{\top}\Sigma_y^{-1}\right)\mathbf{y}^* \\
    & = P^{-1}\mathbf{J}^{\top}\Sigma_y^{-1}\mathbf{y}^* - P^{-1}S(P+S)^{-1}\mathbf{J}^{\top}\Sigma_y^{-1}\mathbf{y}^* \\
    & = P^{-1}\left( I - S(P+S)^{-1} \right)S\hat{\bf w} \\
    & = P^{-1}P(P+S)^{-1}S\hat{\bf w} \\
    & = (P+S)^{-1}S\hat{\bf w} \ .
\end{align*}
where $\hat{\bf w} = (\mathbf{J}^{\top}\Sigma_y^{-1}\mathbf{J})^{-1}\mathbf{J}^{\top}\Sigma_{y}^{-1}\mathbf{y}^*$ is the generalized least square estimator. Finally, using the identity
\[
    (P+S)^{-1}S = I - (P+S)^{-1}P,
\]
we obtain the shrinkage matrix
\[
    K := (P+S)^{-1}S = I - (P+S)^{-1}P,
\]
which is the form used in the main text for the analysis of shrinkage in the
diagonal and general cases of $S$.

\subsubsection*{Whitening the shrinkage matrix}
\label{ap:shrinkage_matrix}
Note that $S$ and $P$ are symmetric PSD matrices, with $P$ diagonal.  
We can express the shrinkage operator as
\begin{align*}
    \left(I - (P+S)^{-1}P\right) &= (P+S)^{-1}S \\
    & = \left(P^{1/2}(I+P^{-1/2}SP^{-1/2})P^{1/2}\right)^{-1}S \\
    & = \left(P^{1/2}(I+G)P^{1/2}\right)^{-1}S \\
    & = P^{-1/2}\left(I+G\right)^{-1}P^{-1/2}P^{1/2}GP^{1/2} \\
    & = P^{-1/2}\left(I+G\right)^{-1}GP^{1/2} \ ,
\end{align*}
where $G = P^{-1/2}SP^{-1/2}$.  
Since $S$ and $P$ are PSD, $G$ is also symmetric and PSD
\begin{align*}
    G^\top = G, \qquad x^\top G x = (P^{-1/2}x)^\top S (P^{-1/2}x) \ge 0 \ .
\end{align*}
By the spectral theorem, let $G = U\Omega U^\top$ with diagonal $\Omega = \mathrm{diag}(\omega_i)$, giving
\begin{align*}
    (I - (P+S)^{-1}P)
    &= P^{-1/2}U (I+\Omega)^{-1}\Omega U^\top P^{1/2} \\
    &= P^{-1/2}U\,\mathrm{diag}\!\left(\frac{\omega_j}{1+\omega_j}\right)\!U^\top P^{1/2} \qquad j=1, ..., pH \ .
\end{align*}
The eigenvalues $\omega_i$ are the generalized eigenvalues of $(S,P)$:
\begin{align*}
    Su_j = \omega_j P u_j, \qquad
    \omega_j = \frac{u_j^\top S u_j}{u_j^\top P u_j}
    = \frac{\tau^2\,u_j^\top S u_j}{u_j^\top \Psi^{-1}u_j} \ .
\end{align*} 
Defining the effective local scale
\[
\psi_{\mathrm{eff},j}^2(u) := \frac{1}{u_j^\top \Psi^{-1}u_j} \ ,
\]
we can rewrite
\begin{align}
\frac{\omega_j}{1+\omega_j}
&= 1-\frac{1}{1+\psi_{\mathrm{eff},j}^2(u)\,\tau^{2}\,u_j^\top S u_j} \ ,
\label{eq:mode-shrink}
\end{align}
identifying the mode-wise shrinkage factor. To relate this to \citet{piironen2017sparsity}, recall
\begin{align*}
S = \mathbf{J}^{\top} \Sigma_y^{-1} \mathbf{J}, \qquad
\Sigma_y = \mathbf{J}_b \mathbf{J}_b^{\top}+ \Phi_0\Phi_0^{\top}+ \mathbf{1}_n\mathbf{1}_n^{\top} + \sigma^2 I_n 
= QQ^{\top}+ \sigma^2 I_n \ ,
\end{align*}
where \(Q = [\,\mathbf{J}_b\ \ \Phi_0\, \ \mathbf{1}_n ]\).  
Applying the Woodbury identity gives
\begin{align*}
\Sigma_y^{-1}
&= \sigma^{-2}\!\left(I - Q(\sigma^2 I + Q^\top Q)^{-1}Q^\top\right) \ .
\end{align*}
Since \(Q Q^{\top}\succeq 0\), we have \(\Sigma_y = \sigma^2 I + Q Q^{\top}\succeq \sigma^2 I\),
which implies
\[
\Sigma_y^{-1} \preceq \sigma^{-2} I \ .
\]
Conversely, because \(\lambda_{\max}(Q Q^{\top}) = \lVert Q\rVert_2^2\),
the largest eigenvalue of \(\Sigma_y\) satisfies
\(\lambda_{\max}(\Sigma_y) \le \sigma^2 + \lVert Q\rVert_2^2\),
and thus
\[
\Sigma_y^{-1} \succeq \frac{1}{\sigma^2 + \lVert Q\rVert_2^2}\,I \ .
\]
Combining these inequalities gives the spectral bounds
\begin{align*}
\frac{1}{\sigma^2+\lVert Q\rVert_2^2}I \preceq \Sigma_y^{-1} \preceq \sigma^{-2}I \ ,
\qquad
\frac{1}{\sigma^2+\lVert Q\rVert_2^2}\mathbf{J}^{\top} \mathbf{J} \preceq S \preceq \frac{1}{\sigma^2}\mathbf{J}^{\top} \mathbf{J} \ ,
\end{align*}
and for any unit vector $v$,
\[
\frac{\|\mathbf{J}v\|_2^2}{\sigma^2+\|Q\|_2^2}
\le v^\top S v
\le \sigma^{-2}\|\mathbf{J}v\|_2^2 \ .
\]
Furthermore, let $A_0:=XW_{1, 0}^{\top} + \mathbf{1}_n\mathbf{b}_{1, 0}$ denote the activation in the reference point, and define the elementwise derivative matrix $\boldsymbol{\Phi}'_0 := \varphi'(A_0) \in \mathbb{R}^{n \times H}$. Then let $R:=\boldsymbol{\Phi}'_0 \mathrm{diag}(\mathbf{w}_{L, 0}) \in \mathbb{R}^{n \times H}$ with columns $R_h=\mathbf{w}_{2, 0, h}\boldsymbol{\Phi}'_{0, h}$, to obtain
\begin{align}
\label{eq:Jacobian}
\mathbf{J} = \frac{\partial (\Phi(\mathbf{w}_1,\mathbf{b}_1)\mathbf{w}_2)}{\partial\mathbf{w}_{1}}
&= \big[\mathrm{diag}(R_{1})X \cdots \mathrm{diag}(R_{H})X \big] \ .
\end{align}
Each block $\mathrm{diag}(R_h)X$ corresponds to one hidden unit and contributes one row per data point. 
Hence, for any Euclidean unit vector $v\in\mathbb{R}^{pH}$,
\[
v^{\top}\mathbf{J}^{\top}\mathbf{J}v = \|\mathbf{J}v\|_2^2 = \sum_{i=1}^n (J_i v)^2 \ ,
\]
which shows that $\|\mathbf{J}v\|_2^2 = \Theta(n)$ whenever the rows of $\mathbf{J}$ have bounded norm. If the same bounded-rows argument applies to the columns of $Q$, then $\|Q\|_2^2 = \Theta(n)$. Consequently, $v^{\top}Sv$ scales approximately linearly with $n$ if $X$ is approximately orthonormal with bounded rows, $|\varphi'| \leq 1$, $H$ is fixed and $\mathbf{w}_2$ is bounded. Recalling that $u$ denotes the generalized eigenvectors of $(S,P)$ satisfying $Su=\omega Pu$, the mode-wise shrinkage \eqref{eq:mode-shrink} satisfies
\begin{align*}
1-\frac{1}{1+\psi_{\mathrm{eff}, j}^2(u)\tau^{2}\tfrac{\Theta(n)}{\sigma^2+\Theta(n)}}
\;\le\;
1-\frac{1}{1+\psi_{\mathrm{eff}, j}^2(u)\tau^{2}u_j^\top S u_j}
\;\le\;
1-\frac{1}{1+\psi_{\mathrm{eff}, j}^2(u)\tau^{2}\sigma^{-2}\Theta(n)} \ .
\end{align*}
This mirrors the scalar Piironen form 
\(\kappa_j = 1/(1+n\sigma^{-2}\tau^2 s_j^2\lambda_j^2)\) exactly.

\subsubsection*{Empirical analysis of the shrinkage matrix}
\label{ap:shrinkage_imposed}
To analyse the shrinkage matrix, we use all our $4000$ posterior samples of parameters as reference points inserted in the linearization. This yields one set of matrices for all samples, and these are what we now look at. 
\subsubsection*{Friedman}
\begin{figure}[tbp]
    \centering
    \subfloat[Independent features\label{fig:eigencurve_friedman_independent}]{
        \includegraphics[width=0.46\linewidth]{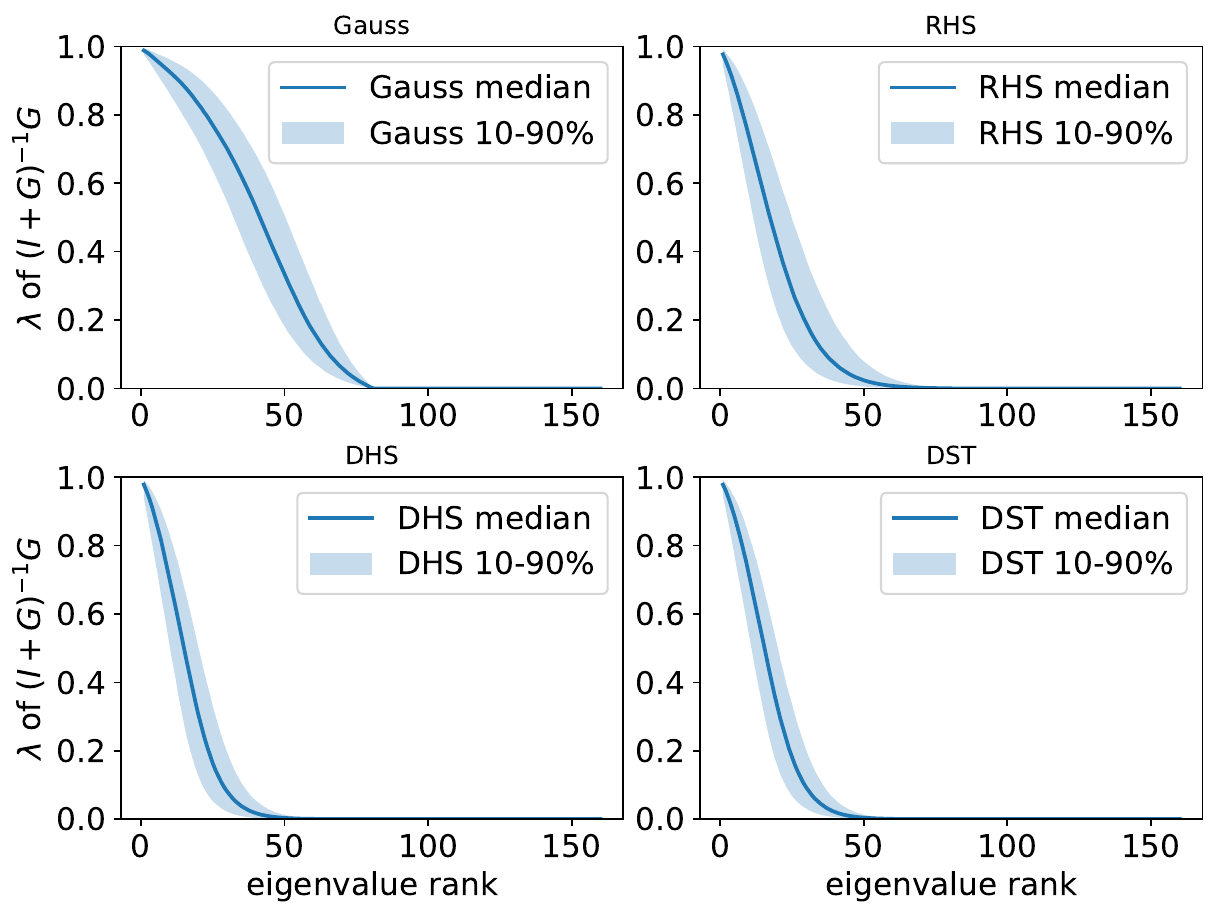}
    }
    \hfill
    \subfloat[Correlated features\label{fig:eigencurve_friedman_correlated}]{
        \includegraphics[width=0.46\linewidth]{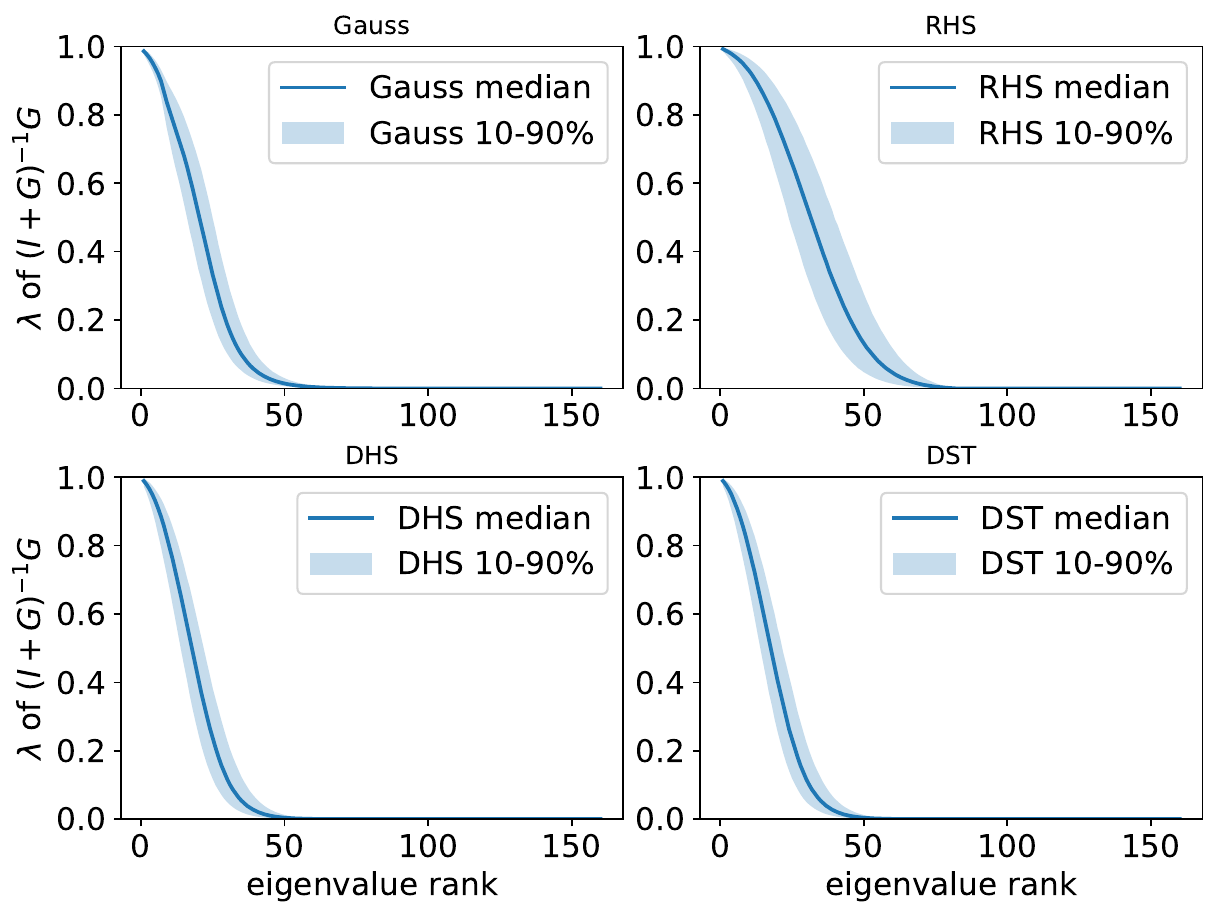}
    }
    \caption{Eigenvalue spectra of $(I+G)^{-1}G$ across different priors on the Friedman dataset under two input settings.}
    \label{fig:eigencurve_friedman_side_by_side}
\end{figure}

We also give the sorted eigenvalue curve for the whitened shrinkage matrix $(I+G)^{-1}G$ in Figure \ref{fig:eigencurve_friedman_side_by_side}, which shows that DSM priors yield shrinkage matrices with far more sparse eigenvalues.

\newpage

\subsubsection*{Abalone}
\begin{figure}[htbp]
    \centering
    \includegraphics[width=0.6\linewidth]{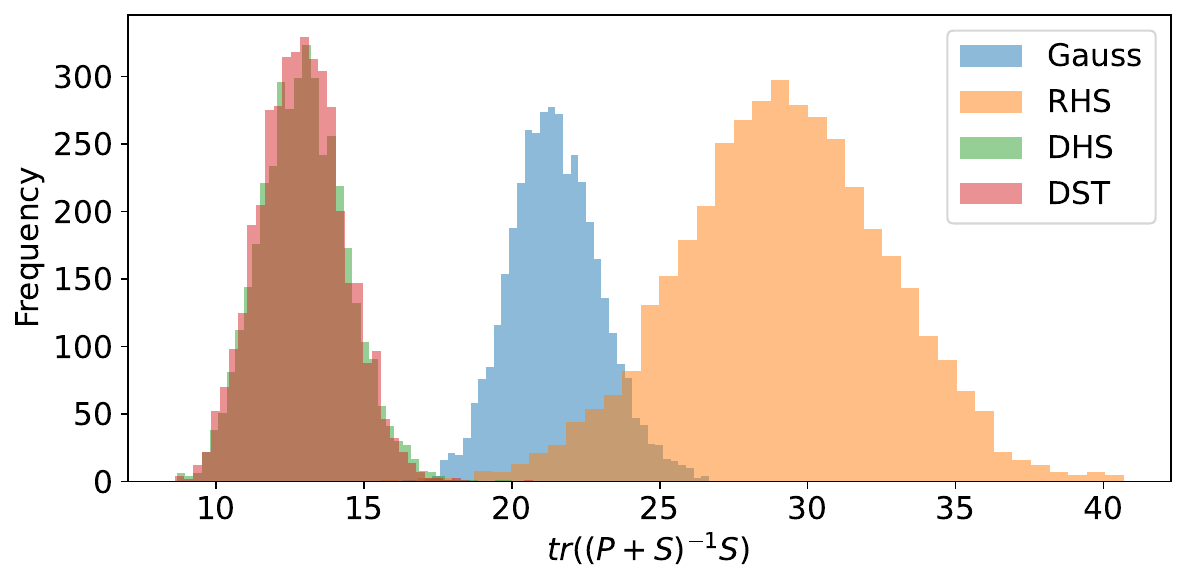}
    \caption{The effective number of non-zero parameters for the different models, as calculated from $m_{\mathrm{eff}} = \mathrm{tr}\left( (P+S)^{-1}S \right)$}
    \label{fig:m_eff_abalone}
\end{figure}

\begin{figure}[htbp]
    \centering
    \includegraphics[width=0.6\linewidth]{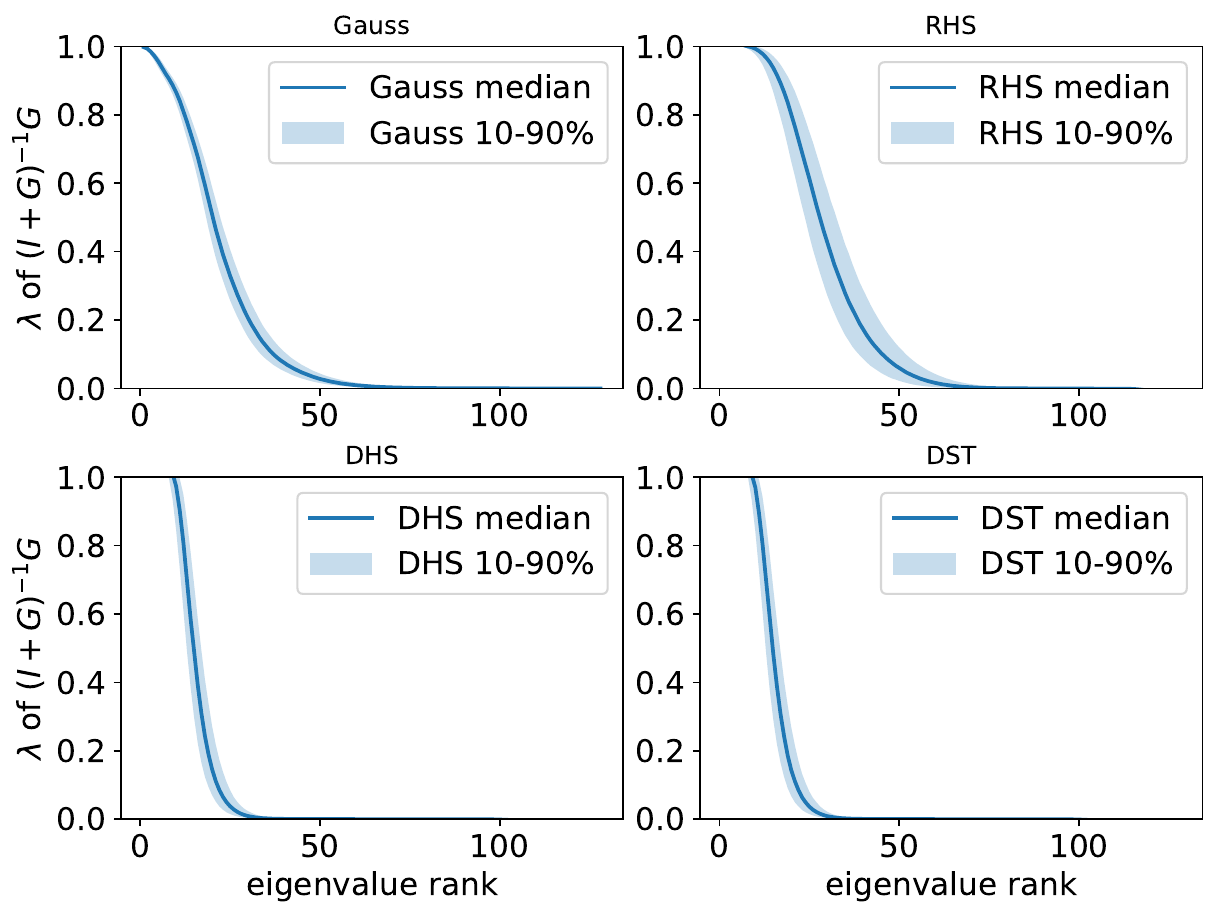}
    \caption{Eigenvalue curve}
    \label{fig:eigencurve_abalone}
\end{figure}

We now perform the same complexity analysis for the Abalone models as was done for the Friedman models. A particularly interesting aspect of the Abalone model is how much more sparsifiable the Dirichlet models are, compared to the Gaussian and the regularized horseshoe. For the Gaussian, this was perhaps expected, but for the regularized horseshoe the poor performance is not obvious. It is surprising to see that that to model the Abalone dataset the regularized horseshoe model needs even more effective parameters than the Gaussian model (Figure \ref{fig:m_eff_abalone}). The estimated number of nonzero parameters are still far less for the Dirichlet models than for the Gaussian model. This can also be seen from the eigenvalue curves in Figure \ref{fig:eigencurve_abalone}, where the regularized horseshoe model produces far more non-zero eigenvalues than the Dirichlet models and the Gaussian model.

\subsection*{Supplementary D: Additional results}
\label{ap:supplementary}
Here we include supplementary material, additional results and convergence diagnostics.

\subsubsection*{Experimental details}
We conducted several additional checks to assess the sensitivity of the models to alternative scaling choices. In particular, we verified that replacing the sample size $N$ by the hidden-layer width $H$ in the definition of $\tau_0$ did not lead to qualitatively different posterior behavior. For Dirichlet-based priors, the normalization constraint $\sum_{i=1}^p \xi_{ji} = 1$ alters the marginal scale of the conditional variances, since $\mathbb{E}[\xi_{ji}] = 1/p$ under a symmetric Dirichlet prior. We therefore considered rescaling the global parameter $\tau$ by a factor of $\sqrt{p}$ to match the marginal variance of the standard horseshoe prior. In practice, this adjustment had negligible impact on posterior shrinkage or predictive behavior, and all results in the main text are reported without this rescaling. 

\subsubsection*{Dirichlet and Beta type priors}
\label{ap:Dirichlet_vs_Beta}
Our theoretical investigations tackle the marginal shrinkage imposed by the DSM priors. This exploits that the components of a symmetric Dirichlet distribution marginally follow a Beta distribution. It is therefore natural to compare the DSM priors to the pure marginal model, defined by 
\begin{equation*}
\begin{aligned}
w_{jk} \mid \tau, \lambda_j, \xi_{jk}
&\sim \mathcal{N}\!\left(0,\, \tau^2 \lambda_j^2 \xi_{jk}\right), \\
(\xi_{j1}, \dots, \xi_{jp})
&\sim \mathrm{Beta}(\alpha, (p-1)\alpha), \\
\lambda_j &\sim \mathcal{P}_{\lambda}, \\
\tau &\sim \mathcal{P}_{\tau} \ .
\end{aligned}
\end{equation*}
To compare, we investigate the Beta Horseshoe prior, in which $\mathcal{P}_{\lambda}=C^+(0, 1)$ and the Beta Student's T prior, in which $\mathcal{P}_{\lambda}=t^+_{3}(0, 1)$. As mentioned previously, two components of a symmetric Dirichlet distribution have a correlation determined solely by $p$, such that the number of covariates will be the largest contributor to the differences between marginal and joint effects.

\begin{figure}[htbp]
\centering
\begin{minipage}{0.48\linewidth}
    \centering
    \includegraphics[width=\linewidth]{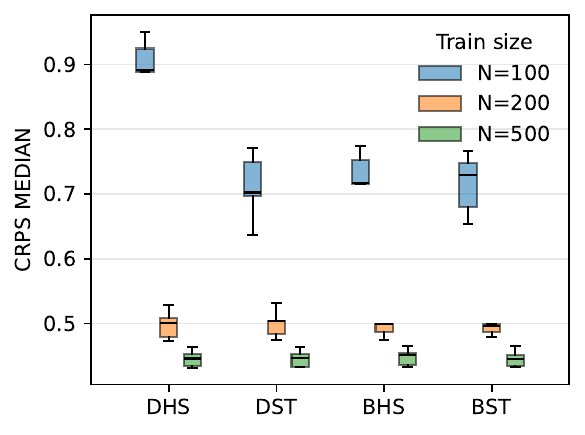}
\end{minipage}
\hfill
\begin{minipage}{0.48\linewidth}
    \centering
    \includegraphics[width=\linewidth]{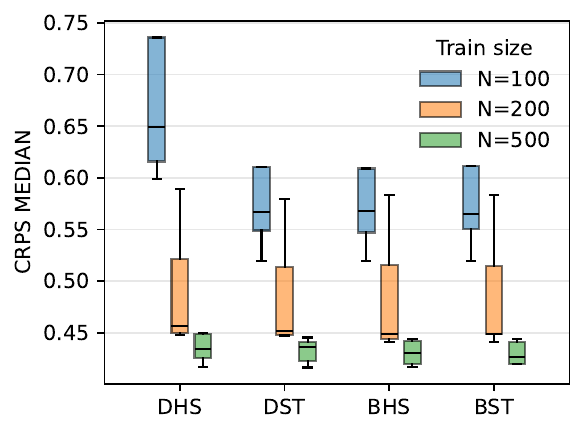}
\end{minipage}
\caption{
Boxplot of aggregated median CRPS across models and training sample sizes for the Friedman data.
}
\label{fig:Friedman_crps_beta}
\end{figure}

We present the same performance metrics on the Friedman data as previously seen (Figure \ref{fig:Friedman_crps_beta} and Table \ref{tab:friedman_rmse_beta}), but now compare the Dirichlet models with the Beta models. In terms of predictive performance the models are nearly indistinguishable. We present the same performance metrics on the Friedman data as previously seen, but now compare the Dirichlet models with the Beta models. In terms of predictive performance the models are nearly indistinguishable, except for the Dirichlet Horseshoe's performance for $N=100$ as we have already seen. This is not surprising, as all models induce function classes of comparable expressivity, and the primary role of the different priors is to regularize the parameter space rather than to fundamentally alter the representational capacity of the network.
\FloatBarrier
\begin{table}[htbp]
\centering
\begin{tabular}{lccc|ccc}
\toprule
& \multicolumn{3}{c|}{Uncorrelated} 
& \multicolumn{3}{c}{Correlated} \\
\cmidrule(lr){2-4} \cmidrule(lr){5-7}
Model 
& N=100 & N=200 & N=500
& N=100 & N=200 & N=500 \\
\midrule
DHS   
& 2.359 & \textbf{1.243} & \textbf{1.106}
& 1.846 & 1.232 & 1.057 \\
DST   
& \textbf{1.875} & 1.252 & 1.107
& 1.515 & 1.215 & 1.049 \\
BHS
& 1.912 & 1.247 & 1.107
& 1.513 & \textbf{1.214} & 1.049 \\
BST  
& 1.897 & 1.248 & 1.107
& \textbf{1.510} & 1.215 & \textbf{1.048} \\
\bottomrule
\end{tabular}
\caption{Comparison of aggregated posterior RMSE for different models and training sample sizes.}
\label{tab:friedman_rmse_beta}
\end{table}

\begin{figure}[htbp]
    \centering
    \subfloat[Independent features\label{fig:m_eff_friedman_independent_beta}]{
        \includegraphics[width=0.46\linewidth]{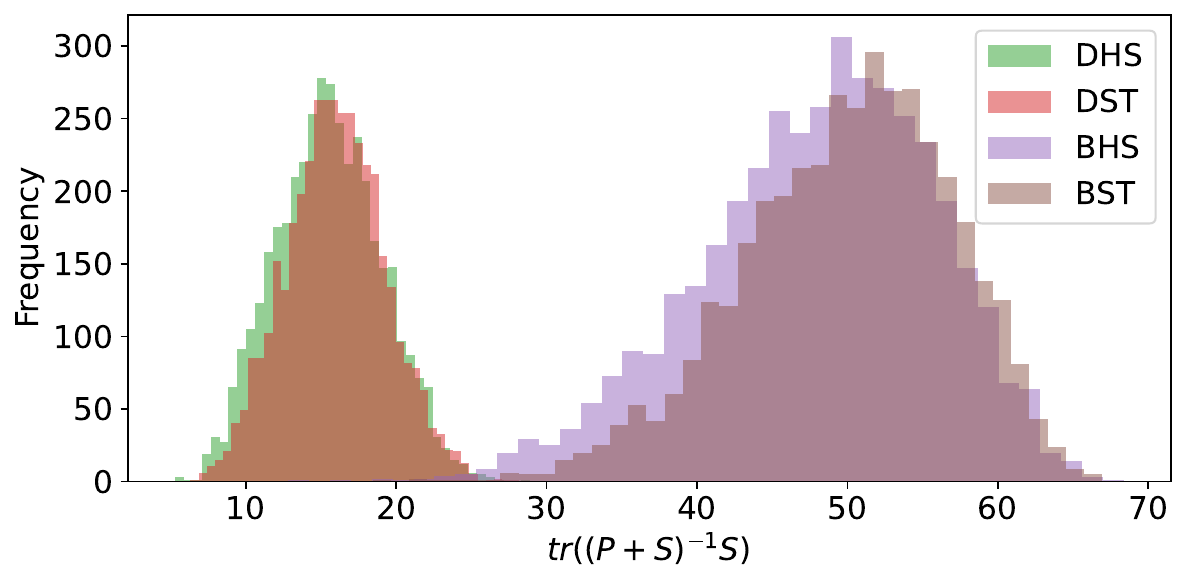}
    }
    \hfill
    \subfloat[Correlated features\label{fig:m_eff_friedman_correlated_beta}]{
        \includegraphics[width=0.46\linewidth]{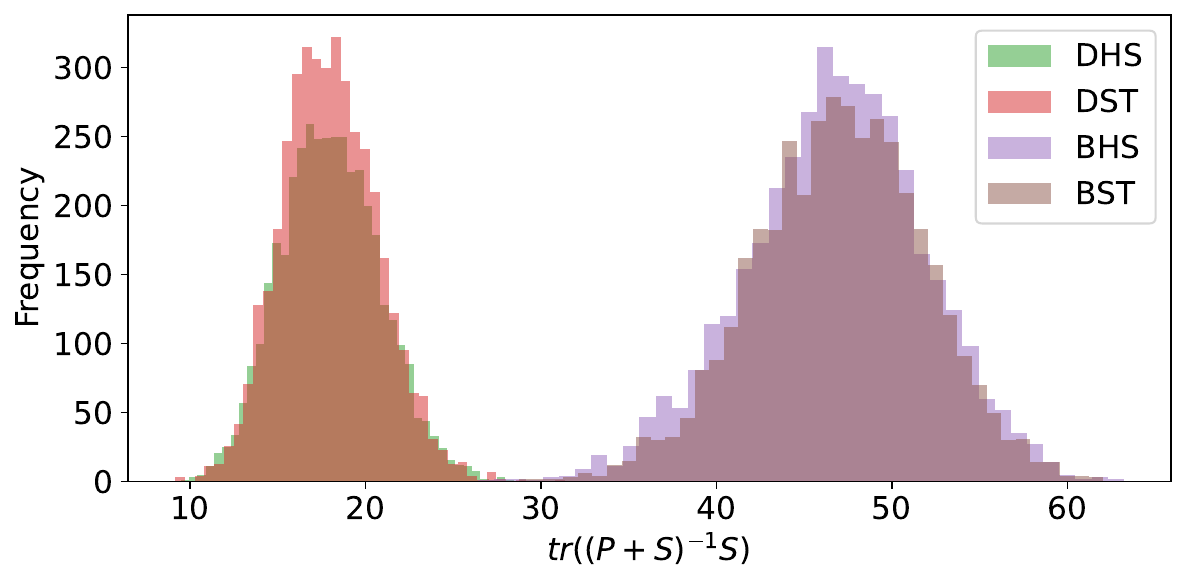}
    }
    \caption{Effective number of non-zero parameters $m_{\mathrm{eff}} = \mathrm{tr}\!\big((P+S)^{-1} S\big)$ for different models on the Friedman dataset with independent and correlated input features.}
    \label{fig:m_eff_friedman_side_by_side_beta}
\end{figure}

\noindent Consequently, differences between the priors are more naturally reflected in the modelling complexity and robustness to pruning. As seen from Figure \ref{fig:m_eff_friedman_side_by_side_beta}, the models using independent Beta distributions seem to use far more effective non-zero parameters than the Dirichlet models. This can possibly be attributed to the lack of constraints on the Beta variables, allowing more to be active simultaneously. 
\par
\begin{figure}[htbp]
    \centering
    \includegraphics[height=0.45\linewidth]{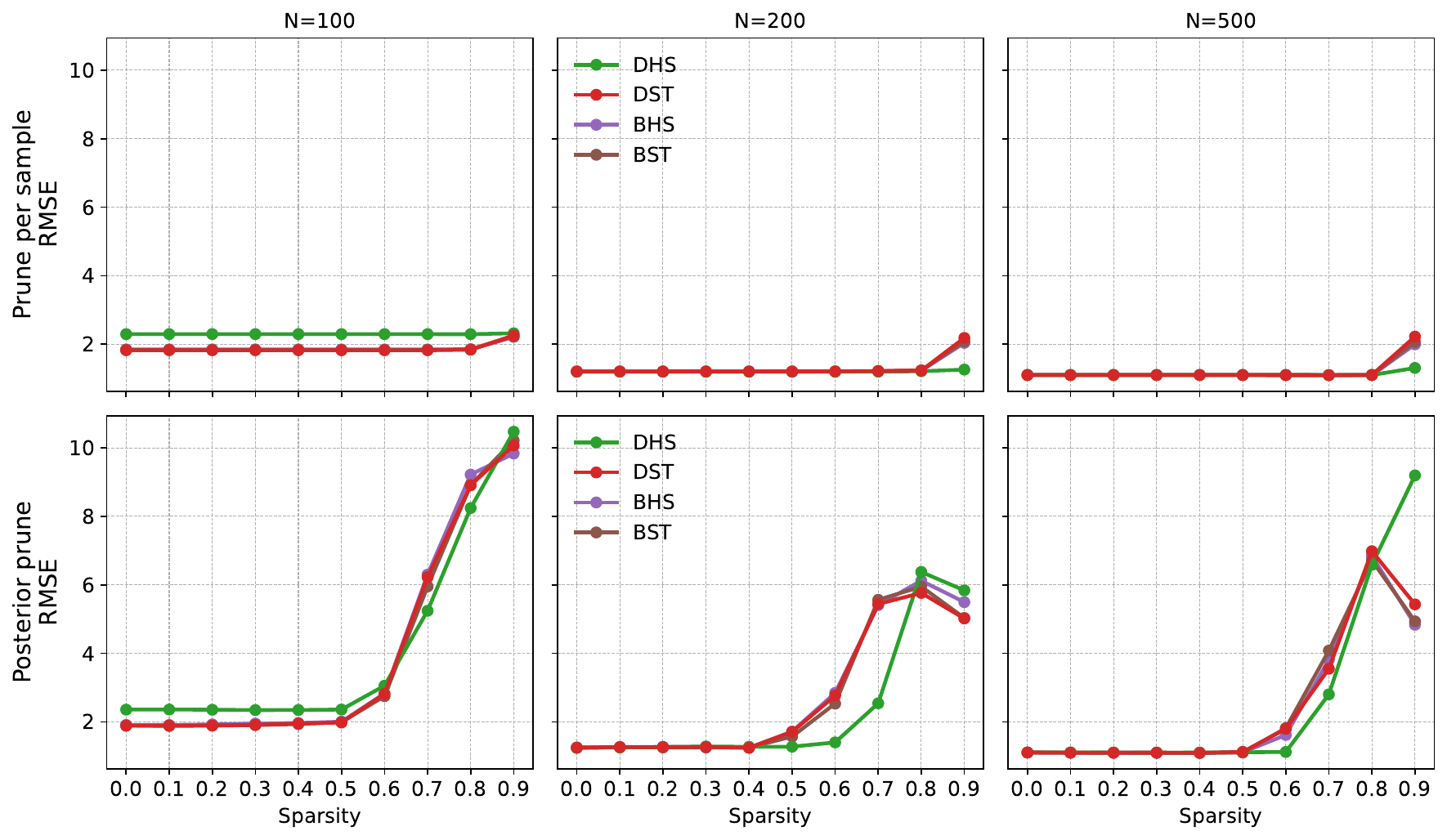}
    \caption{
    Posterior mean RMSE as a function of sparsity level in the BNNs on the independent Friedman datasets. The upper panels show results for the prune per sample scheme, whereas the lower panel shows posterior pruning.}
    \label{fig:Friedman_sparsity_beta}
\end{figure}

\vspace{0.5em}
\noindent Furthermore, Figure \ref{fig:Friedman_sparsity_beta} looks at the behaviour of the models when subject to pruning. We have previously observed that the DHS prior outperforms the Gaussian, RHS, and DST models. In the present comparison, one might expect the BHS prior to exhibit similar behavior. However, this is not the case. Instead, the DST, BST, and BHS models display broadly comparable pruning patterns, with the DHS prior remaining the only model that consistently is robust to intensive pruning. 
\FloatBarrier
\begin{figure}[htbp]
    \centering
    \includegraphics[height=0.45\linewidth]{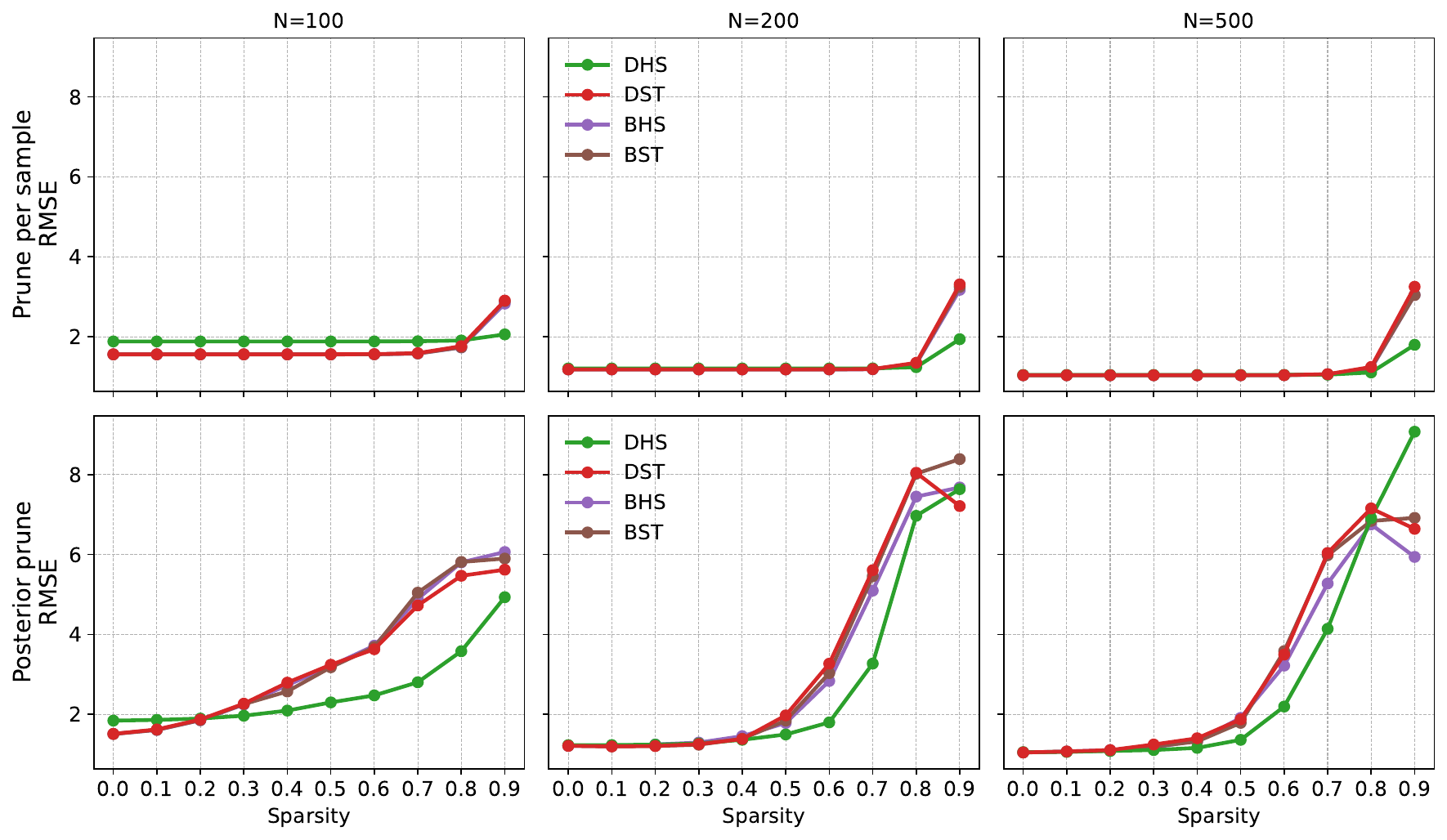}
    \caption{
    Posterior mean RMSE as a function of sparsity level in the BNNs on the correlated Friedman datasets. The upper panels show results for the prune per sample scheme, whereas the lower panel shows posterior pruning.}
    \label{fig:Friedman_correlated_sparsity_beta}
\end{figure}

\subsubsection*{Friedman regression}
\begin{figure}[htbp]
    \centering
    \includegraphics[height=0.4\linewidth]{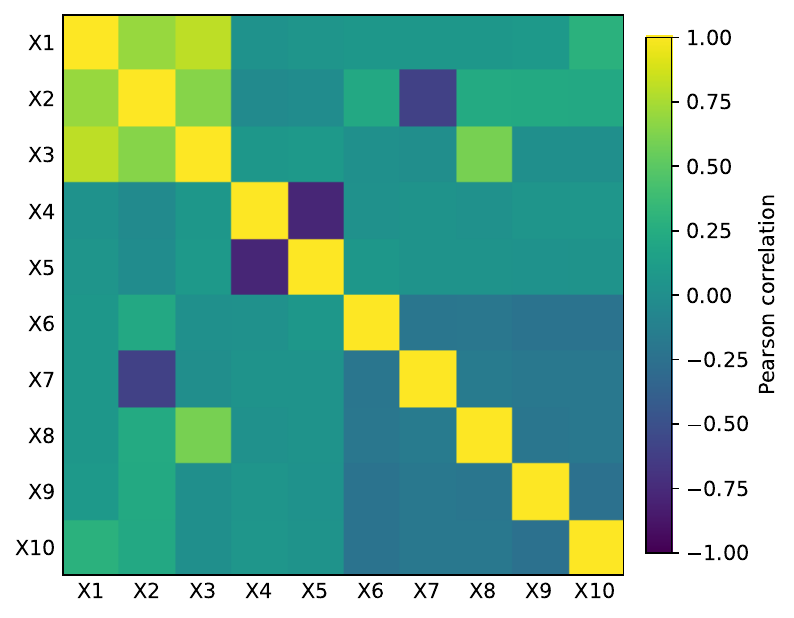}
    \caption{
    Correlation coefficient matrix for the Friedman dataset}
    \label{fig:Friedman_correlation_strucutre}
\end{figure}

In Figure \ref{fig:Friedman_correlation_strucutre} we display the correlation coefficient matrix used to generate the correlated Friedman data.

\label{appendix:friedman_additional}

\subsubsection*{Convergence results} 
\label{ap:convergence}
\noindent Convergence diagnostics for BNNs are a difficult task, as multimodality can hinder mixing of Markov chains for individual weights \citep{chandra_bayesian_2024}. As noted by \citet{chandra_bayesian_2024}, poor convergence of weights does not necessarily hinder good performance, but it does compromise the uncertainty quantification. To assess the convergence, we increase the burn-in period to $M_{\mathrm{warmup}}=5000$ and draw $M=2000$ samples per chain. 
\par
\begin{figure}[htbp]
    \centering
    \includegraphics[width=\linewidth]{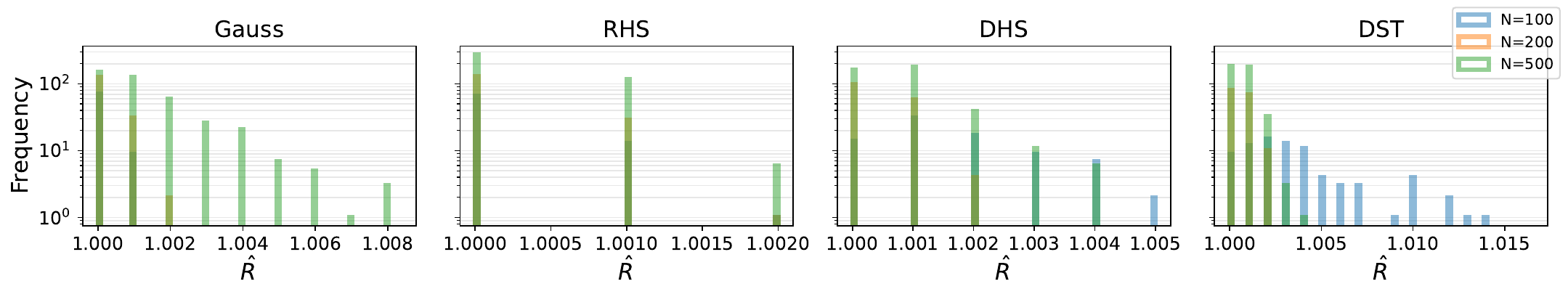}
    \caption{
    Plot of $\hat{R}$ for the network output variable across models and datasets for the Friedman dataset with $N \in \{100, 200, 500\}$}
    \label{fig:Friedman_rhats}
\end{figure}

\begin{figure}[htbp]
    \centering
    \subfloat[Gaussian\label{fig:trace_gauss}]{
        \includegraphics[width=0.46\linewidth]{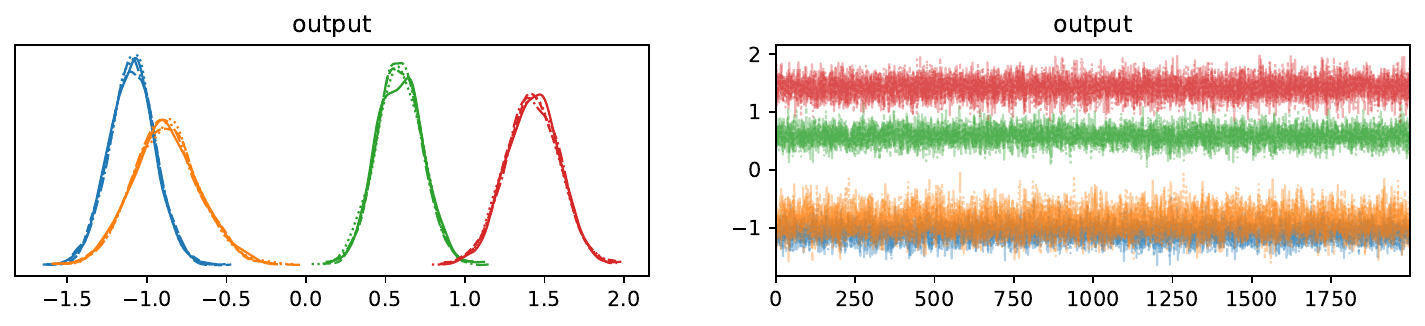}
    }
    \hfill
    \subfloat[Regularized Horseshoe\label{fig:trace_RHS}]{
        \includegraphics[width=0.46\linewidth]{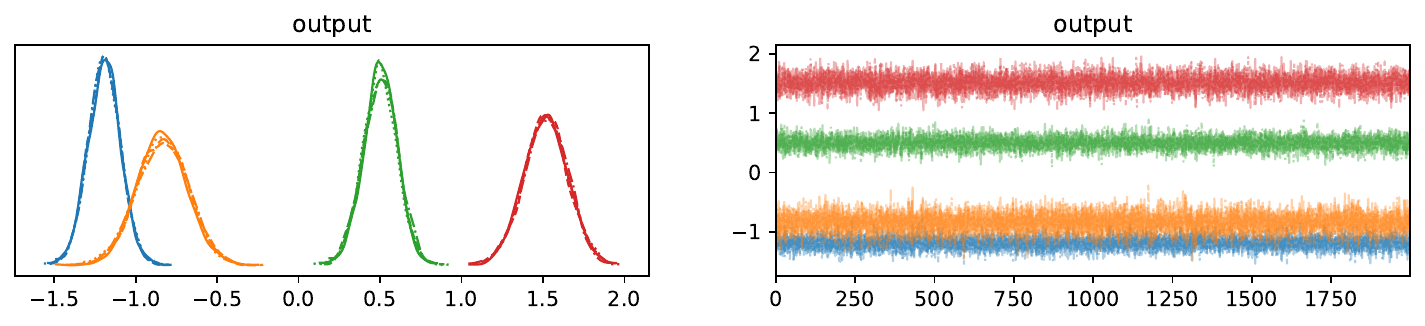}
    }
    \hfill
    \subfloat[Dirichlet Horseshoe\label{fig:trace_DHS}]{
        \includegraphics[width=0.46\linewidth]{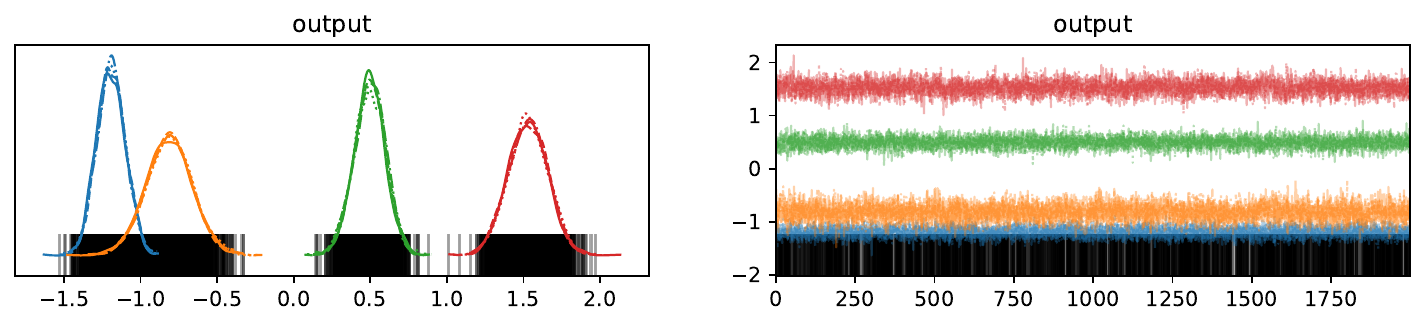}
    }
    \hfill
    \subfloat[Dirichlet Student T\label{fig:trace_DST}]{
        \includegraphics[width=0.46\linewidth]{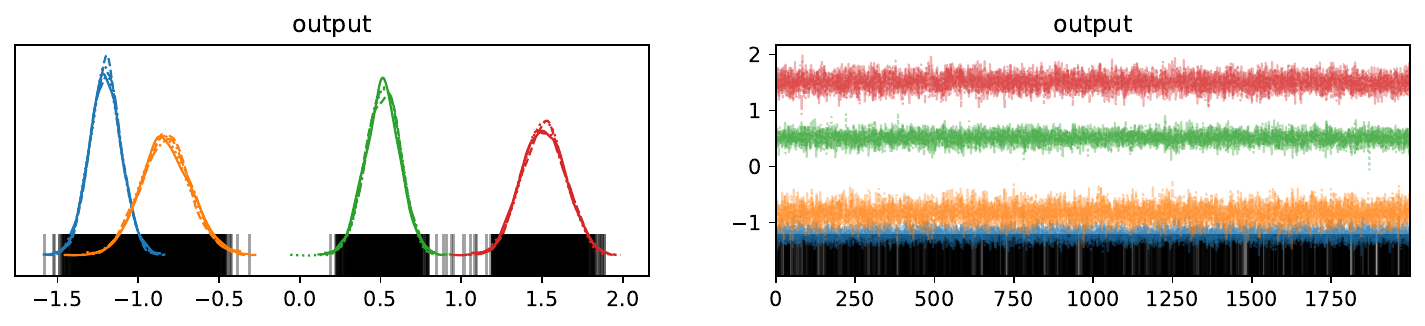}
    }
    \caption{Trace plots for the four priors.}
    \label{fig:trace_plots}
\end{figure}

\vspace{0.5em}
\noindent The convergence diagnostics are summarized in Table \ref{tab:Convergence_tanh}, a plot of $\hat{R}$ values are given in Figure \ref{fig:Friedman_rhats} and Figure \ref{fig:trace_plots} display traceplots of four output instances. The $\hat{R}$ values are computed for the output parameters, whereas the remaining diagnostics summarize behavior at the level of the sampled model parameters. This distinction is important, as the output-level diagnostics indicate reasonably good mixing and the trace plots suggest stable posterior behavior, while the corresponding diagnostics in weight space are generally weaker. This discrepancy is not unexpected in Bayesian neural networks and likely reflects a combination of structural non-identifiability, multimodality induced by symmetries in the parameterization, and the highly curved and anisotropic geometry of the posterior distribution. At the same time, clear differences across prior specifications are observed. 
The Gaussian and RHS priors exhibit comparatively favorable diagnostics, whereas the DSM priors show more challenging sampling behavior. To further investigate the reasons why, we explored a range of sampler configurations, including smaller step sizes, increased tree depths, alternative weakly informative hyperpriors, and less restrictive constraints in the parameterization.  While these adjustments generally improved convergence diagnostics in weight space, they did not lead to appreciable differences in predictive performance or posterior summaries at the output level. We therefore interpret the observed diagnostics primarily as indicative of the general challenges associated with sampling in deep Bayesian models, rather than as definitive evidence of pathological behavior of the proposed method.
\begin{table}[htbp]
\centering
\caption{Sampler diagnostics}
\begin{tabular}{l|c|c|c|c|c|c}
\toprule
Model & $\max\hat{R}$ & Med $\hat{R}$ & $N_{\mathrm{div}} / M$ & Med $\mathrm{ESS}_{\mathrm{tail}}/M$ & Med $\mathrm{ESS}_{\mathrm{bulk}}/M$ & $N$  \\
\midrule
Gauss & 1.001 & 1.000 & 0.004 & 0.901 & 0.895 & 100 \\
Gauss & 1.002 & 1.000 & 0.000 & 0.923 & 0.866 & 200 \\
Gauss & 1.009 & 1.001 & 0.000 & 0.853 & 0.610 & 500 \\
\midrule
RHS & 1.002 & 1.000 & 0.007 & 0.886 & 0.717 & 100 \\
RHS & 1.002 & 1.000 & 0.000 & 0.921 & 0.849 & 200 \\
RHS & 1.002 & 1.000 & 0.000 & 0.845 & 0.644 & 500 \\
\midrule
DHS & 1.005 & 1.001 & 0.586 & 0.560 & 0.236 & 100 \\
DHS & 1.002 & 1.000 & 0.224 & 0.794 & 0.574 & 200 \\
DHS & 1.004 & 1.001 & 0.186 & 0.753 & 0.487 & 500 \\ 
\midrule
DST & 1.021 & 1.003 & 0.346 & 0.544 & 0.235 & 100 \\
DST & 1.002 & 1.000 & 0.222 & 0.776 & 0.574 & 200 \\
DST & 1.004 & 1.001 & 0.160 & 0.759 & 0.502 & 500 \\
\bottomrule
\end{tabular}
\label{tab:Convergence_tanh}
\end{table}
\FloatBarrier
\newpage

\subsection*{Supplementary E: Stan code and hyperparameter details}
\label{ap:code}
In Stan, all hyperparameters in the DSM priors were generated directly from their respective priors. For each hidden unit, the node-specific scale $c_j^2$ was drawn from an $\mathrm{Inv\text{-}Gamma}(2,4)$ distribution, and the group scales $\lambda_{j}$ from independent $\mathrm{Cauchy}(0,1)$ distributions. The Dirichlet weights $\phi_j$ were sampled from a symmetric Dirichlet distribution with concentration parameter $\alpha=0.1$. All first-layer weights were constructed using a non-centred parameterization,
\begin{equation}
    \mathbf{w}_{j} = \tau \tilde{\lambda}_{j}\sqrt{\phi_{ji}}\, z_{ij},
    \qquad
    z_{ij} \sim \mathcal{N}(0,1),
\end{equation}
where $\tilde{\lambda}_{j}$ denotes the regularized local scale. Bias parameters and output weights were given standard normal priors, and the noise scale $\sigma$ an $\mathrm{Inv\text{-}Gamma}(3,2)$ prior. The prior guess $p_0$ can be modified based on the task. Below we include the Stan code for the DHS with tanh activation, which can easily be modified by changing activation, group scales and local scales to obtain the models used in the paper. The full repository can be found on the authors github, \url{https://github.com/AugustArnstad/DirichletScaleMixtures}.
\begin{lstlisting}[language=Stan, basicstyle=\ttfamily\small]
// =====================
// Prior predictive model with non-centered parameterization
// =====================

functions {
  matrix nn_predict(matrix X,
                    matrix W_1,
                    array[] matrix W_internal,
                    array[] row_vector hidden_bias,
                    matrix W_L,
                    row_vector output_bias,
                    int L) {
    int N = rows(X);
    int output_nodes = cols(W_L);
    int H = cols(W_1);
    array[L] matrix[N, H] hidden;

    hidden[1] = tanh(X * W_1 + rep_vector(1.0, N) * hidden_bias[1]);

    if (L > 1) {
      for (l in 2:L)
        hidden[l] = tanh(hidden[l - 1] * W_internal[l - 1] 
                    + rep_vector(1.0, N) * hidden_bias[l]);
    }

    matrix[N, output_nodes] output = hidden[L] * W_L;
    output += rep_matrix(output_bias, N);
    return output;
  }
}

data {
  int<lower=1> N;
  int<lower=1> P;
  matrix[N, P] X;
  int<lower=1> output_nodes;
  matrix[N, output_nodes] y;
  
  int<lower=1> L;
  int<lower=1> H;
  
  int<lower=1> N_test;
  matrix[N_test, P] X_test;
  
  int<lower=1> p_0;
  real<lower=0> a;
  real<lower=0> b;
  vector<lower=0>[P] alpha;
}

parameters {
  vector<lower=0>[H] lambda_node;   
  array[H] simplex[P] phi_data;               
  real<lower=1e-6> tau;
  vector<lower=0>[H] c_sq;

  matrix[P, H] W1_raw;

  array[max(L - 1, 1)] matrix[H, H] W_internal;
  array[L] row_vector[H] hidden_bias;
  matrix[H, output_nodes] W_L;
  row_vector[output_nodes] output_bias;
  real<lower=1e-6> sigma;
}


transformed parameters {
  real<lower=1e-6> tau_0 = (p_0 * 1.0) / (P - p_0) * 1 / sqrt(N);

  vector<lower=0>[H] lambda_tilde_node;

  for (j in 1:H) {
    lambda_tilde_node[j] = fmax(
      1e-12,
      c_sq[j] * square(lambda_node[j]) /
      (c_sq[j] + square(lambda_node[j]) * square(tau))
    );
  }

  matrix[P, H] W_1;
  for (j in 1:H) {
    for (i in 1:P) {
      real stddev = fmax(1e-12, tau *
                sqrt(lambda_tilde_node[j]) * sqrt(phi_data[j][i])) 
                / sqrt(P);
      W_1[i, j] = stddev * W1_raw[i, j];
    }
  }

  matrix[N, output_nodes] output = nn_predict(X, W_1, 
                    W_internal, hidden_bias, 
                    W_L, output_bias, L);
}


model {
  tau ~ cauchy(0, tau_0);
  c_sq ~ inv_gamma(a, b);

  lambda_node ~ cauchy(0, 1);
  for (j in 1:H)
    phi_data[j] ~ dirichlet(alpha);

  to_vector(W1_raw) ~ normal(0, 1);

  if (L > 1) {
    for (l in 1:(L - 1)) {
      for (j in 1:H) {
        W_internal[l][, j] ~ normal(0, 1);
      }
    }
  }

  for (l in 1:L)
    hidden_bias[l] ~ normal(0, 1);

  for (j in 1:output_nodes)
    W_L[, j] ~ normal(0, 1);

  output_bias ~ normal(0, 1);
  sigma ~ inv_gamma(3, 2);

  // Likelihood
  for (n in 1:N)
    for (j in 1:output_nodes)
      y[n, j] ~ normal(output[n, j], sigma);
}

\end{lstlisting}

\newpage

\end{document}